\documentclass{article}

\usepackage{PRIMEarxiv}

\usepackage[utf8]{inputenc} 
\usepackage[T1]{fontenc}    
\usepackage{hyperref}       
\usepackage{url}            
\usepackage{booktabs}       
\usepackage{amsfonts}       
\usepackage{nicefrac}       
\usepackage{microtype}      
\usepackage{lipsum}
\usepackage{fancyhdr}       
\usepackage{graphicx}       
\graphicspath{{media/}}     

\usepackage{caption}
\usepackage{float}
\usepackage{glossaries}
\usepackage{subcaption}

\usepackage{xcolor}
\usepackage{multirow}
\usepackage{array}
\usepackage{rotating}
\usepackage{tabularx}

\newacronym{audioprotopnet}{AudioProtoPNet}{Audio Prototypical Part Network}
\newacronym{auroc}{AUROC}{Area Under the Receiver Operating Characteristic Curve}
\newacronym{cmap}{cmAP}{Class Mean Average Precision}
\newacronym{cnn}{CNN}{Convolutional Neural Network}
\newacronym{dl}{DL}{Deep Learning}
\newacronym{xai}{XAI}{Explainable Artificial Intelligence}
\newacronym{ml}{ML}{Machine Learning}
\newacronym{pam}{PAM}{Passive Acoustic Monitoring}
\newacronym{ppnet}{ProtoPNet}{Prototypical Part Network}

\newcolumntype{M}[1]{>{\centering\arraybackslash}m{#1}}
\newcolumntype{Y}{>{\centering\arraybackslash}X}

\title{AudioProtoPNet: An interpretable deep learning model for bird sound classification}

\author{
  René Heinrich\textsuperscript{1,2,*}, Lukas Rauch\textsuperscript{2}, Bernhard Sick\textsuperscript{2}, Christoph Scholz\textsuperscript{1,2} \\ \\
  \textsuperscript{1} Fraunhofer Institute for Energy Economics and Energy System Technology (IEE) \\ 
  Joseph-Beuys-Straße 8, 34117 Kassel, Germany \\ \\
  \textsuperscript{2} Intelligent Embedded Systems (IES), University of Kassel \\
  M\"onchebergstraße 19, 34127 Kassel, Germany \\ \\
  * rene.heinrich@iee.fraunhofer.de \\
}

\begin{document}
\maketitle

\begin{abstract}
Deep learning models have significantly advanced acoustic bird monitoring by being able to recognize numerous bird species based on their vocalizations. 
However, traditional deep learning models are black boxes that provide no insight into their underlying computations, limiting their usefulness to ornithologists and machine learning engineers. 
Explainable models could facilitate debugging, knowledge discovery, trust, and interdisciplinary collaboration. 
This study introduces AudioProtoPNet, an adaptation of the Prototypical Part Network (ProtoPNet) for multi-label bird sound classification. 
It is an inherently interpretable model that uses a ConvNeXt backbone to extract embeddings, with the classification layer replaced by a prototype learning classifier trained on these embeddings. 
The classifier learns prototypical patterns of each bird species' vocalizations from spectrograms of training instances. 
During inference, audio recordings are classified by comparing them to the learned prototypes in the embedding space, providing explanations for the model's decisions and insights into the most informative embeddings of each bird species. 
The model was trained on the BirdSet training dataset, which consists of 9,734 bird species and over 6,800 hours of recordings. 
Its performance was evaluated on the seven test datasets of BirdSet, covering different geographical regions. AudioProtoPNet outperformed the state-of-the-art model Perch, achieving an average AUROC of 0.90 and a cmAP of 0.42, with relative improvements of 7.1\% and 16.7\% over Perch, respectively. 
These results demonstrate that even for the challenging task of multi-label bird sound classification, it is possible to develop powerful yet inherently interpretable deep learning models that provide valuable insights for ornithologists and machine learning engineers.
\end{abstract}


\section{Introduction}
\label{sec:introduction}
Global biodiversity has declined dramatically in recent decades. 
A study by Cornell's Lab of Ornithology estimates the loss of approximately 3 billion wild birds in North America since 1970, a population decline of 29\% \cite{rosenberg2019decline}. 
Overall, the average abundance of native species in most major terrestrial habitats has decreased by at least 20\% since 1900, according to \cite{watson2019summary}. 
Changes in land and sea use, direct exploitation of natural resources, pollution, climate change, and invasive alien species are among the driving forces behind the recent loss of biodiversity worldwide \cite{jaureguiberry2022direct}. \\ 
In order to track and improve the effectiveness of efforts to combat biodiversity loss, a coherent global biodiversity monitoring system is essential \cite{butchart2010global}. 
Birds play a central role in this effort, as they are valuable indicators of environmental health and biodiversity in different habitats \cite{sekercioglu2019birds}. 
As a result, \gls*{pam} of birds has become increasingly important as it provides a cost-effective way to collect large amounts of acoustic data while minimizing disturbance to natural habitats \cite{ross2023passive, shonfield2017autonomous}. 
However, the traditional use of \gls*{pam} is time-consuming, requiring ornithologists to carefully analyze extensive audio recordings. 
The development of automated, computer-aided analysis methods is therefore of great interest and has contributed significantly to the emergence of the research field of computational avian bioacoustics \cite{stowell2022computational, kahl2021birdnet}. \\
However, despite advances in automation using \gls*{dl}, the extraordinary diversity and complexity of bird vocalizations remains a major challenge. 
The vocalizations of individual bird species include a variety of different sounds, including different songs, call types, regional dialects, and imperfect juvenile vocalizations.  
Although \gls*{dl} models have been shown to be particularly effective at classifying bird sounds \cite{stowell2019automatic, kahl2021birdnet}, their black-box nature makes it unclear what types of bird vocalizations they can actually recognize.
While training models on fine-grained labels reflecting vocalization types and dialects would help to address this issue, labels are typically only available at the species level, rather than for specific vocalization types or dialects. 
Thus, to ensure the generalizability of a model to different real-world scenarios, it is important to determine whether a \gls*{dl} model has learned to recognize all the different types of songs and calls, as well as the different dialects of a bird species, or only some of the most common vocalizations.
Furthermore, it is important that \gls*{dl} models actually recognize birds based on their vocalizations and do not perform shortcut learning \cite{geirhos2020shortcut}, i.e., identify individual bird species based on spurious features such as certain background noise in audio recordings. 
Shortcut learning can severely limit the generalizability of a model, as it will no longer be able to recognize bird species in recordings that lack these spurious features. 
A better understanding of the types of vocalizations that the model has learned to recognize can greatly facilitate collaboration between ornithologists and \gls*{ml} engineers, helping them to validate, debug, and improve \gls*{dl} models.
In addition, this understanding can serve educational purposes by helping citizen scientists learn how to identify bird species, thereby improving the quality of volunteer contributions to citizen science platforms such as eBird \cite{sullivan2009ebird} or iNaturalist \cite{iNaturalist2024}.
Therefore, not only the performance of bird sound classification models is important, but also their explainability, as illustrated in Figure \ref{fig:illustration_importance_xai} and discussed in more detail in Appendix \ref{secA1:importance_xai}.

\begin{figure}[ht]
    \centering
    \includegraphics[trim={0cm 0cm 0cm 0cm}, clip, scale=0.5]{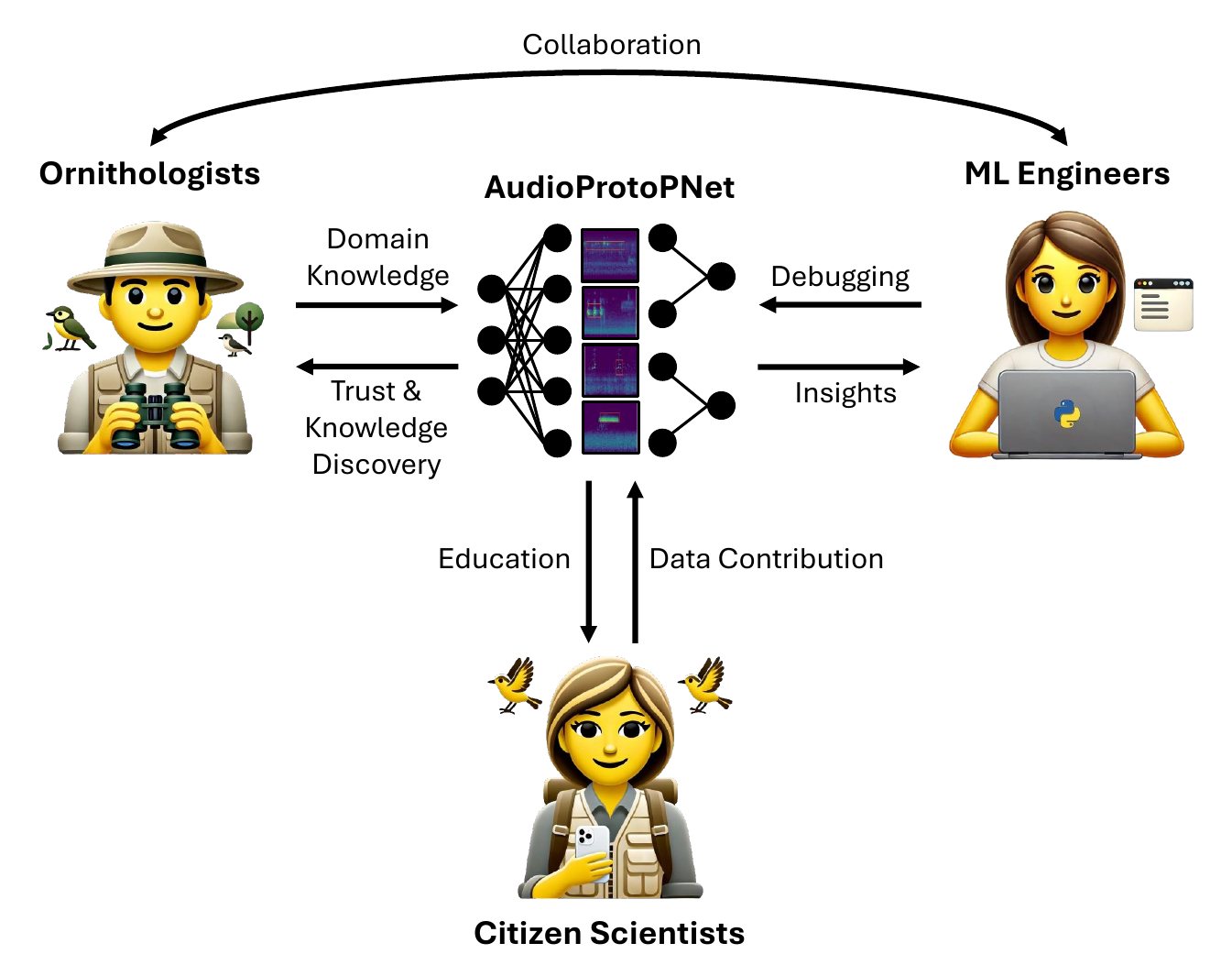}
    \caption{Illustration of the importance of inherently interpretable models such as AudioProtoPNet for different audiences of the bird sound classification task.
    }
\label{fig:illustration_importance_xai}
\end{figure}

In this respect, \gls*{xai} is particularly useful because it includes techniques that allow humans to understand how black-box models arrive at their classifications. 
Much of the research in \gls*{xai} focuses on post-hoc explanation approaches \cite{arrieta2020explainable}. 
Post-hoc explanation methods provide explanations after a black-box model has been trained and classifications have been made. 
Their main advantage is their model independence, which allows flexible application to a variety of black-box models. 
However, recent studies have shown the limitations of post-hoc explanation approaches \cite{rudin2019stop}. 
In particular, post-hoc explanation methods often locally approximate complex models with simpler, interpretable models to explain individual classifications. 
For example, LIME \cite{ribeiro2016should} is often used with a linear lasso regression model to locally approximate the original model around a given data point. 
However, these approximations can be subject to approximation error, so the generated explanations may not accurately reflect the actual model calculations. 
Such inconsistencies can undermine confidence in the explanations and thus in the models themselves. \\
These challenges highlight the need for inherently interpretable models. 
Inherently interpretable models are models whose architecture is designed so that explanations are an integral part of their classification processes. 
Unlike post-hoc explanation methods, these models provide explanations without approximation errors. 
Although it is often claimed that such models sacrifice performance for interpretability, the study by \cite{rudin2019stop} refutes this claim. 
A promising approach for inherently interpretable \gls*{dl} models are prototype learning models \cite{li2018deep}. These models learn to characterize each class through representative examples, called prototypes.
For each class, one or more prototypes are learned during training to represent the essential characteristics of that class. 
A prototype is a representative instance, such as an entire image in the case of image classification, or a characteristic part of an instance, such as a patch of an image, from the training data that is understandable to humans and shows the typical characteristics of a class. 
During training, these models automatically identify important training instances or parts of them as prototypes. 
When classifying new inputs, they compare them to the stored prototypes and make their decisions based on the similarity between the inputs and the prototypes.
The use of prototypes contributes significantly to the explainability of the models, since the classifications are based directly on understandable examples. 
Instead of having to interpret abstract feature spaces or complex model weights, the prototypes can be used to understand which features led to the classifications. 
This makes it easier not only to understand the model's classifications, but also to validate them and correct the model if necessary. 
A prominent example of a prototype learning model is \gls*{ppnet}, developed by \cite{chen2019looks}. 
\gls*{ppnet} has shown that it is possible to develop inherently interpretable models that are comparable to black-box models in terms of classification performance. 
It is an image classification architecture that learns prototypical parts of training images as representative prototypes of their respective classes. 
By using visually interpretable prototypes, \gls*{ppnet} enables a deep understanding of how the model makes its classifications. \\
This concept can easily be applied to the audio domain, as spectrograms are often used as input for audio classification models \cite{kahl2021birdnet, hamer2023birb}. 
Spectrograms are visual representations of audio signals that plot frequency content over time, converting audio data into an image-like format. 
They are also often used as input for bird sound classification models \cite{kahl2021birdnet,hamer2023birb}. 
By applying a prototype learning model such as \gls*{ppnet} to bird sound classification, intuitive explanations can be generated by learning prototypical sound patterns in the input spectrograms. 
For example, the model could learn prototypical sound patterns that represent specific song types, call types, or dialects from the sound repertoire of different bird species. 
For each bird species, one or more prototypes would be learned that represent characteristic sound patterns of that species.
When classifying a new audio recording, the model compares segments of the input spectrogram with these prototypes and classifies the recording based on its similarity to the prototypes. 
In addition, the prototypes learned by the model make it possible to identify the sound patterns that are crucial for the classifications. 
Thus, the prototypes not only serve as a basis for classifying new recordings, but also provide information about which specific features the model considers relevant.
In addition, the prototypes in the form of spectrograms or parts thereof can be converted into audio files using transformations such as the Griffin-Lim algorithm \cite{griffin1984signal, perraudin2013fast}, in order to analyze the classification results not only visually but also auditorily. 
This allows ornithologists to listen to the prototypes, which further increases explainability and promotes a deeper understanding of model behavior.

\subsection{Related Work}
While \gls*{xai} methods have been well studied in the field of image and text processing, there is little research on this topic in the context of audio data. 
In the field of post-hoc explanation methods for audio classification, several techniques have been developed to explain \gls*{dl} models. 
These include counterfactual explanations \cite{zhang2022towards}, gradient-based feature attribution methods \cite{becker2018interpreting, schiller2019relevance, lim2022detecting}, perturbation-based feature attribution methods \cite{dissanayake2020robust, weitz2021let, sanakkayala2022explainable}, and attention-based feature attribution methods \cite{won2019toward, pal2021pay, ren2022deep}. 
Much of this work is focused on medical applications. \\
Deep prototype learning was originally proposed by \cite{li2018deep} and further developed with \gls*{ppnet} by \cite{chen2019looks} to learn prototypical parts in images.
Since then, numerous extensions have been introduced to improve performance and explainability. 
Deformable \gls*{ppnet} \cite{donnelly2022deformable} allows prototypes to flexibly deform and adapt their spatial structure to better capture variations in pose and context.
ProtoTree \cite{nauta2021neural} combines prototype learning with decision trees. 
TesNet \cite{wang2021interpretable} uses a transparent embedding space based on category-aware orthogonal basis concepts. 
ProtoPool \cite{rymarczyk2022interpretable} simplifies training through a fully differentiable assignment of prototypes to classes. 
ProtoPFormer \cite{xue2022protopformer} and ProtoViT \cite{ma2024interpretable} adapt \gls*{ppnet} to vision transformers. 
PIP-Net \cite{nauta2023pip} uses prototypes learned in a self-supervised manner to better match human visual perception. 
In addition, ProtoConcepts \cite{ma2024looks} improves the explanations of \gls*{ppnet} by visualizing multiple image patches for each prototype. \\
The application of deep prototype learning to audio classification is a relatively new area of research. 
\cite{ren2022prototype} developed a deep prototype learning model for classifying human breath sounds to detect early signs of respiratory disease. 
In addition, \cite{zinemanas2021interpretable} explored the use of deep prototype learning to classify speech, music, and environmental sounds, achieving results comparable to state-of-the-art black-box models.
\cite{liu2023interpretability} developed a \gls*{ppnet} for binary audio classification of wood-boring pests. 
However, the models developed in these studies either represent the prototypes as full instances or are limited to binary classification problems. 
To date, there are no studies on the use of \gls*{ppnet} in the context of complex multi-label classification problems with thousands of classes, as required for bird sound classification.

\subsection{Contributions}
As described above, previous research in audio classification has focused on the development of prototype learning models, where the prototypes either represent whole instances or are designed for binary classification tasks only. 
In models where prototypes represent whole instances, these instances correspond to complete audio recordings from the training dataset. 
Typically, the training instances are in the form of spectrograms with a length of a few seconds \cite{kahl2021birdnet, hamer2023birb, rauch2024birdset}. 
This is particularly challenging for bird sound multi-label classification, because the recordings often contain multiple bird species, different overlapping bird vocalizations, as well as vocalizations from other animal species and environmental noise.
This complexity makes it difficult to assign prototypes that represent complete instances to specific sound patterns in an audio recording because it is not clear which of several sound patterns a prototype exactly represents, or whether it is just background noise. 
This problem severely limits the explainability of such models.
Especially in species-rich ecosystems such as the Amazon rainforest, which is home to about 1,300 bird species \cite{mittermeier2003wilderness}, vocalizations overlap considerably. 
To meet this challenge, multi-label classification approaches to bird sound classification are needed instead of simple binary or multi-class classification models \cite{rauch2024birdset}. 
On the other hand, it is crucial that the prototype learning model learns specific, characteristic parts of each class within the training instances and uses them for classification. 
This can improve both the performance of the model, by focusing on the discriminative features specific to each bird species, and the explainability, by clearly highlighting which sound patterns of an instance are represented by a prototype. \\
In this study, we address these challenges and make the following contributions:

\begin{enumerate}
    \item[(C1)] We present the \gls*{audioprotopnet}, a prototype learning classifier for bird sound classification that extends the work of \cite{chen2019looks} and \cite{donnelly2022deformable} to a multi-label audio classification setting. 
    Our classifier learns class-specific prototypes that represent characteristic parts of spectrograms from the training data.
    These prototypes correspond to different sound patterns associated with each bird species. 
    \gls*{audioprotopnet} can be applied to arbitrary translation-equivariant black-box models by replacing their original classifier with the prototype learning classifier. 
    Thus, it can be used either to transform black-box models into inherently interpretable models, or as a post-hoc explanation method to analyze their learned embeddings. 
    The latter is particularly beneficial when the original black-box model outperforms the inherently interpretable model.
    \item[(C2)] We demonstrate the scalability of our approach by successfully training \gls*{audioprotopnet} with a ConvNeXt backbone on the large BirdSet training dataset \cite{rauch2024birdset}, which contains over 6,800 hours of audio from nearly 10,000 different bird species.
    \item[(C3)] We evaluate \gls*{audioprotopnet} on the BirdSet test dataset \cite{rauch2024birdset}, which contains seven different datasets of bird sounds from different geographical regions. 
    Over the seven test datasets, \gls*{audioprotopnet} achieves an average \gls*{auroc} of 0.90 and an average \gls*{cmap} of 0.42. 
    {audioprotopnet} outperforms the current state-of-the-art black-box \gls*{dl} model Perch \cite{hamer2023birb} on most of the test datasets, as Perch only achieves an average \gls*{auroc} of 0.84 and an average \gls*{cmap} of 0.36.
    This corresponds to a relative improvement of 7.1\% in \gls*{auroc} and 16.7\% in \gls*{cmap} compared to Perch.
    \item[(C4)] We investigate the influence of choosing different numbers of prototypes per class on \gls*{audioprotopnet} and show how our prototype learning classifier can be used to qualitatively analyze the learned embeddings of a black-box ConvNeXt model. 
    We find that five prototypes per class are sufficient to represent the most important embeddings learned by the ConvNeXt model. 
    Our qualitative analysis also provides insight into the model's internal representations of the training and test data in the embedding space, thereby adding explainability.
\end{enumerate}

\section{Methodology}
\label{sec:methodology}
\begin{figure}[ht]
    \centering
    \includegraphics[trim={0cm 0cm 0cm 0cm}, clip, width=\textwidth]{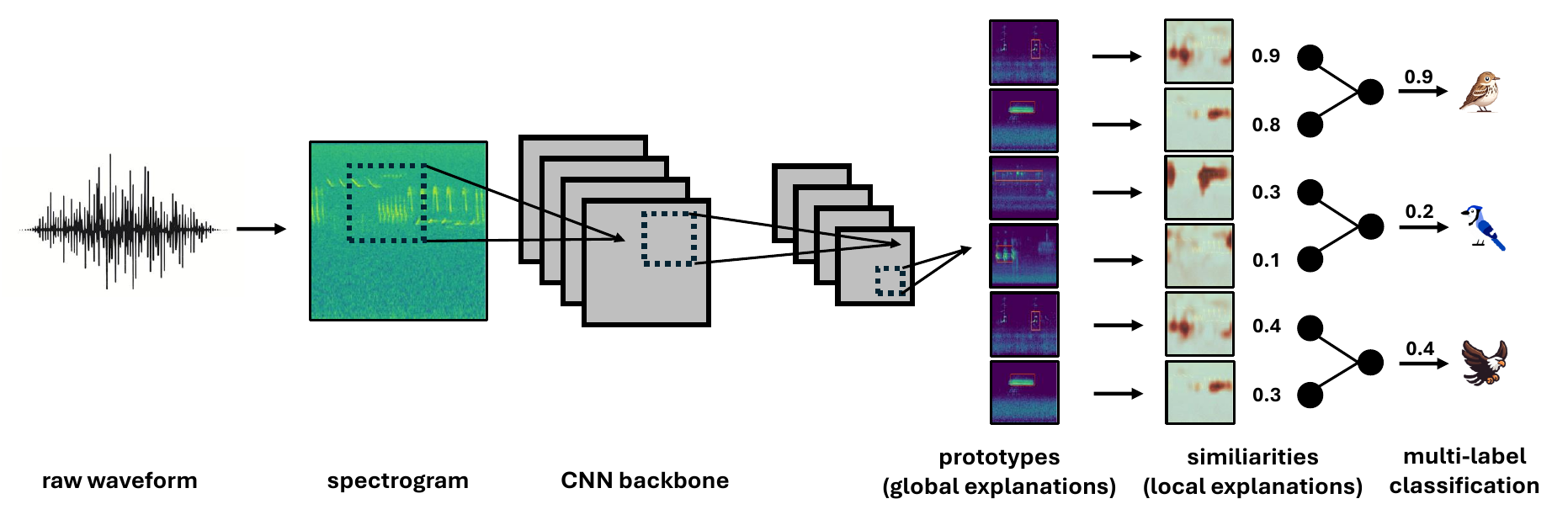}
    \caption{Illustration of the \gls*{audioprotopnet} inference process. The raw audio waveforms are first converted to spectrograms and then mapped to embedding space via a CNN backbone. There the similarities between the spectrogram embeddings and the learned prototypes are computed. While the similarities serve as local explanations for the models' classification, the learned prototypes themselves represent global explanations. Finally, the confidence score for each class is computed by weighting the similarities to the different prototypes of the respective class with a linear layer and then applying a sigmoid activation function.
    }
\label{fig:illustration_audioprotopnet}
\end{figure}

As contribution (C1), we present \gls*{audioprotopnet}, an adaptation of the \gls*{ppnet} architecture specifically designed to address the challenges of multi-label bird sound classification.
This section describes the architecture, training process, and visualization of the prototypes in \gls*{audioprotopnet}.
The inference process of \gls*{audioprotopnet} is illustrated in Figure \ref{fig:illustration_audioprotopnet}.

\subsection{Architecture}
Our model $f$ consists of a regular \gls*{cnn} backbone $f_b$, a prototype layer $f_p$, and a final linear layer $f_f$.
Given an input spectrogram $\mathbf x$, the model aims to predict multi-label ground truths $\mathbf{\hat{y}} = f(\mathbf{x})$, where $\mathbf{\hat{y}} \in [0, 1]^C$ and $C$ is the number of classes. 
The backbone model $f_b$ extracts a feature embedding $\mathbf{z} = f_b(\mathbf{x})$ from the input data. This embedding $\mathbf z \in \mathbb{R}^{H_z \times W_z \times D}$ is a feature map with a height of $H_z$, a width of $W_z$, and $D$ channels. 
The prototype layer contains several representative prototypes $\mathbf{P}^{(c)} = \{ \mathbf p^{(c,j)} \}_{j=1}^J$ for each class $c$, where $J$ is the number of prototypes per class. 
Each prototype has dimensions $H_p \times W_p \times D$, where $1 \leq H_p \leq H_z$ and $1 \leq W_p \leq W_z$. 
Following the approach in \cite{chen2019looks}, we set $H_p = W_p = 1$ in our work.
Each prototype $\mathbf p^{(c,j)}$ of the prototype layer is then compared to the embedding $\mathbf z$ at each spatial position $(h_z, w_z)$, where $1 \leq h_z \leq H_z$ and $1 \leq w_z \leq W_z$, by computing the cosine similarity in embedding space. 
The similarity score for a prototype $\mathbf{p}^{(c,j)}$ and the embedding $\mathbf z$ at position $\left( h_z, w_z \right)$ is defined as \cite{donnelly2022deformable}:
\begin{equation}
    s^{(c,j)}_{h_z, w_z} \left( \mathbf{z} \right) = \mathbf{\tilde{p}}^{(c,j)} \cdot \mathbf{\tilde{z}}_{h_z, w_z} = \frac{\mathbf{p}^{(c,j)}}{\left\|\mathbf{p}^{(c,j)}\right\|_2} \cdot \frac{\mathbf{z}_{h_z, w_z}}{\left\|\mathbf{z}_{h_z, w_z}\right\|_2},
\end{equation}
where $\mathbf{\tilde{z}}_{h_z, w_z}$ and $\mathbf{\tilde{p}}^{(c,j)}$ are the embedding patch and prototype, respectively, each scaled to a unit vector and thus having a norm of 1.
The similarity scores for an individual prototype form an activation map, which is reduced by global max-pooling to a single value that reflects the presence intensity of the prototype in the entire embedding of the input spectrogram:
\begin{equation}
    s^{(c,j)} \left( \mathbf{z} \right) = \max_{h_z, w_z} s^{(c,j)}_{h_z, w_z} \left( \mathbf{z} \right).
    \label{eq:max_similiarity_score}
\end{equation}
The resulting similarity scores are then processed by the final linear layer $f_f$, which uses a non-negative weight matrix $\mathbf{\omega} _f \in \mathbb R ^{J \times C}$, as suggested in \cite{nauta2023pip}.
Finally, we apply the sigmoid function to the logits to obtain an independent confidence score for each class. 
From these confidence scores, we can infer the classes to which the spectrogram belongs in the multi-label classification problem.
Using a non-negative weight matrix in the final linear layer $f_f$ ensures that the presence of a class-relevant prototype can only increase the confidence score for that class, not decrease it. 
This avoids negative reasoning and improves explainability. 
Each prototype is connected exclusively to its associated class, with no connections to other classes. 
This design ensures that predictions for a class are based only on the prototypes of that class. 
In this way, the model is forced to learn prototypes that capture unique features for a class that distinguish that class from other classes.
As in \cite{chen2019looks}, we initialize the connection weight of each prototype to its associated class with a value of 1, so that all prototypes are equally weighted at the start.
In addition, we initialize all bias terms of the final layer with -2. 
This initialization means that if an instance has no similarity to the prototypes of a class, the model will predict a logit of -2 for that class. 
Applying the sigmoid function to this logit yields a confidence score close to 0 for that class, which is appropriate in a multi-label setting where the absence of certain classes needs to be accurately reflected. 

\subsection{Training process}
Let $\mathbf{Y} \in \{0,1\}^{N \times C}$ be the binary matrix of true labels for a multi-label classification problem, where $N$ is the number of instances and $C$ is the number of classes. 
Each element $y_{n,c}$ of the matrix $\mathbf{Y}$ is defined as:
\begin{equation}
 y_{n,c} = 
\begin{cases}
1, & \text{if instance } n \text{ is labeled with class } c\\
0, & \text{otherwise}
\end{cases}   
\end{equation}
Similarly, let $\mathbf{\hat{Y}} \in [0,1]^{N \times C}$ be the matrix of predicted confidence scores, where each element $\hat{y}_{n,c}$ corresponds to the predicted confidence score that instance $n$ belongs to class $c$.
For each class $c$, we define a vector $\mathbf{Y}_{:,c}$ representing the true labels for class $c$ across all instances, and a vector $\mathbf{\hat{Y}}_{:,c}$ representing the predicted confidence scores for class $c$ across all instances.
Analogously, for the $n$-th instance, $\mathbf{Y}_{n,:}$ denotes the true label vector over all classes, while $\mathbf{\hat{Y}}_{n,:}$ is the predicted confidence score vector over all classes.
To train our model, we use a special multi-component loss function that extends the previous work of \cite{chen2019looks} and \cite{donnelly2022deformable} to the multi-label setting. 
It replaces the traditional cross-entropy loss component with an asymmetric loss function $\mathcal{L}_{\text{asym}} \left( \mathbf{\hat{Y}}, \mathbf{Y} \right)$, details of which can be found in \cite{ridnik2021asymmetric}. 
The asymmetric loss component penalizes misclassifications in the training data to increase the accuracy of the model. 
The overall loss function also includes an orthogonality loss, borrowed from \cite{donnelly2022deformable}, which supports learning a diverse set of prototypes and thus avoiding too similar prototypes within a class:
\begin{equation}
    \mathcal{L}_{\text{ortho}} \left( \mathbf{P} \right) = \frac{1}{C \cdot J^2} \sum_{c=1}^C \left\| \mathbf{\tilde{P}}^{(c)} \mathbf{\tilde{P}}^{(c)T} - \mathbf{I} \right\|_F^2,
\end{equation}
Here $\tilde{\mathbf{P}}$ denotes the prototypes normalized to unit length, and $\mathbf{I}$ is the identity matrix subtracted to exclude self-orthogonality.
In contrast to \cite{donnelly2022deformable}, in this study we normalize the orthogonality loss by dividing the Frobenius norm by the total number of pairwise orthogonalities given by $C \cdot J^2$. 
This ensures that the loss is scale invariant and not affected by the number of classes and prototypes. \\
The individual components of the loss function are weighted and aggregated into an overall loss function: 
\begin{equation}
    \mathcal{L} \left( \mathbf{P}, \mathbf{\hat{Y}}, \mathbf{Y} \right) = \mathcal{L}_{\text{asym}} \left( \mathbf{\hat{Y}}, \mathbf{Y} \right) + \lambda_1 \mathcal{L}_{\text{ortho}} \left( \mathbf{P} \right)
\end{equation}

Training is performed in two phases.
In the first phase, the backbone model $f_b$ is trained by combining it with a traditional fully connected black-box classifier in a supervised training setting using only the asymmetric loss function $\mathcal{L}_{\text{asym}} \left( \mathbf{\hat{Y}}, \mathbf{Y} \right)$.
In the second phase of training, the fully-connected classification layer is removed and replaced by the prototype learning classifier, consisting of the prototype layer $f_p$ and the final layer $f_f$.
Then the prototype layer and the final layer are fine-tuned on the embeddings of the backbone model. 
However, the weights of the backbone model itself are frozen.
In this way, the prototype learning classifier can be used either to create an inherently interpretable classification model or to generate post-hoc explanations for the learned embeddings of the backbone model.
The latter is especially useful when the black-box model outperforms the inherently interpretable model.

\subsection{Prototype visualization}
\label{subsec:prototype_visualization}
In order to use the learned prototypes to explain a model, they must be transformed into a form that is easily understood by humans. 
Since the prototypes are learned in the embedding space, they must first be projected from the embedding space back into the input space to represent them as spectrograms.
Following \cite{nauta2023pip}, in our approach we visualize each prototype in the input space by identifying the top $K$ most similar instances from the training dataset, plotting them as spectrograms, and highlighting the regions of these spectrograms that are most similar to the prototype, as shown in Figures \ref{fig:prototypes_train_mouchi} and \ref{fig:prototypes_train_yerwar}. 
By looking at the top $K$ most similar instances rather than a single instance, we gain a deeper understanding of how a prototype is associated with different sound patterns.
To identify the top $K$ most similar instances for each prototype $\mathbf{p}^{(c,j)}$, we compute the similarity $s^{(c,j)}(\mathbf{z}_n)$ between the prototype and the embedding $\mathbf{z}_n = f_b(\mathbf{x}_n)$ of each instance $\mathbf{x}_n$ of the training dataset. 
Then, for each prototype $\mathbf{p}^{(c,j)}$, we select the set $\mathcal{X}^{(c,j)}$ of $K$ instances with the highest similarity values:
\begin{equation} 
\mathcal{X}^{(c,j)} = \underset{\mathbf{x}_n}{\operatorname*{arg\,top-K}} \quad s^{(c,j)}( \mathbf{z}_n ). 
\end{equation}
This approach provides the instances that are most similar to the prototype and thus the most relevant examples for visualization.
To determine which parts of the spectrograms of the most similar instances resemble a prototype, an exact projection of the prototype from the embedding space into the input space is required. 
For this purpose, the similarities between the individual embeddings of the spectrogram patches and the prototype are computed for the spectrograms of each of the top $K$ most similar instances. 
By computing the similarities between the prototype and the spectrogram embeddings, we obtain an activation map for each spectrogram. 
Each pixel of an activation map for a spectrogram indicates the similarity value of the prototype to the embedding of one of the spectrogram patches.
Due to the translation equivariance of the underlying CNN backbone, the spatial relationships in these activation maps are preserved, allowing them to be scaled to the dimensions of the original spectrograms. 
This allows the creation of a heatmap for each spectrogram, highlighting the areas of the spectrogram that are most similar to the prototype. 
Following the methodology of \cite{chen2019looks}, the visual representation of a prototype is done by identifying the smallest rectangular spectrogram patch that contains the pixels whose activation values in the scaled activation map exceed the 95\% percentile of all activation values on the map.

\section{Experimental setup}
\label{sec:experiments}
\subsection{Data}
\label{subsec:data}
As contribution (C2) to our study, we demonstrate the scalability of \gls*{audioprotopnet} by successfully training it on the extensive XCL training dataset from BirdSet \cite{rauch2024birdset}. 
The XCL training dataset contains audio recordings of 9,734 bird species, totaling over 6,800 hours of recordings at a sampling rate of 32 kHz. 
The XCL training dataset is curated from publicly available bird sound recordings on Xeno-Canto \cite{vellinga2015xeno}, excluding all recordings under a non-derivative license. 
We used the XCL training dataset to train all of our bird sound classification models.
To balance the class distribution and reduce the computational effort, we limited the number of recordings per class in the XCL dataset to a maximum of 500 for training. 
As in \cite{rauch2024birdset}, we used the Powdermill Nature (POW) soundscape dataset \cite{powdermill_chronister_2021_4656848} as a validation dataset. 
We evaluated the models using the following seven soundscape test datasets from the BirdSet framework: High Sierra Nevada (HSN) \cite{high_sieras_mary_clapp_2023_7525805}, NIPS4Bplus (NBP) \cite{morfi2019nips4bplus}, Colombia Costa Rica (NES) \cite{columbia_alvaro_vega_hidalgo_2023_7525349}, Amazon Basin (PER) \cite{amazon_basin_w_alexander_hopping_2022_7079124}, Sierra Nevada (SNE) \cite{sierra_nevada_stefan_kahl_2022_7050014}, Sapsucker Woods (SSW) \cite{sapsucker_stefan_kahl_2022_7079380}, and Hawaiian Islands (UHH) \cite{hawaii_amanda_navine_2022_7078499}. 
A more detailed description of these datasets can be found in \cite{rauch2024birdset}. \\
As in BirdSet \cite{rauch2024birdset}, we used the \textit{bambird} event detection package \cite{michaud2023unsupervised} to extract the bird vocalizations from the Xeno-Canto training data, resulting in a total of 1,528,068 training instances.  
This generated exactly one 5-second audio clip per recording. 
For the validation and test datasets, each recording was segmented into non-overlapping 5-second intervals. 
We then converted the audio data into log-mel spectrograms with 256 mel bins, using an FFT size of 2048, a hop length of 256, and 1025 bins in the STFT. 
In addition, we applied z-score standardization\footnote{The z-score is a method of normalizing data by transforming the features to have a mean of 0 and a standard deviation of 1. 
The z-score $z$ of an individual instance $x$ is calculated by subtracting the mean $\mu$ of the training dataset from the instance and dividing the result by the standard deviation $\sigma$ of the training dataset, i.e. $z=\left( x - \mu \right) / \sigma$.}, based on a mean of $-13.369$ and a standard deviation of $13.162$ calculated from the training dataset. \\
We used several data augmentation techniques to improve the robustness and generalizability of the models. 
We applied waveform augmentations before converting to spectrograms. 
A time shift with probability $p = 1.0$ randomly shifted the temporal position of the audio signal within an 8 second window from the original recording around the detected event to increase the adaptability of the model to temporal variations. 
A background noise mixing with probability $p = 0.5$ added background noise without bird vocalizations from the BirdVox-DCASE-20k dataset \cite{lostanlen_2018_dcase} of the DCASE18 Bird Audio Detection Challenge \cite{stowell2019automatic} to help the model distinguish signal from noise. 
Random colored noise was also added with probability $p = 0.2$. 
A gain adjustment with probability $p = 0.2$ randomly varied the volume to increase the robustness of the model to amplitude variations. 
A multi-label mixup \cite{zhang2017mixup, hendrycks2019augmix} with probability $p = 0.8$ mixed up to three random instances from a batch and combined their labels using hard labels to increase data diversity and simulate multi-label conditions. 
A no-call mixing with probability $p = 0.075$ replaced some instances with no-call instances from the BirdVox-DCASE-20k dataset \cite{lostanlen_2018_dcase}, labeled with a zero vector, to simulate the absence of bird vocalizations in the recordings.
In addition, we applied spectrogram augmentation techniques after conversion to spectrograms. 
Frequency masking \cite{park2019specaugment} with a probability of $p = 0.5$ randomly masked a range of frequency bands in the spectrogram to increase robustness to spectral variations. 
Time masking \cite{park2019specaugment} with a probability of $p = 0.3$ randomly masked a time segment in the spectrogram to increase robustness to temporal discontinuities. 
More detailed information about these augmentation techniques can be found in \cite{rauch2024birdset}.

\subsection{Classification models}
\label{subsec:classification_models}
In this study, we select ConvNeXt-B \cite{liu2022convnet} as a black-box model for bird sound classification because it can achieve performance comparable to other state-of-the-art models \cite{rauch2024birdset}. 
We further trained a ConvNeXt-B model, pre-trained on ImageNet \cite{deng2009imagenet}, on the XCL dataset using the AdamW optimizer \cite{loshchilov2017decoupled} with a weight decay of $1 \times 10^{-4}$ and a learning rate of $5 \times 10^{-4}$. 
We used a learning rate that followed a cosine annealing learning rate schedule \cite{loshchilov2016sgdr} with a warm-up ratio of 0.05. 
The ConvNeXt black-box model had a total of 97.5 million parameters.
We trained the model for 10 epochs using five different random seeds and a batch size of 64. 
For each random seed, we selected the model checkpoint with the lowest validation loss for evaluation. 
To create the inherently interpretable \gls*{audioprotopnet} model, we modified the black-box ConvNeXt model after training by removing the fully connected classifier and replacing it with the prototype learning classifier described in Section \ref{sec:methodology}.
We froze the ConvNeXt backbone of the \gls*{audioprotopnet} model to preserve the learned embeddings while optimizing the prototype learning classifier over 10 epochs. 
We created a total of four different \gls*{audioprotopnet} models, each with a different number of prototypes per class, namely one, five, ten, and twenty prototypes per class.
These four models will be referred to as \gls*{audioprotopnet}-1, \gls*{audioprotopnet}-5, \gls*{audioprotopnet}-10 and \gls*{audioprotopnet}-20 respectively.
We trained the \gls*{audioprotopnet} models using the AdamW optimizer with a weight decay of $1 \times 10^{-4}$. 
We set the learning rate to $0.05$ for the prototype vectors and $5 \times 10^{-4}$ for the final layer. 
We applied a cosine annealing learning rate schedule with a warmup ratio of 0.05 to both learning rates. 
We weighted the asymmetric loss and the orthogonality loss equally, i.e. $\lambda _1 = 1.0$. 
Analogous to the training of the black-box ConvNeXt model, we trained each of the \gls*{audioprotopnet} models with five different random seeds. 
\gls*{audioprotopnet}-1 had a total of 97.6 million parameters, \gls*{audioprotopnet}-5 had 137.5 million parameters, \gls*{audioprotopnet}-10 had 187.4 million parameters, and \gls*{audioprotopnet}-20 had 287.2 million parameters, including the backbone parameters.
While a batch size of 64 was used for \gls*{audioprotopnet}-1, \gls*{audioprotopnet}-5 and \gls*{audioprotopnet}-10, a batch size of 32 was used for \gls*{audioprotopnet}-20 due to its large number of parameters.
For each random seed, we selected the model checkpoint with the lowest validation loss for evaluation. \\
In addition, we used the Perch model \cite{hamer2023birb}, which represents the state-of-the-art in bird sound classification, as another baseline. 
Perch is based on the EfficientNet B1 architecture \cite{tan2019efficientnet} and has about 80 million parameters.
It has proven to be superior to competing models such as BirdNet \cite{kahl2021birdnet}. 
This superiority was confirmed in a comprehensive analysis by \cite{ghani2023global}, which showed that Perch was slightly more accurate and consistent in its results than BirdNet and similar models across a wide range of datasets.

\subsection{Evaluation metrics}
\label{subsec:evaluation_metrics}
As in \cite{rauch2024birdset}, we masked the logits of the models during testing so that only the corresponding logits of the models for the target classes of the respective test dataset were included in the evaluation. 
Following the BirdSet evaluation protocol \cite{rauch2024birdset}, the following metrics were used to quantify model performance:
\begin{itemize}
    \item \textbf{AUROC}: 
    This threshold-free metric quantifies the ability of the model to discriminate between classes across all thresholds by measuring the area under the receiver operating characteristic curve. It is equivalent to the probability that a random pair of positive and negative examples is ranked correctly and is defined as follows \cite{hamer2023birb}:
    \begin{equation}
        \text{AUROC} \left( \mathbf{\hat{Y}}, \mathbf{Y} \right) = \frac{1}{C} \sum_{c=1}^{C} \left( \frac{1}{|\mathbf{Y}_{+,c}| \cdot |\mathbf{Y}_{-,c}|} \sum_{n \in \mathbf{Y}_{+,c}} \sum_{m \in \mathbf{Y}_{-,c}} \chi _{\left\{ \hat{y}_{n,c} < \hat{y}_{m,c} \right\}} \right),
    \end{equation}
    where $\mathbf{Y}_{+,c}$ is the set of indices where the true label for class $c$ is positive, and $\mathbf{Y}_{-,c}$ is the set of indices where the true label for class $c$ is negative.
    The indicator function $\chi$ is 1 if the predicted score for class $c$ is higher for a positive instance than for a negative instance, and 0 otherwise. 
    \gls*{auroc} provides a balanced assessment of model performance because it generates scores that can be compared across datasets and classes that have different numbers of positive and negative instances \cite{van2024birds}. \\
        
    \item \textbf{cmAP}: Determines the average precision of each class and then averages across all classes. 
    This approach provides detailed insight into the model's ability to consistently classify positive instances higher than negative instances across all possible thresholds \cite{rauch2024birdset}. 
    The \gls*{cmap} score is defined as follows:
    \begin{equation}
        \text{cmAP} \left( \mathbf{\hat{Y}}, \mathbf{Y} \right) = \frac{1}{C} \sum_{c=1}^{C} \text{AP}(\mathbf{\hat{Y}}_{:,c}, \mathbf{Y}_{:,c}),
    \end{equation}
    where $\text{AP}(\mathbf{\hat{Y}}_{:,c}, \mathbf{Y}_{:,c})$ is the average precision for class $c$. 
    An advantage of \gls*{cmap} is that all classes are weighted equally, regardless of the number of instances in the dataset \cite{kahl2023overview}. \\ 

    \item \textbf{Top-1 accuracy}: This metric evaluates whether the class with the highest predicted confidence score is among the correct classes for each instance \cite{rauch2024birdset}:
    \begin{equation}
        \text{T1-Acc} \left( \mathbf{\hat{Y}} , \mathbf{Y} \right) = \frac{1}{N} \sum_{n=1}^{N} \chi_{\left\{ \hat{y}_n^{(\text{top})} \in \mathbf{Y}_{n,:} \right\}},
    \end{equation}
    where $\mathbf{Y}_{n,:}$ represents the set of correct classes for the $n^{th}$ instance, $\hat{y}_n^{(\text{top})}$ is the class with the highest predicted confidence score for the $n^{th}$ instance, and $\chi$ is the indicator function, which is 1 if $\hat{y}_n^{(\text{top})}$ is within the set of correct classes $\mathbf{Y}_{n,:}$ and 0 otherwise.
    Although less meaningful for instances with multiple labels, it provides a quick assessment of the model's ability to correctly predict at least one relevant class for each instance. 
\end{itemize}

\section{Results}
\label{sec:results}
\subsection{Performance evaluation} 
\subsubsection{Comparison of AudioProtoPNet architectures}
As part of contribution (C4), we investigated the influence of the number of prototypes per class on the performance of \gls*{audioprotopnet}.
As shown in Table \ref{tab:results_audioprotopnet_xcl_table}, the \gls*{audioprotopnet} models with five, ten and twenty prototypes per class showed similar performance in terms of \gls*{cmap}, \gls*{auroc} and top-1 accuracy. 
Although the model with twenty prototypes per class, i.e. \gls*{audioprotopnet}-20, often achieved the best results on many test datasets, the improvements over the \gls*{audioprotopnet} models with five and ten prototypes are marginal. 
This is illustrated by the average metric values over all seven test datasets shown in the Score column of Table \ref{tab:results_audioprotopnet_xcl_table}, where \gls*{audioprotopnet}-5, \gls*{audioprotopnet}-10, and \gls*{audioprotopnet}-20 each achieved the same \gls*{cmap} score of 0.42, the same \gls*{auroc} score of 0.90, and the same top-1 accuracy score of 0.62.

\begin{table}[ht]
    \caption{Mean performance of \gls*{audioprotopnet} models with one, five, ten and twenty prototypes per class for the validation dataset POW and the seven test datasets over five different random seeds. The score corresponds to the average of the respective metric over the test datasets. The best values for each metric are highlighted in bold, while the second best values are underlined. While the \gls*{audioprotopnet} models with five, ten, and twenty prototypes per class performed similarly, the \gls*{audioprotopnet} model with only one prototype per class performed slightly worse.}
    \centering
    \begin{tabularx}{\textwidth}{M{1cm} | c | Y | Y Y Y Y Y Y Y !{\vrule width 1pt} Y}
        \toprule
        &  & \textbf{POW} & \textbf{PER} & \textbf{NES} & \textbf{UHH} & \textbf{HSN} & \textbf{NBP} & \textbf{SSW} & \textbf{SNE} & \textbf{Score} \\
        \midrule
        \multirow{3}{*}{\rotatebox[origin=c]{90}{\shortstack{\textbf{Audio} \\ \textbf{Proto} \\ \textbf{PNet-1}}}}
        & cmAP & 0.49 & \textbf{0.30} & \underline{0.36} & 0.28 & \underline{0.50} & \underline{0.66} & 0.40 & 0.32 & \underline{0.40} \\ 
        & AUROC  & 0.88 & \underline{0.79} & 0.92 & 0.85 & \underline{0.91} & \underline{0.92} & \underline{0.96} & 0.84 & \underline{0.88} \\
        & T1-Acc & 0.87 & \underline{0.59} & \underline{0.49} & 0.42 & \underline{0.64} & \underline{0.71} & 0.64 & 0.70 & \underline{0.60} \\ 
        \midrule
        \multirow{3}{*}{\rotatebox[origin=c]{90}{\shortstack{\textbf{Audio} \\ \textbf{Proto} \\ \textbf{PNet-5}}}}
        & cmAP & 0.50 & \textbf{0.30} & \textbf{0.38} & \textbf{0.31} & \textbf{0.54} & \textbf{0.68} & \underline{0.42} & \underline{0.33} & \textbf{0.42} \\ 
        & AUROC  & 0.88 & \underline{0.79} & \underline{0.93} & \textbf{0.87} & \textbf{0.92} & \textbf{0.93} & \textbf{0.97} & \underline{0.86} & \textbf{0.90} \\
        & T1-Acc & 0.84 & \underline{0.59} & \textbf{0.52} & \textbf{0.49} & \textbf{0.65} & \underline{0.71} & 0.66 & \underline{0.74} & \textbf{0.62} \\ 
        \midrule
        \multirow{3}{*}{\rotatebox[origin=c]{90}{\shortstack{\textbf{Audio} \\ \textbf{Proto} \\ \textbf{PNet-10}}}}
        & cmAP & 0.50 & \textbf{0.30} & \textbf{0.38} & \underline{0.30} & \textbf{0.54} & \textbf{0.68} & \underline{0.42} & \textbf{0.34} & \textbf{0.42} \\ 
        & AUROC  & 0.88 & \textbf{0.80} & \textbf{0.94} & \underline{0.86} & \textbf{0.92} & \textbf{0.93} & \textbf{0.97} & \underline{0.86} & \textbf{0.90} \\
        & T1-Acc & 0.85 & \underline{0.59} & \textbf{0.52} & \underline{0.47} & \underline{0.64} & \textbf{0.72} & \underline{0.67} & \underline{0.74} & \textbf{0.62} \\ 
        \midrule
        \multirow{3}{*}{\rotatebox[origin=c]{90}{\shortstack{\textbf{Audio} \\ \textbf{Proto} \\ \textbf{PNet-20}}}}
        & cmAP & 0.50 & \textbf{0.30} & \textbf{0.38} & \textbf{0.31} & \textbf{0.54} & \textbf{0.68} & \textbf{0.43} & \underline{0.33} & \textbf{0.42} \\ 
        & AUROC  & 0.89 & \textbf{0.80} & \textbf{0.94} & \underline{0.86} & \textbf{0.92} & \textbf{0.93} & \textbf{0.97} & \textbf{0.87} & \textbf{0.90} \\
        & T1-Acc & 0.87 & \textbf{0.60} & \textbf{0.52} & 0.42 & \textbf{0.65} & \textbf{0.72} & \textbf{0.68} & \textbf{0.75} & \textbf{0.62} \\ 
        \bottomrule
    \end{tabularx}
    \label{tab:results_audioprotopnet_xcl_table}
\end{table}

The model with only one prototype per class, \gls*{audioprotopnet}-1, performed slightly worse than the other three \gls*{audioprotopnet} models. 
\gls*{audioprotopnet}-1 achieved a \gls*{cmap} value of 0.40, an \gls*{auroc} value of 0.88, and a top-1 accuracy of 0.60 across all test datasets. 
This shows that increasing the number of prototypes from one to five per class slightly improved the classification performance of \gls*{audioprotopnet}.
Since \gls*{audioprotopnet}-5 achieved the same performance as the models with a higher number of prototypes, but used fewer prototypes per class, it can be concluded that five prototypes are sufficient to represent the most important features of each class in the embedding space that the ConvNeXt backbone has learned to extract.
In addition, a smaller number of prototypes improves the explainability of the model, since fewer prototypes per class generate less overhead for human analysis and can be understood more easily.
For these reasons, the following analysis focuses on the \gls*{audioprotopnet}-5 model.

\subsubsection{Comparison of AudioProtoPNet with state-of-the-art models} 
\begin{figure}[ht]
    \centering
    \includegraphics[trim={0cm 0cm 0cm 0cm}, clip, width=\textwidth]{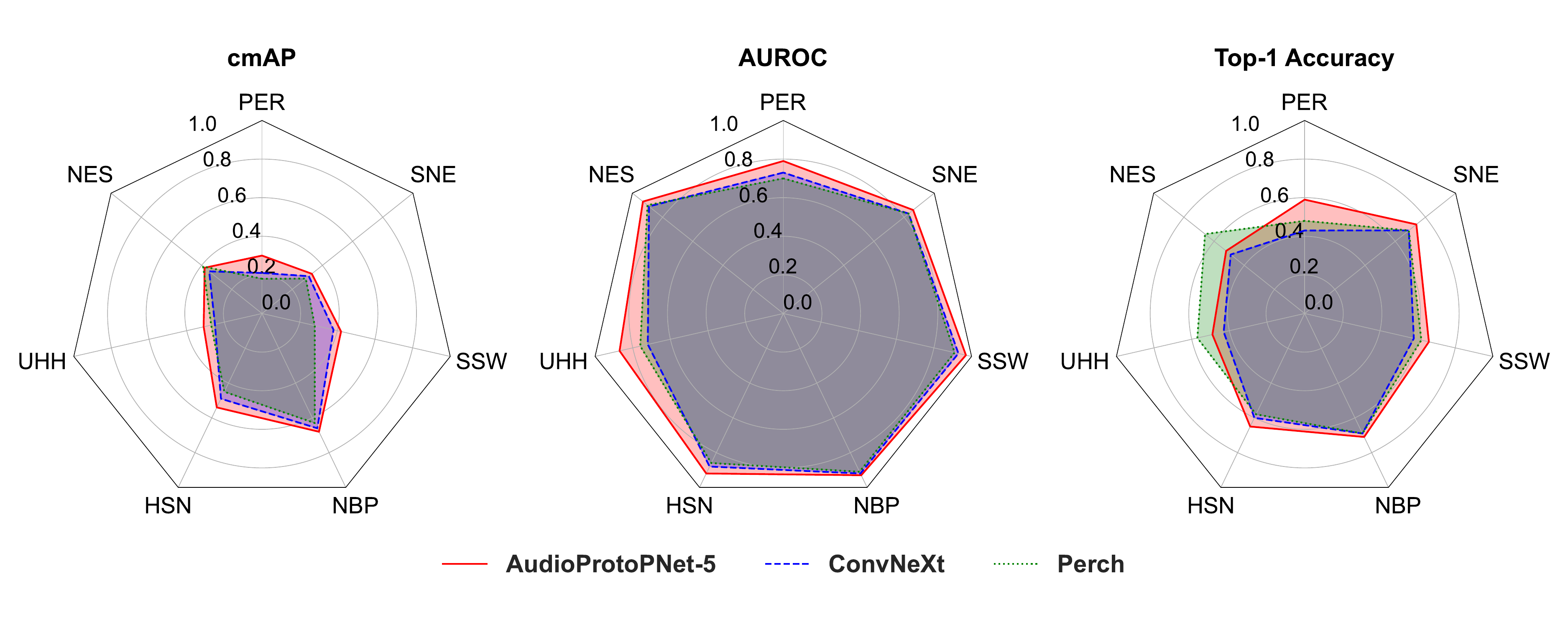}
    \caption{Average performance of \gls*{audioprotopnet}-5 (solid), ConvNeXt (dashed), and Perch (dotted) for the seven test datasets over five different random seeds, shown in a radar plot. With respect to the \gls*{auroc} metric, \gls*{audioprotopnet}-5 outperformed ConvNeXt and Perch on all test datasets. Also, \gls*{audioprotopnet}-5 achieved the best \gls*{cmap} and top-1 accuracy values for most of the test datasets. Perch only achieved a slightly better \gls*{cmap} value for the NES test dataset, and a much higher top-1 accuracy for the NES and UHH datasets.
    }
\label{fig:results_xcl_radar_charts}
\end{figure}

As contribution (C3), we not only compared the performance of \gls*{audioprotopnet}-5 with the black-box ConvNeXt model that served as its backbone, but also tested it directly against Perch \cite{hamer2023birb}, the current state-of-the-art bird sound classification model.
The results are shown in Table \ref{tab:results_baselines_xcl_table} and Figure \ref{fig:results_xcl_radar_charts}.

\begin{table}[ht]
    \caption{Mean performance of \gls*{audioprotopnet}, ConvNeXt and Perch for the validation dataset POW and the seven test datasets over five different random seeds. The score corresponds to the average of the respective metric over the test datasets. The best values for each metric are highlighted in bold, while the second best values are underlined. \gls*{audioprotopnet}-5 outperformed Perch and ConvNeXt in terms of \gls*{cmap}, \gls*{auroc}, and top-1 accuracy scores.}
    \centering
    \begin{tabularx}{\textwidth}{M{1cm} | c | Y | Y Y Y Y Y Y Y !{\vrule width 1pt} Y}
        \toprule
        &  & \textbf{POW} & \textbf{PER} & \textbf{NES} & \textbf{UHH} & \textbf{HSN} & \textbf{NBP} & \textbf{SSW} & \textbf{SNE} & \textbf{Score} \\ 
        \midrule
        \multirow{3}{*}{\rotatebox[origin=c]{90}{\shortstack{\textbf{Audio} \\ \textbf{Proto} \\ \textbf{PNet-5}}}}
        & cmAP & 0.50 & \textbf{0.30} & \underline{0.38} & \textbf{0.31} & \textbf{0.54} & \textbf{0.68} & \textbf{0.42} & \textbf{0.33} & \textbf{0.42} \\ 
        & AUROC  & 0.88 & \textbf{0.79} & \textbf{0.93} & \textbf{0.87} & \textbf{0.92} & \textbf{0.93} & \textbf{0.97} & \textbf{0.86} & \textbf{0.90} \\
        & T1-Acc & 0.84 & \textbf{0.59} & \underline{0.52} & \underline{0.49} & \textbf{0.65} & \textbf{0.71} & \textbf{0.66} & \textbf{0.74} & \textbf{0.62} \\ 
        \midrule
        \multirow{3}{*}{\rotatebox[origin=c]{90}{\shortstack{\textbf{Conv} \\ \textbf{NeXt}}}}
        & cmAP   & 0.41 & \underline{0.21} & 0.35 & 0.25 & \underline{0.49} & \underline{0.66} & \underline{0.38} & \underline{0.31} & \underline{0.38} \\     
        & AUROC  & 0.83 & \underline{0.73} & 0.89 & 0.72 & \underline{0.88} & \underline{0.92} & \underline{0.93} & \underline{0.83} & \underline{0.84} \\
        & T1-Acc & 0.75 & 0.43 & 0.49 & 0.43 & \underline{0.60} & \underline{0.69} & 0.58 & \underline{0.69} & 0.56 \\ 
        \midrule
        \multirow{3}{*}{\rotatebox[origin=c]{90}{\textbf{Perch}}}
        & {cmAP} & 0.30 & 0.18 & \textbf{0.39} & \underline{0.27} & 0.45 & 0.63 & 0.28 & 0.29 & 0.36 \\
        & {AUROC} & 0.84 & 0.70 & \underline{0.90} & 0.76 & 0.86 & 0.91 & 0.91 & \underline{0.83} & \underline{0.84} \\ 
        & {T1-Acc} & 0.85 & \underline{0.48} & \textbf{0.66} & \textbf{0.57} & 0.58 & \underline{0.69} & \underline{0.62} & \underline{0.69} & \underline{0.61} \\
        \bottomrule
    \end{tabularx}
    \label{tab:results_baselines_xcl_table}
\end{table}

Overall, our experiments show that \gls*{audioprotopnet}-5 outperformed the black-box models ConvNeXt and Perch on most test datasets. 
In particular, \gls*{audioprotopnet}-5 achieved the highest \gls*{auroc} values for all test datasets and higher \gls*{cmap} and top-1 accuracy values for most test datasets. 
\gls*{audioprotopnet}-5 achieved an average \gls*{auroc} score of 0.90 across all test datasets, while ConvNeXt and Perch only achieved \gls*{auroc} scores of 0.84 each.
It is worth noting that \gls{audioprotopnet}-5 thus outperformed all models in the BirdSet multi-label bird sound classification benchmark \cite{rauch2024birdset}.
A similar pattern can be seen for the \gls*{cmap} metric, where \gls*{audioprotopnet}-5 also outperformed Perch and ConvNeXt on almost all test datasets.
\gls*{audioprotopnet}-5 achieved an average \gls*{cmap} score of 0.42 across all test datasets, while ConvNeXt and Perch only achieved \gls*{cmap} scores of 0.38 and 0.36, respectively.
In terms of top-1 accuracy, \gls*{audioprotopnet}-5 also achieved the highest score with 0.62, followed by Perch with 0.61 and ConvNeXt with 0.56. 
Although \gls*{audioprotopnet}-5 achieved better top-1 accuracy values for most test datasets, Perch clearly outperformed the top-1 accuracy of \gls*{audioprotopnet}-5 and ConvNeXt for the two test datasets NES and UHH.  
However, the top-1 accuracy is less meaningful in the context of multi-label classification tasks than \gls*{cmap} or \gls*{auroc}, since it considers only the confidence scores of the highest ranked class and ignores the confidence scores of the other classes. \\
Another notable result is the superior performance of \gls*{audioprotopnet}-5 on the PER test dataset. 
This dataset is the most challenging of the seven test datasets due to the high number of different bird species per instance and the diverse soundscape of the Amazon rainforest, which is reflected in the performance of the models \cite{rauch2024birdset}.
However, the much better performance of \gls*{audioprotopnet}-5 with a \gls*{cmap} of 0.30 and an \gls*{auroc} of 0.79 on this dataset indicates an improved generalization ability in difficult and diverse acoustic environments compared to ConvNeXt and Perch, which only achieved \gls*{cmap} values of 0.21 and 0.18 and \gls*{auroc} values of 0.73 and 0.70, respectively.
In summary, our results show that \gls*{audioprotopnet}-5 often outperformed the black-box models Perch and ConvNeXt.

\subsection{Qualitative analysis of learned prototypes}
\begin{figure}[ht]
\centering
\begin{minipage}{\textwidth}
\begin{minipage}{0.025\textwidth}
\rotatebox{90}{\scriptsize\textbf{Prototype 1}}
\end{minipage}
\begin{minipage}{0.025\textwidth}
\rotatebox{90}{\scriptsize $w^{(j,c)} = 2.52$}
\end{minipage}
\begin{minipage}{0.95\textwidth}
\centering
\begin{minipage}{0.19\textwidth}
\centering
\includegraphics[trim={57 35 40 20},clip, width=\textwidth]{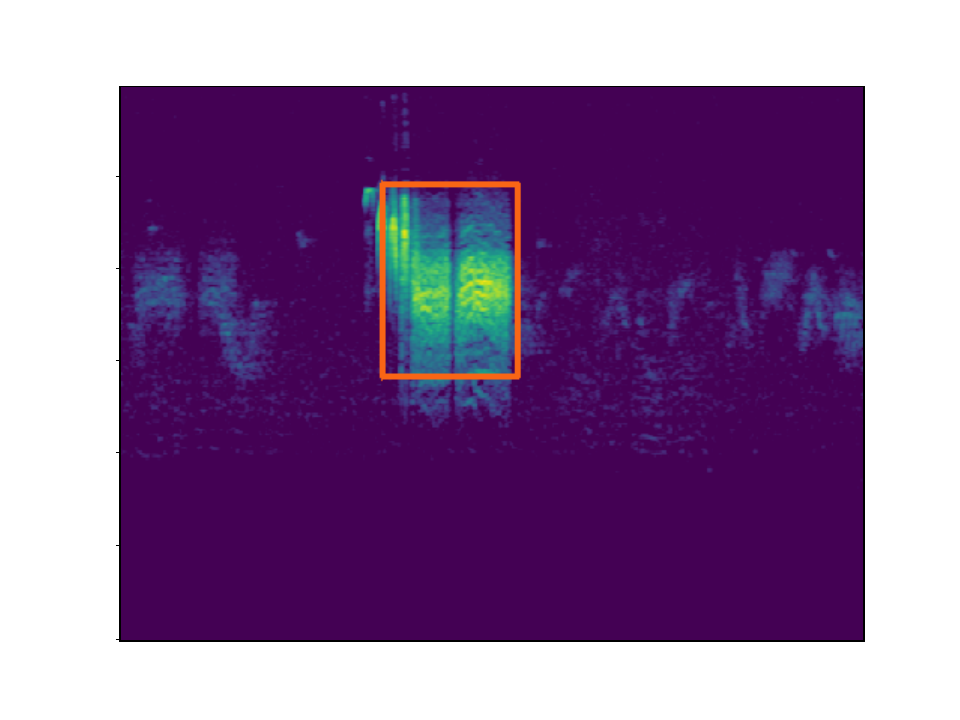}
\par
{\scriptsize $s^{(c,j)} = 0.50$}
\par
\scriptsize\textbf{mouchi}
\end{minipage}
\begin{minipage}{0.19\textwidth}
\centering
\includegraphics[trim={57 35 40 20},clip, width=\textwidth]{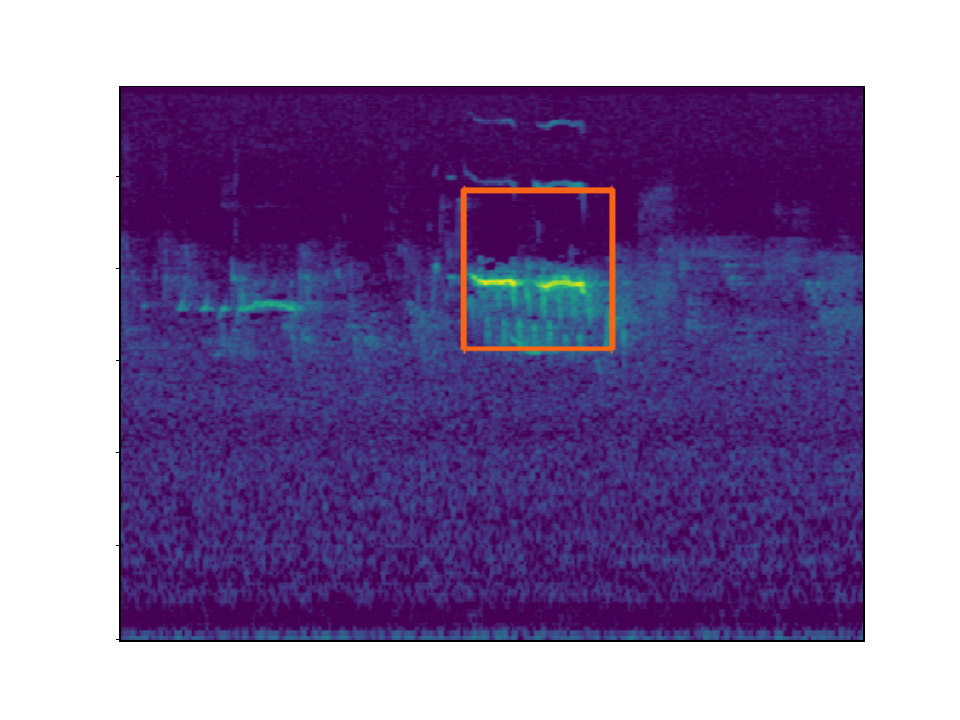}
\par
{\scriptsize $s^{(c,j)} = 0.49$}
\par
\scriptsize\textbf{mouchi}
\end{minipage}
\begin{minipage}{0.19\textwidth}
\centering
\includegraphics[trim={57 35 40 20},clip, width=\textwidth]{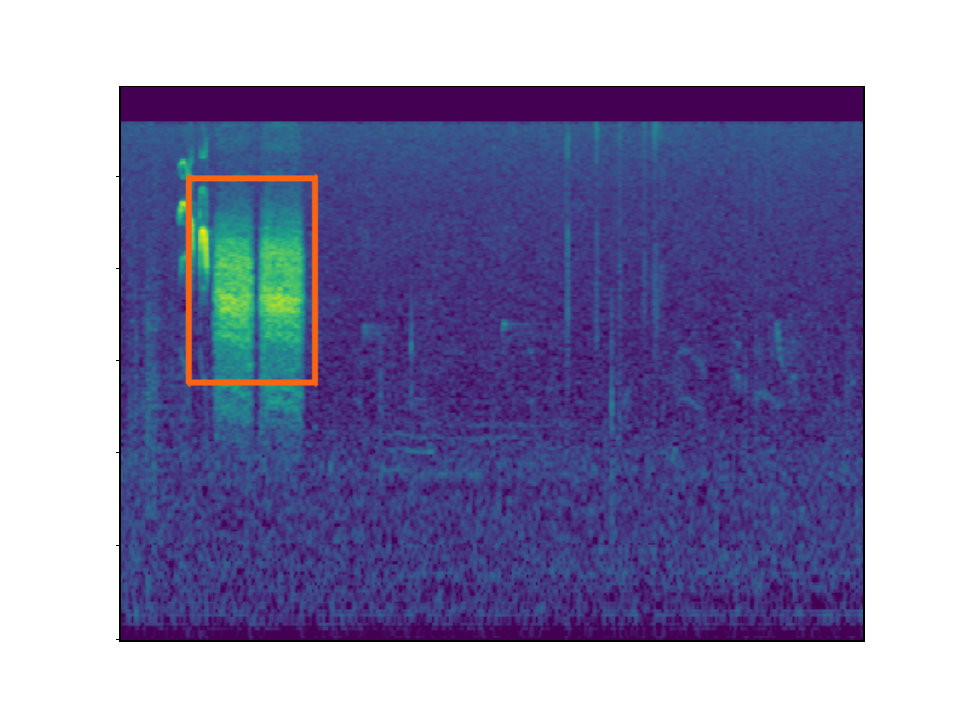}
\par
{\scriptsize $s^{(c,j)} = 0.49$}
\par
\scriptsize\textbf{mouchi}
\end{minipage}
\begin{minipage}{0.19\textwidth}
\centering
\includegraphics[trim={57 35 40 20},clip, width=\textwidth]{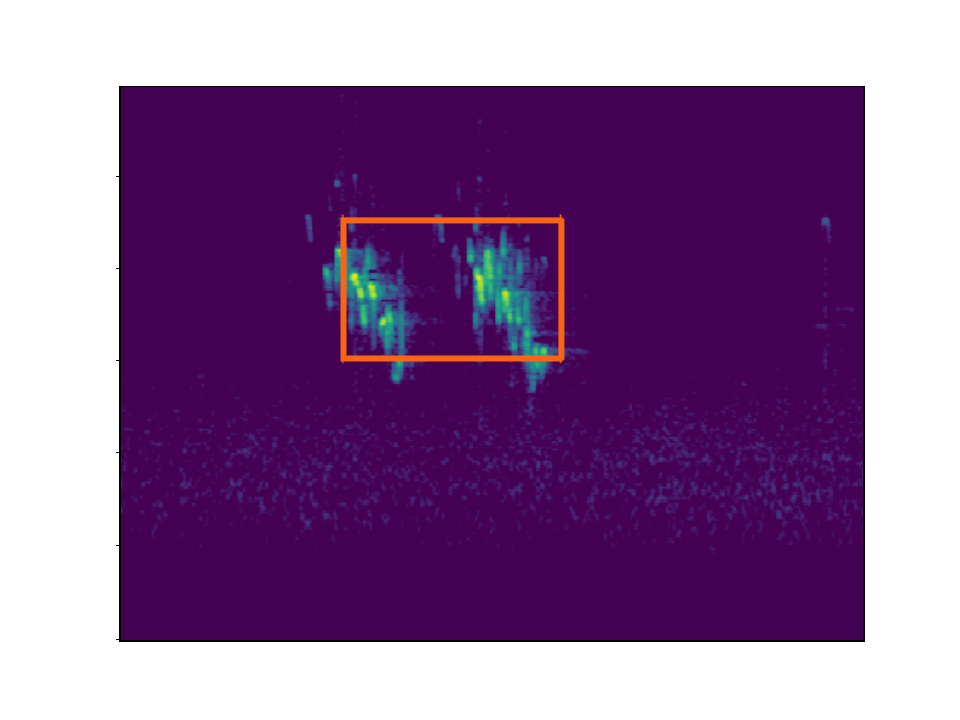}
\par
{\scriptsize $s^{(c,j)} = 0.49$}
\par
\scriptsize\textbf{mouchi}
\end{minipage}
\begin{minipage}{0.19\textwidth}
\centering
\includegraphics[trim={57 35 40 20},clip, width=\textwidth]{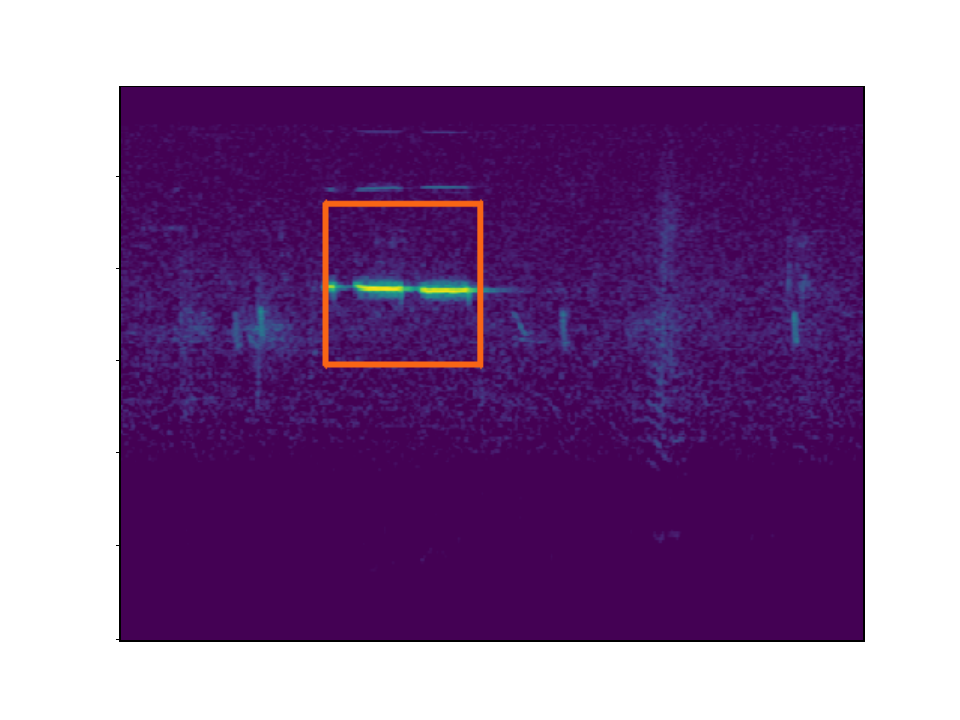}
\par
{\scriptsize $s^{(c,j)} = 0.49$}
\par
\scriptsize\textbf{mouchi}
\end{minipage}
\end{minipage}
\end{minipage}
\begin{minipage}{\textwidth}
\begin{minipage}{0.025\textwidth}
\rotatebox{90}{\scriptsize\textbf{Prototype 2}}
\end{minipage}
\begin{minipage}{0.025\textwidth}
\rotatebox{90}{\scriptsize $w^{(j,c)} = 1.03$}
\end{minipage}
\begin{minipage}{0.95\textwidth}
\centering
\begin{minipage}{0.19\textwidth}
\centering
\includegraphics[trim={57 35 40 20},clip, width=\textwidth]{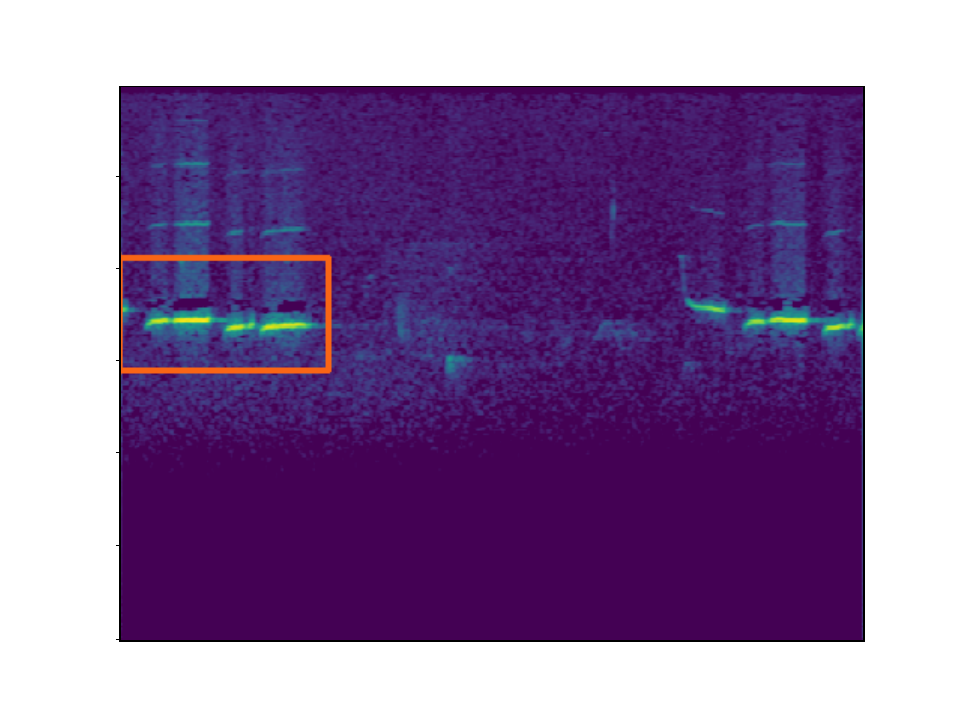}
\par
{\scriptsize $s^{(c,j)} = 0.40$}
\par
\scriptsize\textbf{mouchi}
\end{minipage}
\begin{minipage}{0.19\textwidth}
\centering
\includegraphics[trim={57 35 40 20},clip, width=\textwidth]{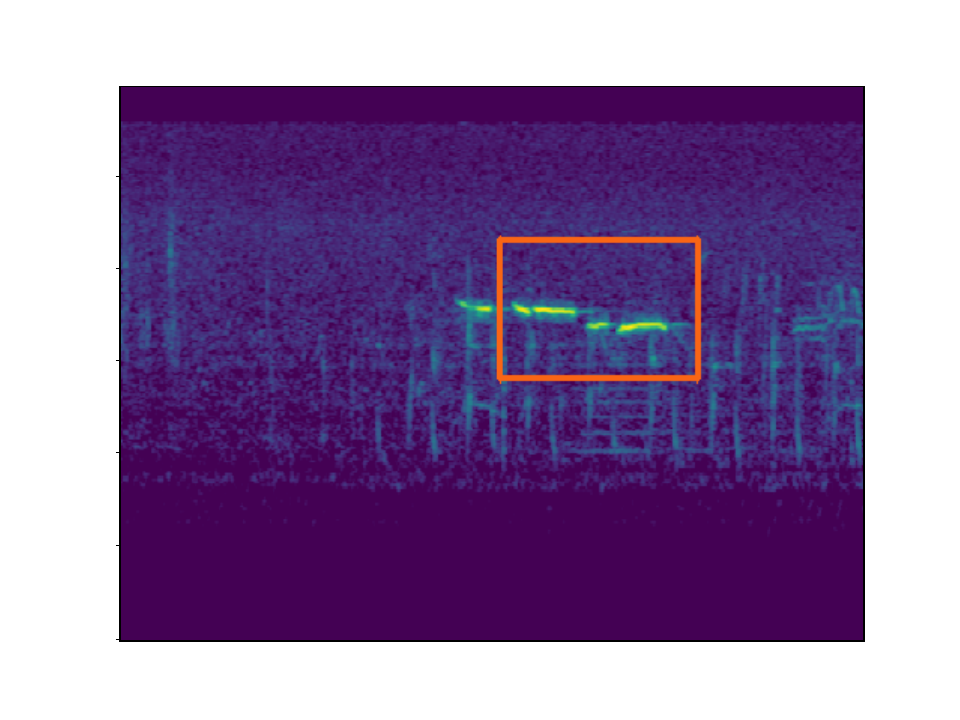}
\par
{\scriptsize $s^{(c,j)} = 0.40$}
\par
\scriptsize\textbf{mouchi}
\end{minipage}
\begin{minipage}{0.19\textwidth}
\centering
\includegraphics[trim={57 35 40 20},clip, width=\textwidth]{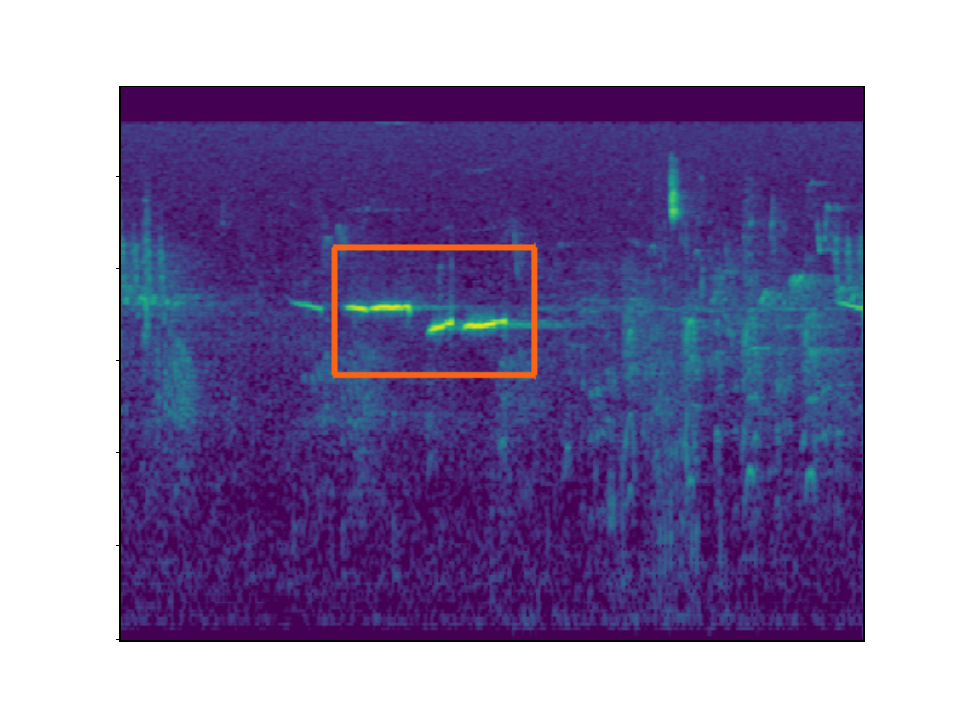}
\par
{\scriptsize $s^{(c,j)} = 0.39$}
\par
\scriptsize\textbf{mouchi}
\end{minipage}
\begin{minipage}{0.19\textwidth}
\centering
\includegraphics[trim={57 35 40 20},clip, width=\textwidth]{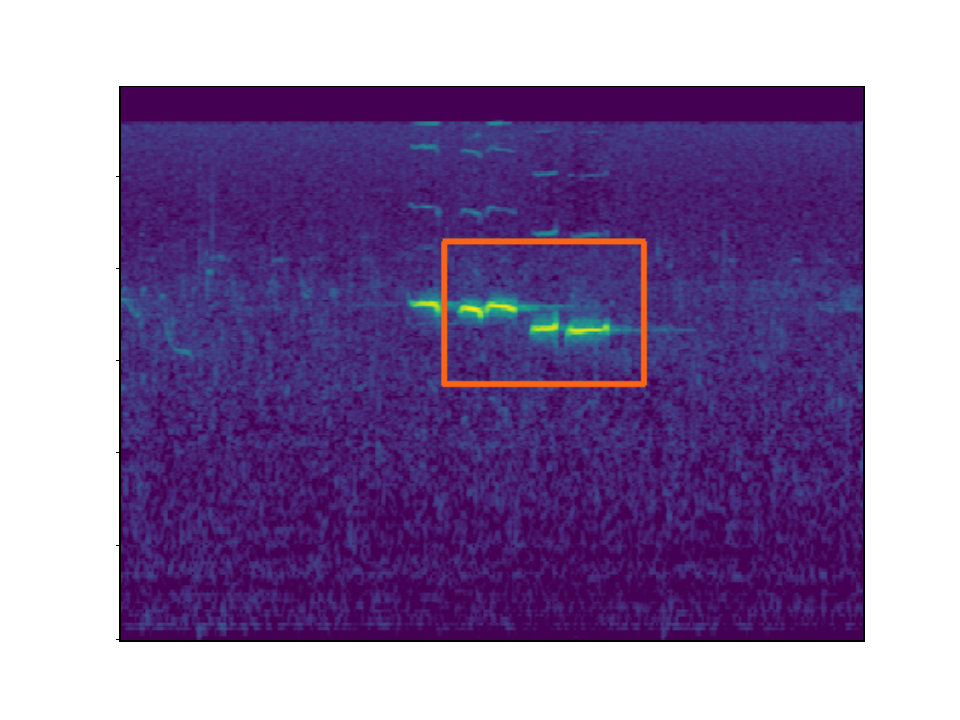}
\par
{\scriptsize $s^{(c,j)} = 0.38$}
\par
\scriptsize\textbf{mouchi}
\end{minipage}
\begin{minipage}{0.19\textwidth}
\centering
\includegraphics[trim={57 35 40 20},clip, width=\textwidth]{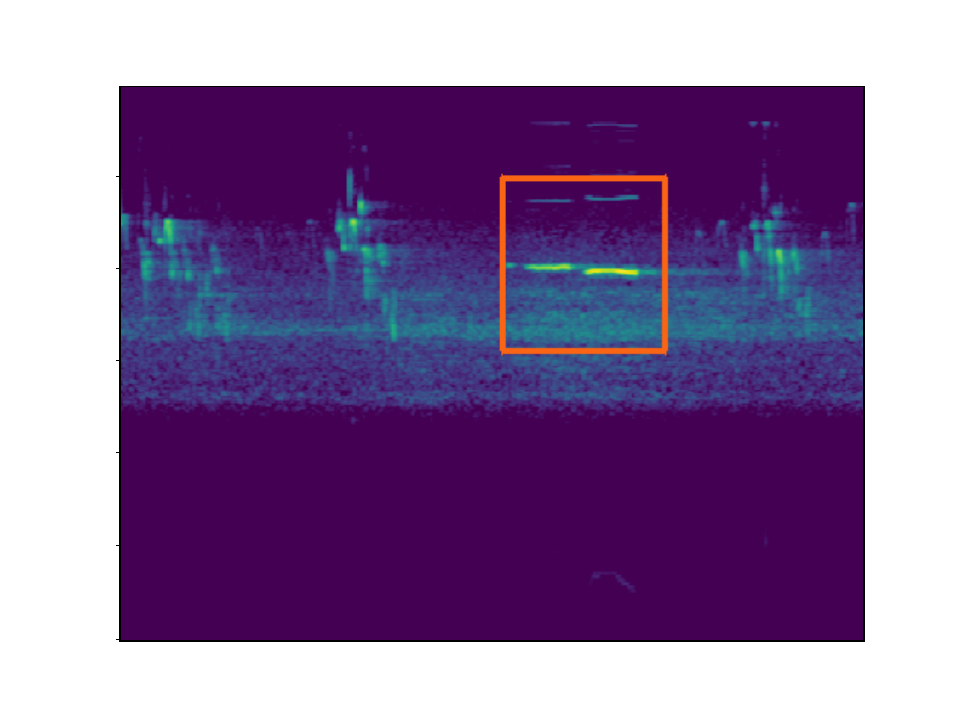}
\par
{\scriptsize $s^{(c,j)} = 0.38$}
\par
\scriptsize\textbf{mouchi}
\end{minipage}
\end{minipage}
\end{minipage}
\begin{minipage}{\textwidth}
\begin{minipage}{0.025\textwidth}
\rotatebox{90}{\scriptsize\textbf{Prototype 3}}
\end{minipage}
\begin{minipage}{0.025\textwidth}
\rotatebox{90}{\scriptsize $w^{(j,c)} = 2.49$}
\end{minipage}
\begin{minipage}{0.95\textwidth}
\centering
\begin{minipage}{0.19\textwidth}
\centering
\includegraphics[trim={57 35 40 20},clip, width=\textwidth]{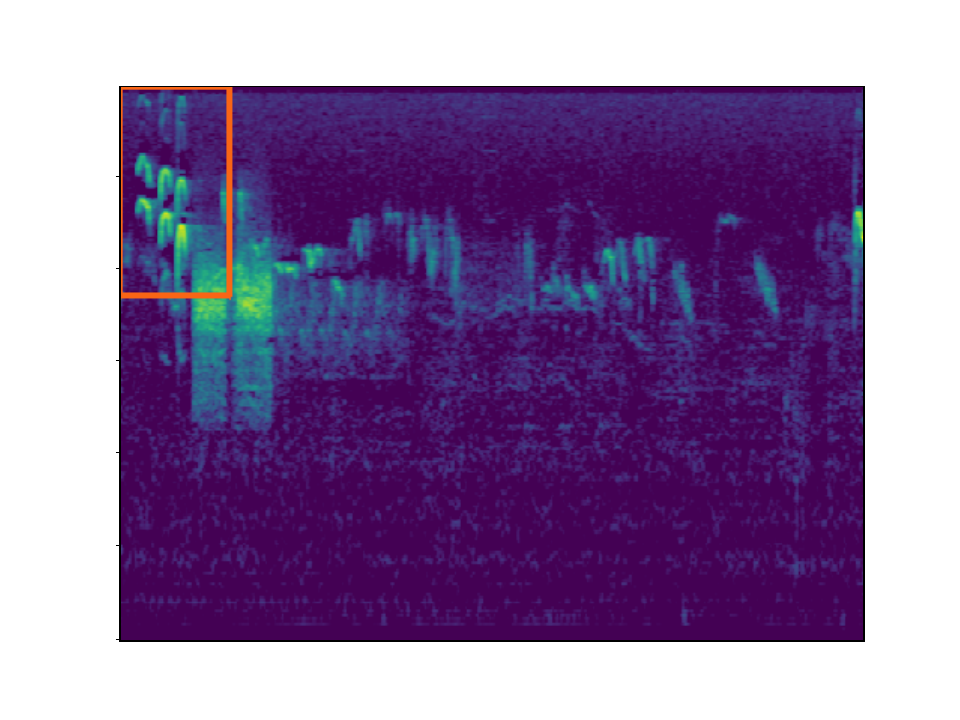}
\par
{\scriptsize $s^{(c,j)} = 0.57$}
\par
\scriptsize\textbf{mouchi}
\end{minipage}
\begin{minipage}{0.19\textwidth}
\centering
\includegraphics[trim={57 35 40 20},clip, width=\textwidth]{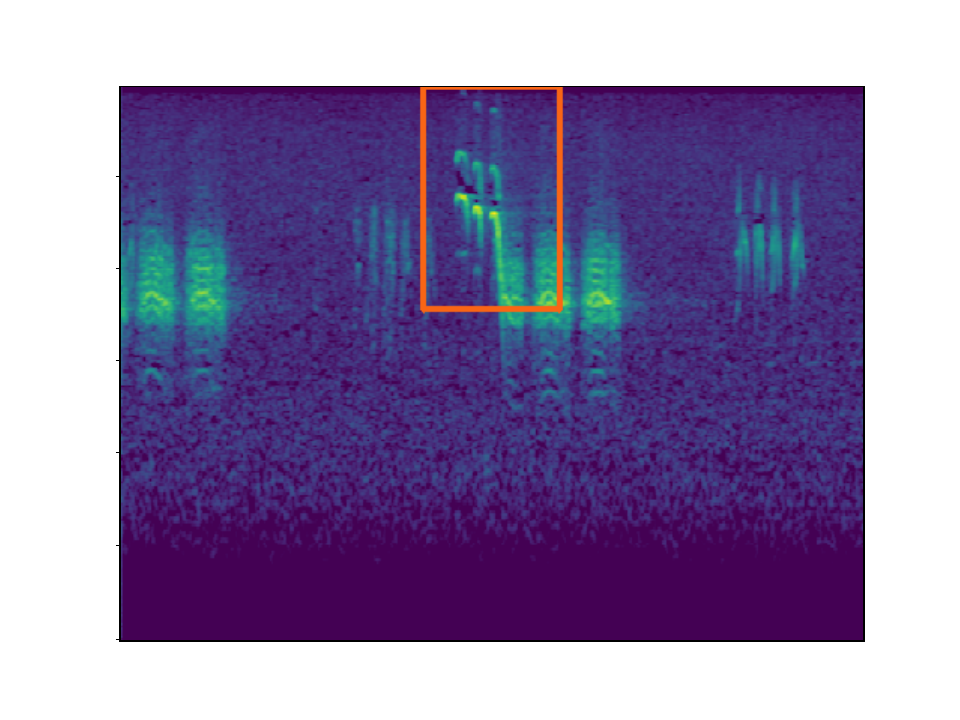}
\par
{\scriptsize $s^{(c,j)} = 0.56$}
\par
\scriptsize\textbf{mouchi}
\end{minipage}
\begin{minipage}{0.19\textwidth}
\centering
\includegraphics[trim={57 35 40 20},clip, width=\textwidth]{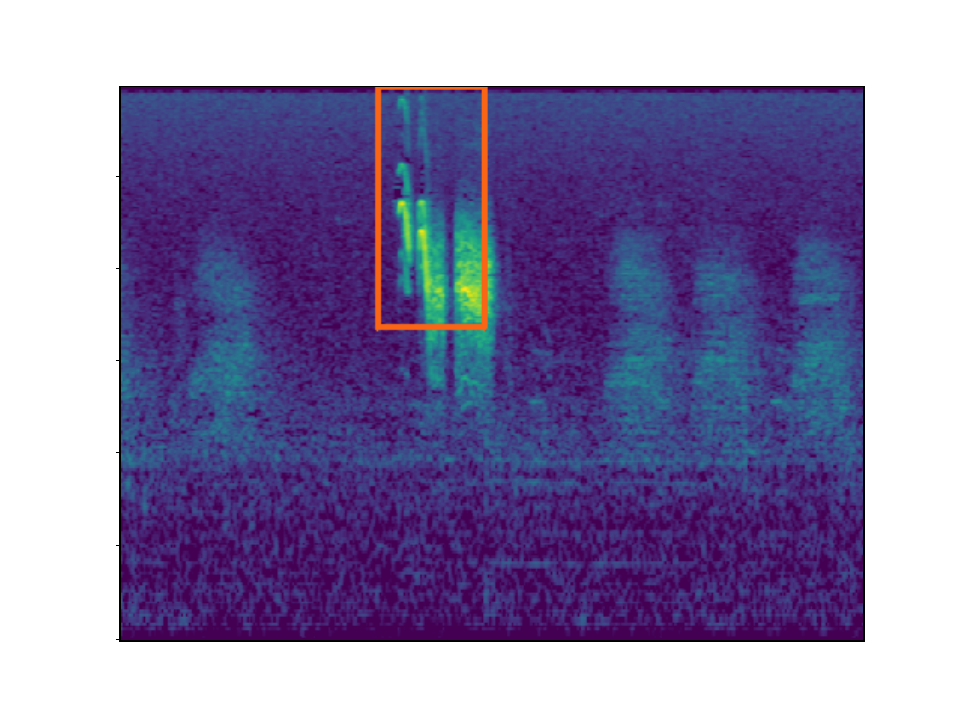}
\par
{\scriptsize $s^{(c,j)} = 0.55$}
\par
\scriptsize\textbf{mouchi}
\end{minipage}
\begin{minipage}{0.19\textwidth}
\centering
\includegraphics[trim={57 35 40 20},clip, width=\textwidth]{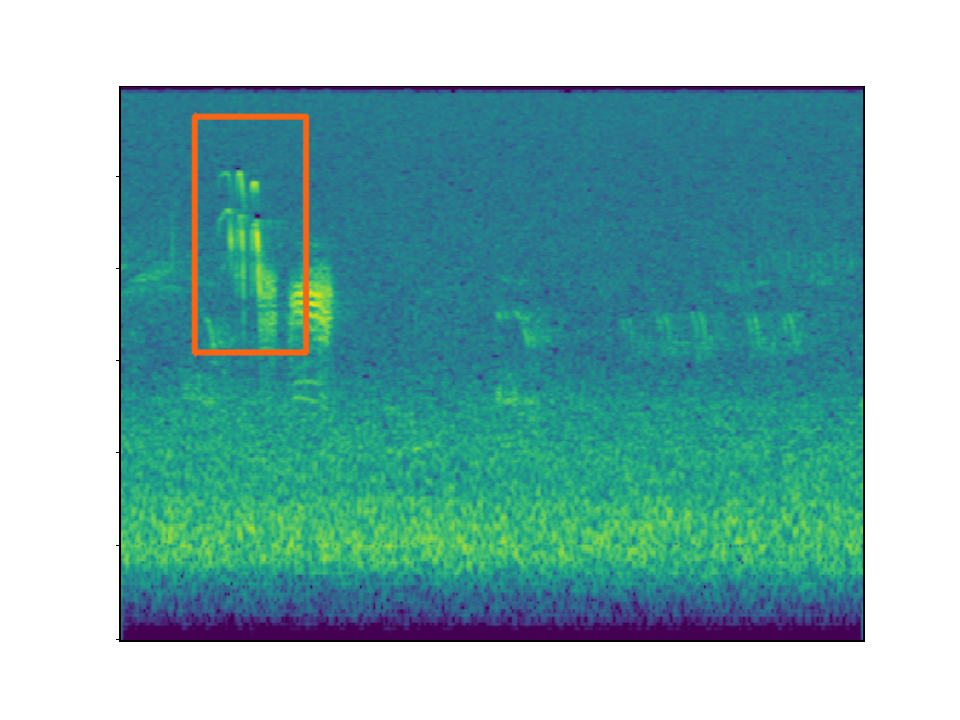}
\par
{\scriptsize $s^{(c,j)} = 0.54$}
\par
\scriptsize\textbf{mouchi}
\end{minipage}
\begin{minipage}{0.19\textwidth}
\centering
\includegraphics[trim={57 35 40 20},clip, width=\textwidth]{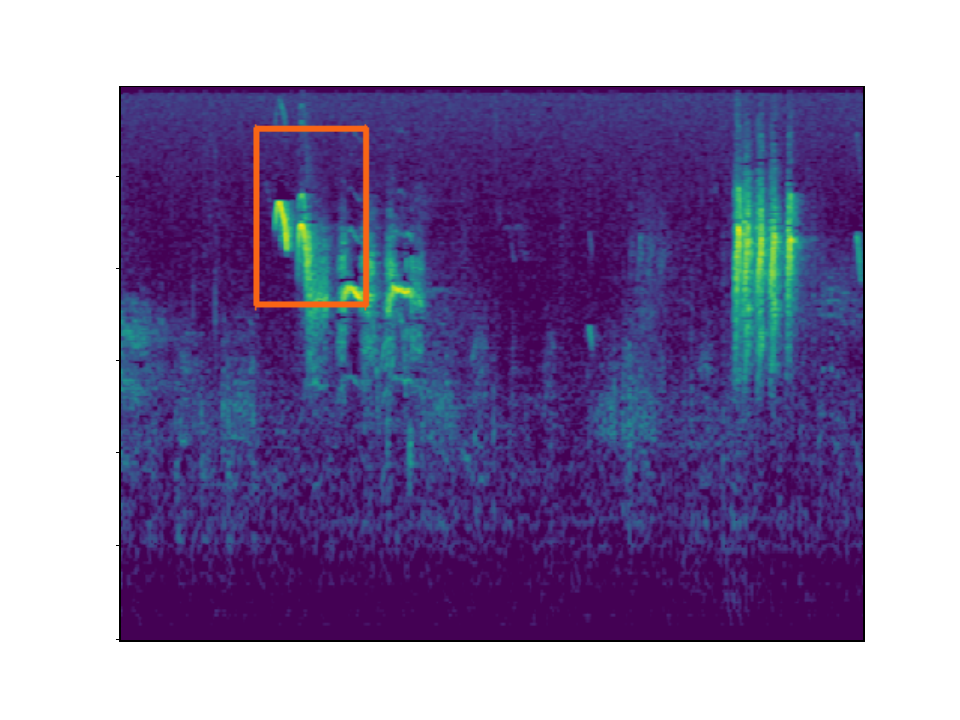}
\par
{\scriptsize $s^{(c,j)} = 0.54$}
\par
\scriptsize\textbf{mouchi}
\end{minipage}
\end{minipage}
\end{minipage}
\begin{minipage}{\textwidth}
\begin{minipage}{0.025\textwidth}
\rotatebox{90}{\scriptsize\textbf{Prototype 4}}
\end{minipage}
\begin{minipage}{0.025\textwidth}
\rotatebox{90}{\scriptsize $w^{(j,c)} = 2.26$}
\end{minipage}
\begin{minipage}{0.95\textwidth}
\centering
\begin{minipage}{0.19\textwidth}
\centering
\includegraphics[trim={57 35 40 20},clip, width=\textwidth]{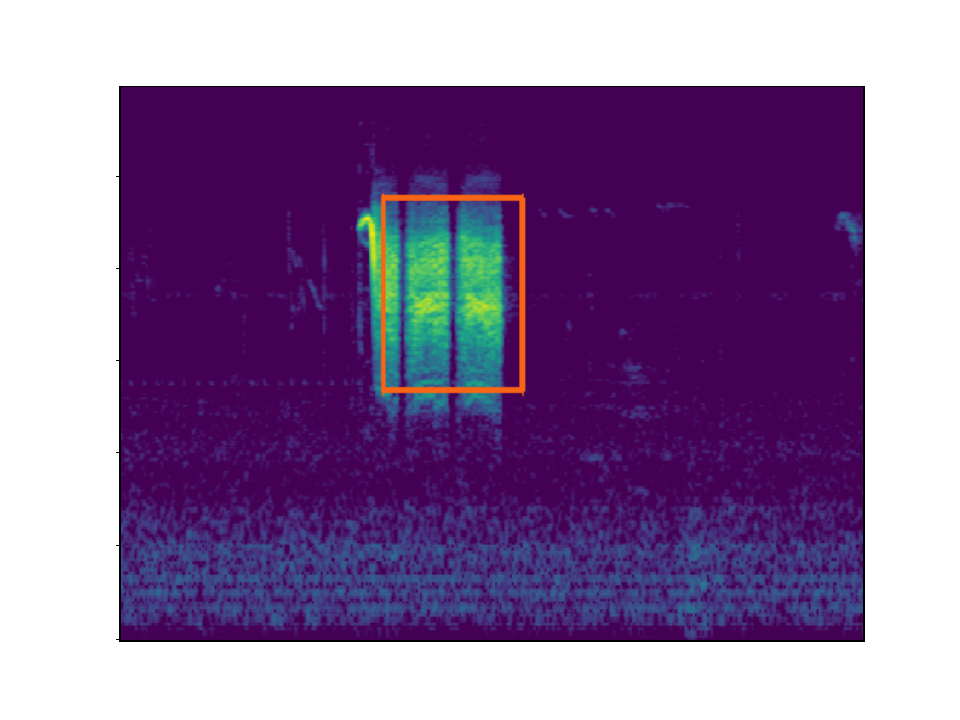}
\par
{\scriptsize $s^{(c,j)} = 0.41$}
\par
\scriptsize\textbf{mouchi}
\end{minipage}
\begin{minipage}{0.19\textwidth}
\centering
\includegraphics[trim={57 35 40 20},clip, width=\textwidth]{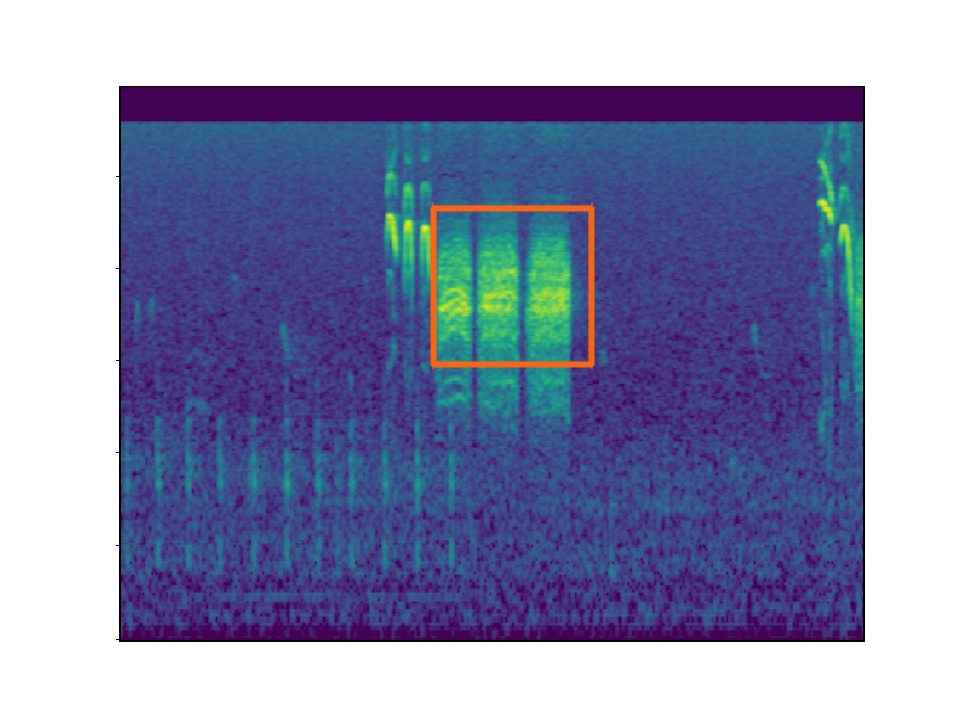}
\par
{\scriptsize $s^{(c,j)} = 0.41$}
\par
\scriptsize\textbf{mouchi}
\end{minipage}
\begin{minipage}{0.19\textwidth}
\centering
\includegraphics[trim={57 35 40 20},clip, width=\textwidth]{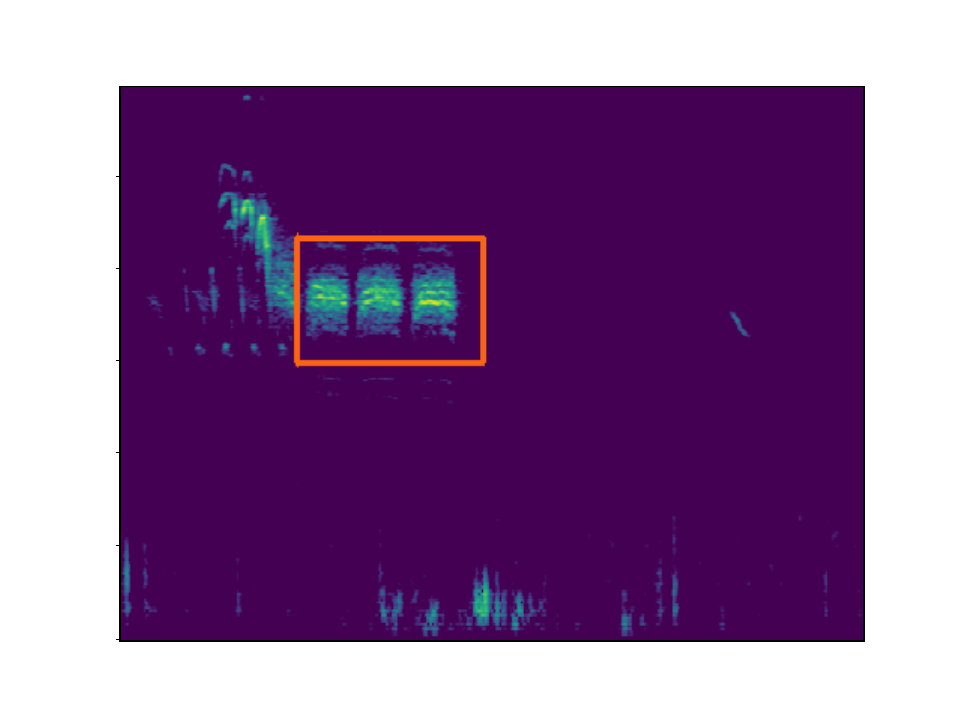}
\par
{\scriptsize $s^{(c,j)} = 0.40$}
\par
\scriptsize\textbf{mouchi}
\end{minipage}
\begin{minipage}{0.19\textwidth}
\centering
\includegraphics[trim={57 35 40 20},clip, width=\textwidth]{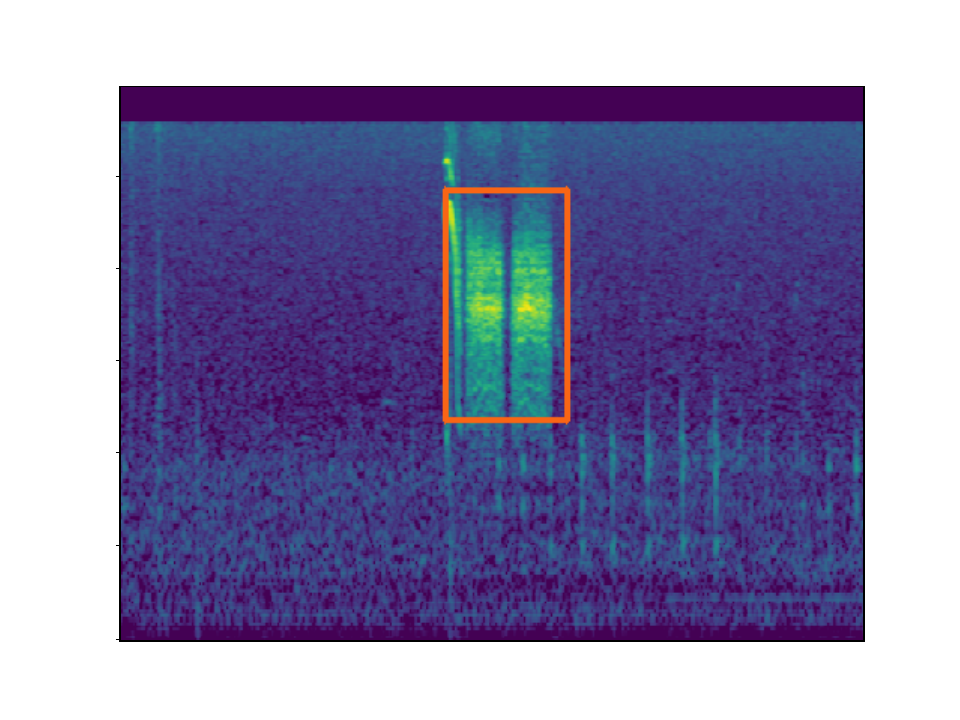}
\par
{\scriptsize $s^{(c,j)} = 0.40$}
\par
\scriptsize\textbf{mouchi}
\end{minipage}
\begin{minipage}{0.19\textwidth}
\centering
\includegraphics[trim={57 35 40 20},clip, width=\textwidth]{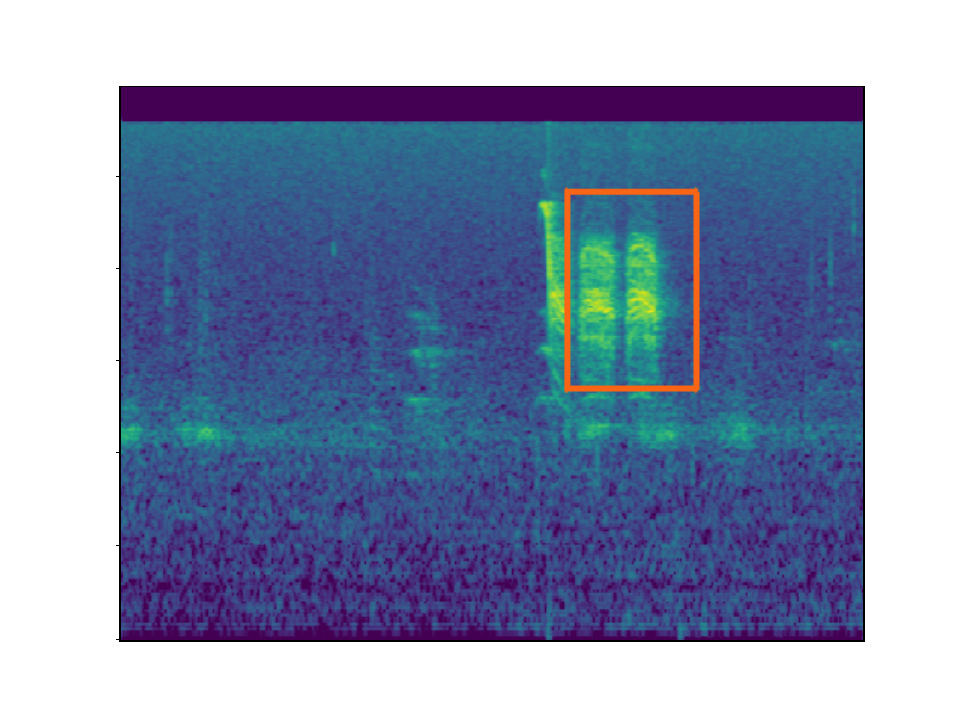}
\par
{\scriptsize $s^{(c,j)} = 0.40$}
\par
\scriptsize\textbf{mouchi}
\end{minipage}
\end{minipage}
\end{minipage}
\begin{minipage}{\textwidth}
\begin{minipage}{0.025\textwidth}
\rotatebox{90}{\scriptsize\textbf{Prototype 5}}
\end{minipage}
\begin{minipage}{0.025\textwidth}
\rotatebox{90}{\scriptsize $w^{(j,c)} = 2.25$}
\end{minipage}
\begin{minipage}{0.95\textwidth}
\centering
\begin{minipage}{0.19\textwidth}
\centering
\includegraphics[trim={57 35 40 20},clip, width=\textwidth]{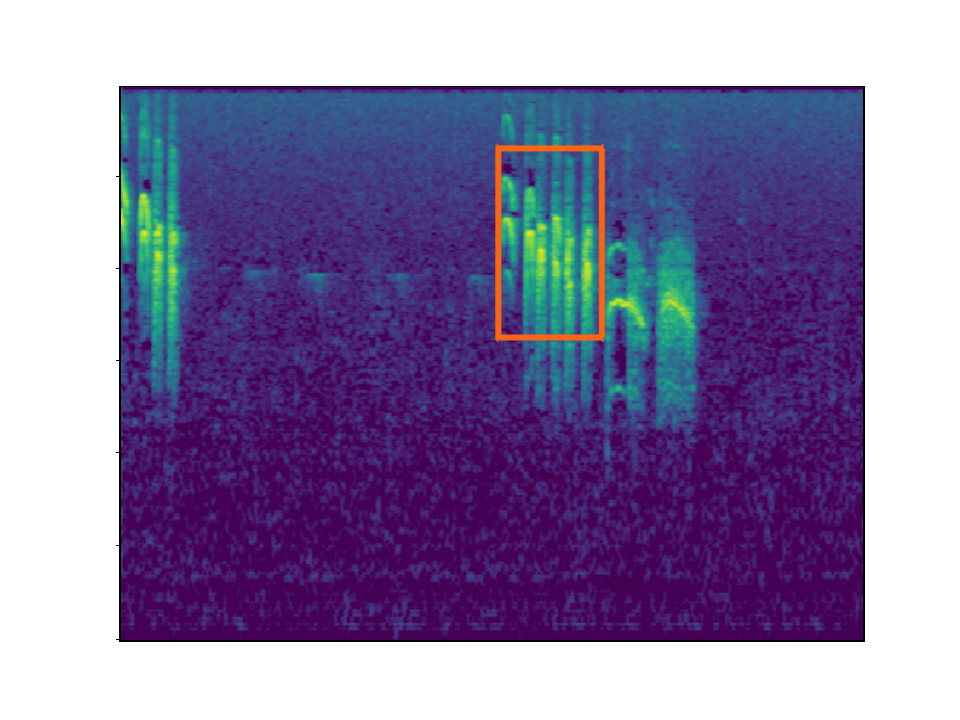}
\par
{\scriptsize $s^{(c,j)} = 0.46$}
\par
\scriptsize\textbf{mouchi}
\end{minipage}
\begin{minipage}{0.19\textwidth}
\centering
\includegraphics[trim={57 35 40 20},clip, width=\textwidth]{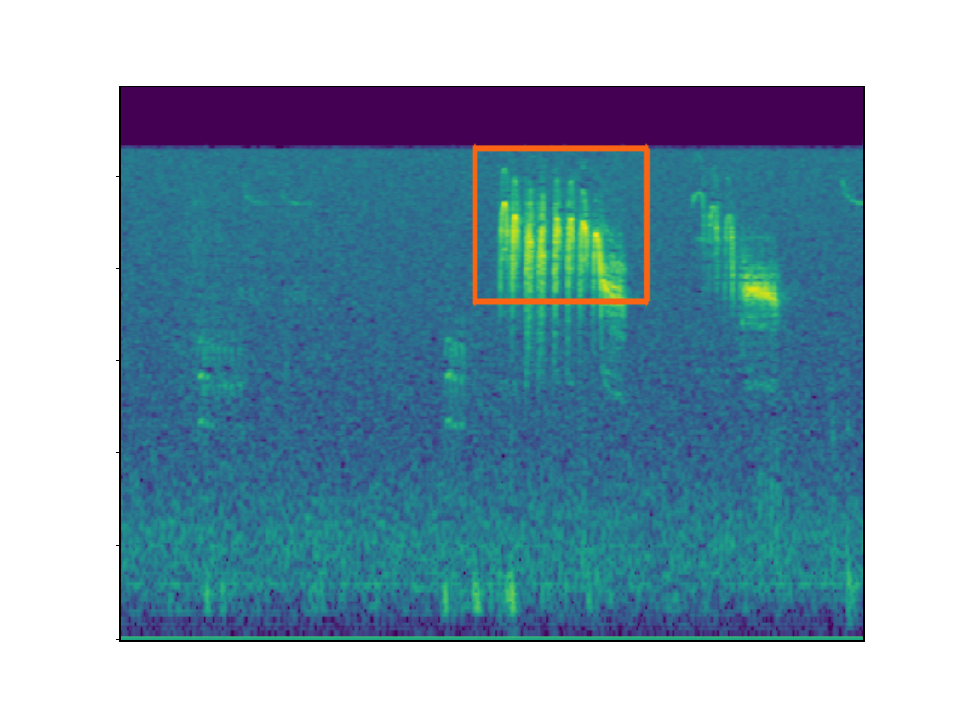}
\par
{\scriptsize $s^{(c,j)} = 0.45$}
\par
\scriptsize\textbf{mouchi}
\end{minipage}
\begin{minipage}{0.19\textwidth}
\centering
\includegraphics[trim={57 35 40 20},clip, width=\textwidth]{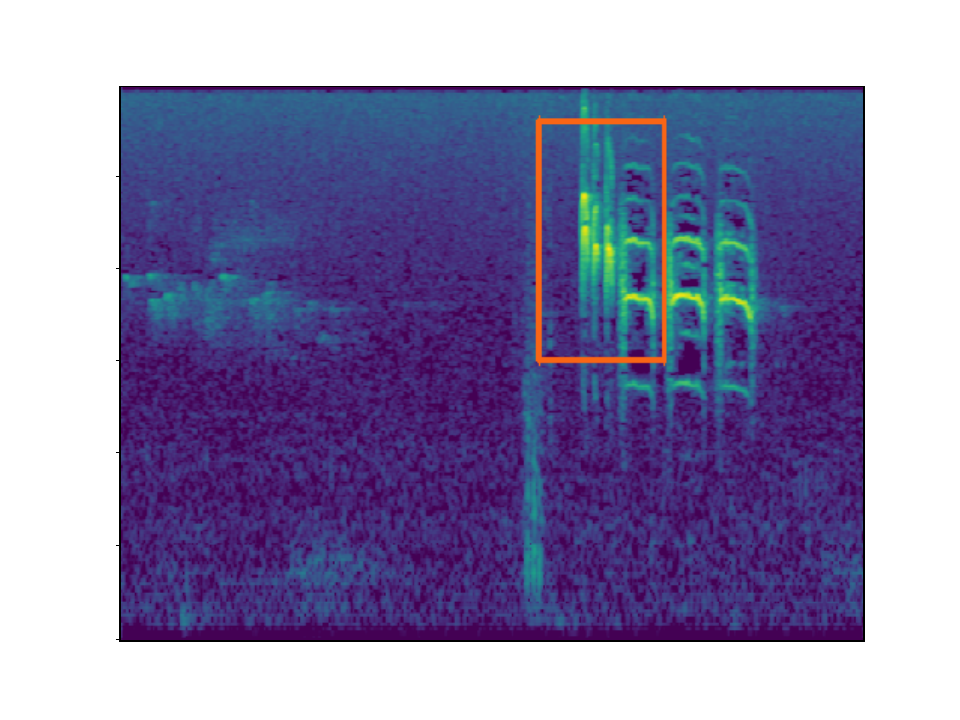}
\par
{\scriptsize $s^{(c,j)} = 0.45$}
\par
\scriptsize\textbf{mouchi}
\end{minipage}
\begin{minipage}{0.19\textwidth}
\centering
\includegraphics[trim={57 35 40 20},clip, width=\textwidth]{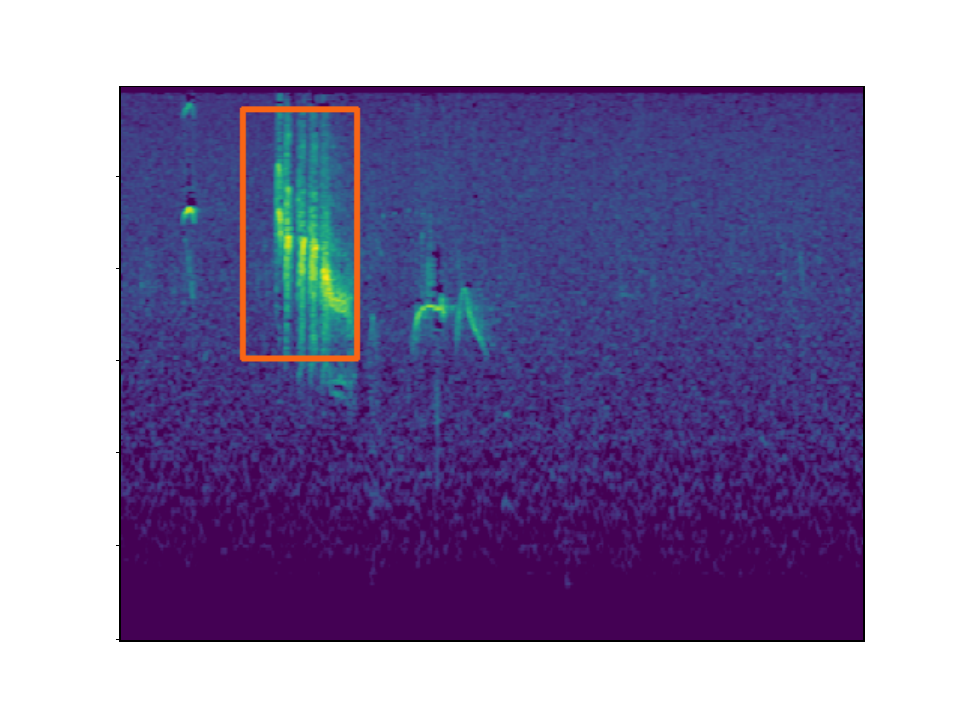}
\par
{\scriptsize $s^{(c,j)} = 0.45$}
\par
\scriptsize\textbf{mouchi}
\end{minipage}
\begin{minipage}{0.19\textwidth}
\centering
\includegraphics[trim={57 35 40 20},clip, width=\textwidth]{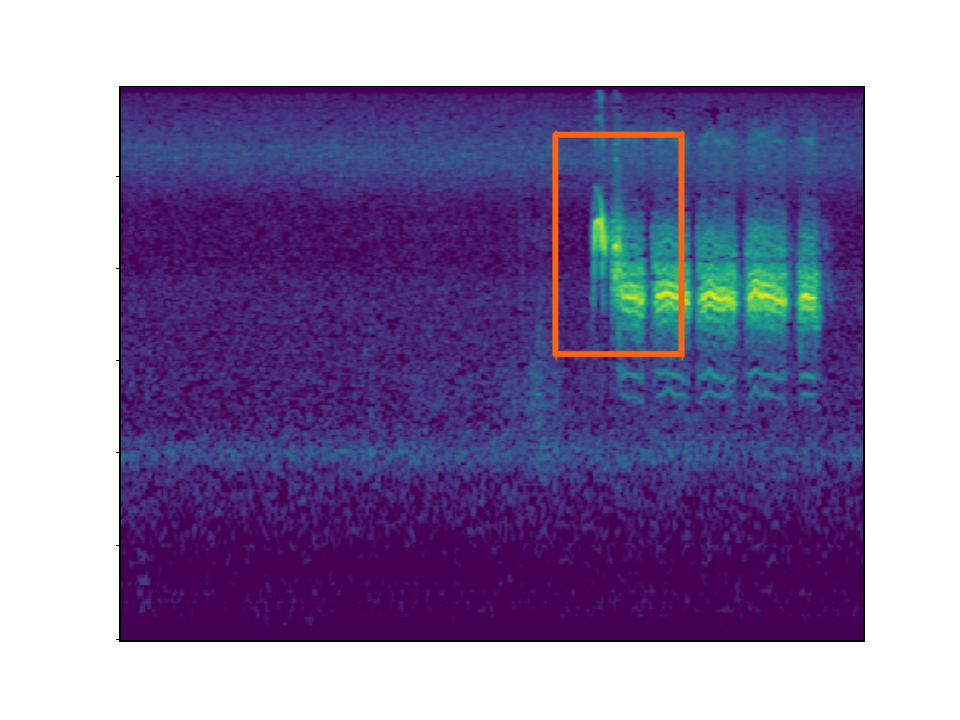}
\par
{\scriptsize $s^{(c,j)} = 0.45$}
\par
\scriptsize\textbf{mouchi}
\end{minipage}
\end{minipage}
\end{minipage}

\caption{The spectrograms of the most similar instances from the SNE subset of the training dataset, with their labels highlighted in bold, for the five prototypes learned by \gls*{audioprotopnet}-5 for the Mountain Chickadee (mouchi). The prototypes
correspond to the parts of the spectrograms surrounded by the orange bounding boxes. Additionally, the similarity values to the prototypes $s^{(c,j)}$ and the weights of the respective prototypes $w^{(j,c)}$ in the final layer are shown.}
\label{fig:prototypes_train_mouchi}
\end{figure}

\begin{figure}[ht]
    
    \centering

    \begin{minipage}{\textwidth}
        \begin{minipage}{0.025\textwidth}
            \rotatebox{90}{\scriptsize\textbf{Prototype 1}}
        \end{minipage}
        \begin{minipage}{0.025\textwidth}
            \rotatebox{90}{\scriptsize $w^{(j,c)} = 2.34$}
        \end{minipage}
        \begin{minipage}{0.95\textwidth}
            \centering
            \begin{minipage}{0.19\textwidth}
                \centering
                \includegraphics[trim={57 35 40 20},clip, width=\textwidth]{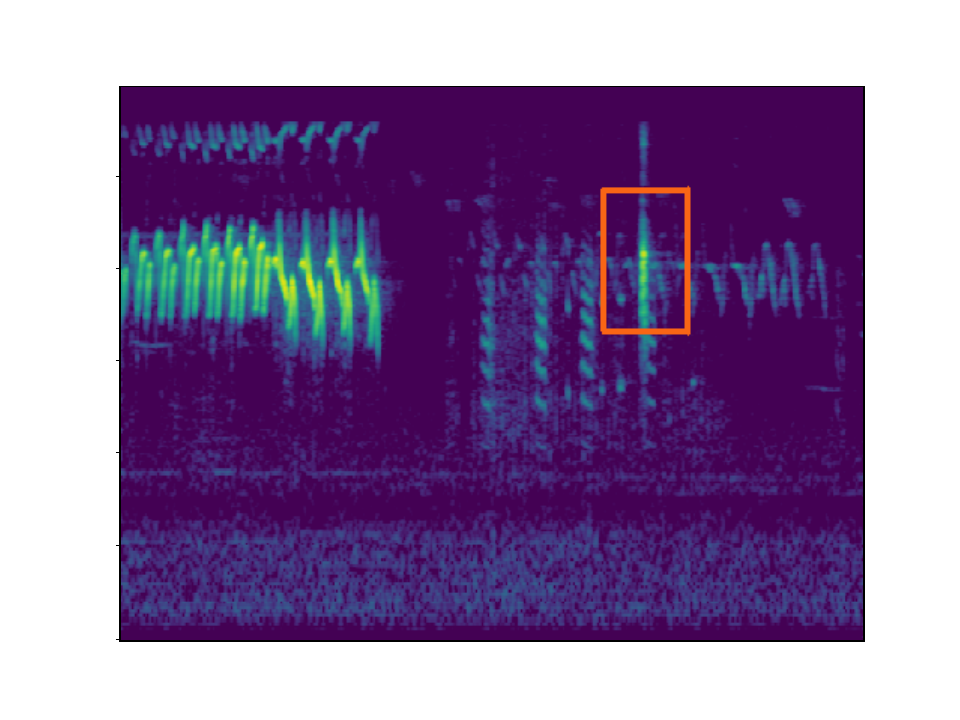}
                \par
                {\scriptsize $s^{(c,j)} = 0.60$}
                \par
                \scriptsize\textbf{yerwar}
            \end{minipage}
            \begin{minipage}{0.19\textwidth}
                \centering
                \includegraphics[trim={57 35 40 20},clip, width=\textwidth]{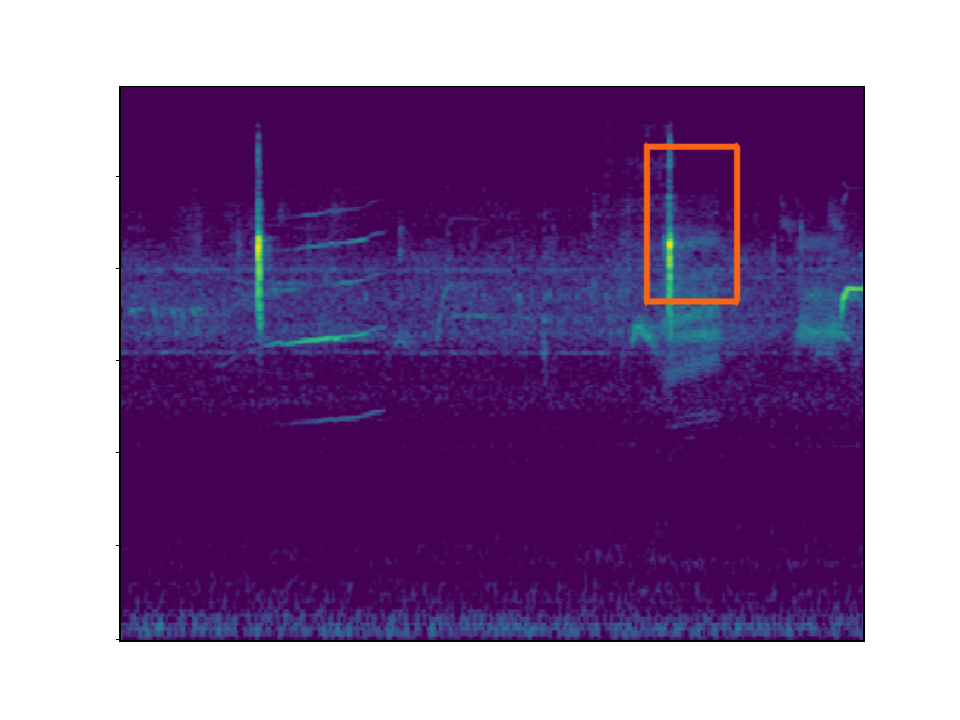}
                \par
                {\scriptsize $s^{(c,j)} = 0.58$}
                \par
                \scriptsize\textbf{yerwar}
            \end{minipage}
            \begin{minipage}{0.19\textwidth}
                \centering
                \includegraphics[trim={57 35 40 20},clip, width=\textwidth]{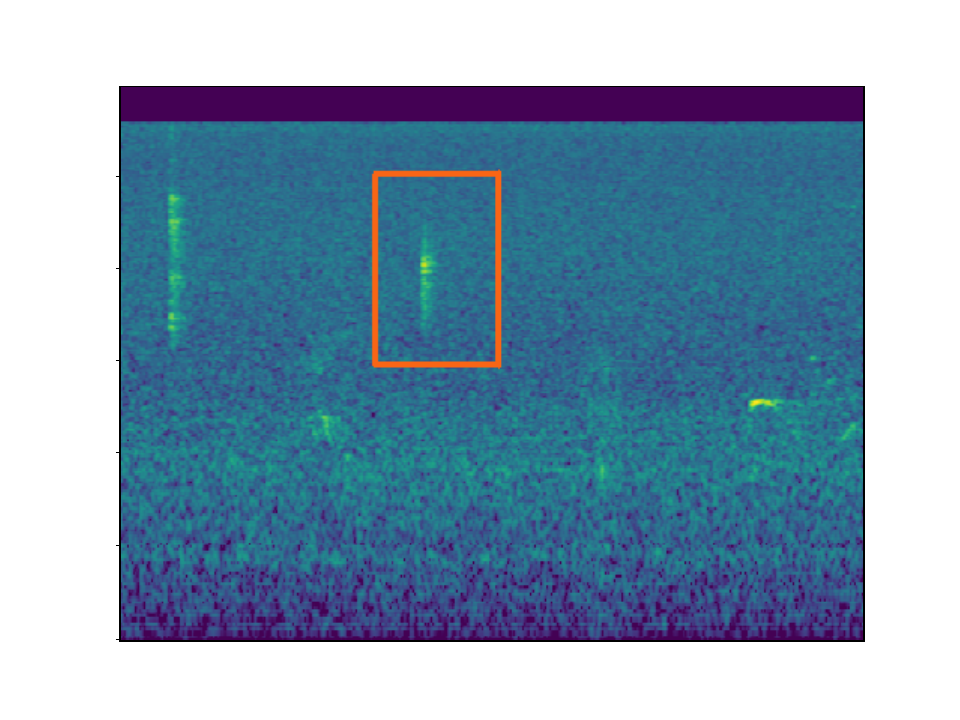}
                \par
                {\scriptsize $s^{(c,j)} = 0.58$}
                \par
                \scriptsize\textbf{wlswar}
            \end{minipage}
            \begin{minipage}{0.19\textwidth}
                \centering
                \includegraphics[trim={57 35 40 20},clip, width=\textwidth]{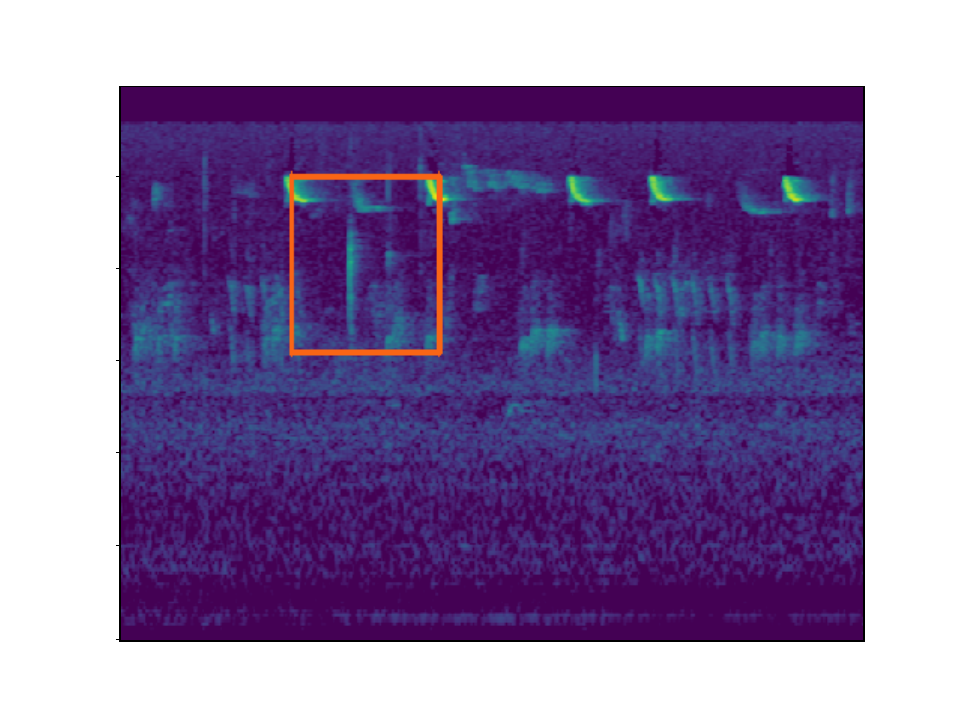}
                \par
                {\scriptsize $s^{(c,j)} = 0.58$}
                \par
                \scriptsize\textbf{brncre}
            \end{minipage}
            \begin{minipage}{0.19\textwidth}
                \centering
                \includegraphics[trim={57 35 40 20},clip, width=\textwidth]{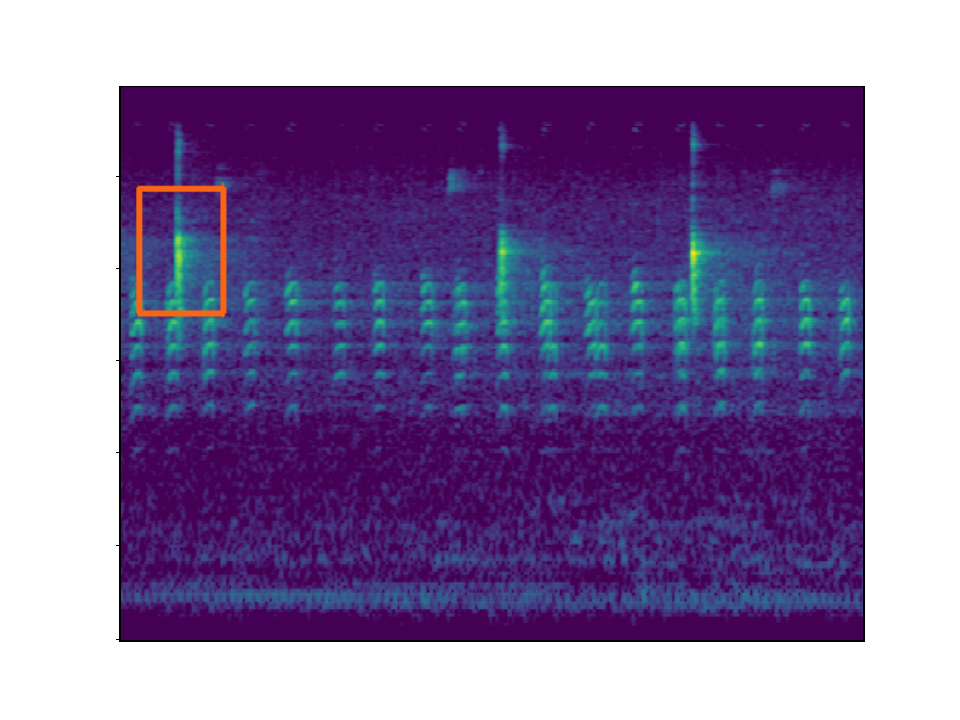}
                \par
                {\scriptsize $s^{(c,j)} = 0.57$}
                \par
                \scriptsize\textbf{yerwar}
            \end{minipage}
        \end{minipage}
    \end{minipage}

    \begin{minipage}{\textwidth}
        \begin{minipage}{0.025\textwidth}
            \rotatebox{90}{\scriptsize\textbf{Prototype 2}}
        \end{minipage}
        \begin{minipage}{0.025\textwidth}
            \rotatebox{90}{\scriptsize $w^{(j,c)} = 2.16$}
        \end{minipage}
        \begin{minipage}{0.95\textwidth}
            \centering
            \begin{minipage}{0.19\textwidth}
                \centering
                \includegraphics[trim={57 35 40 20},clip, width=\textwidth]{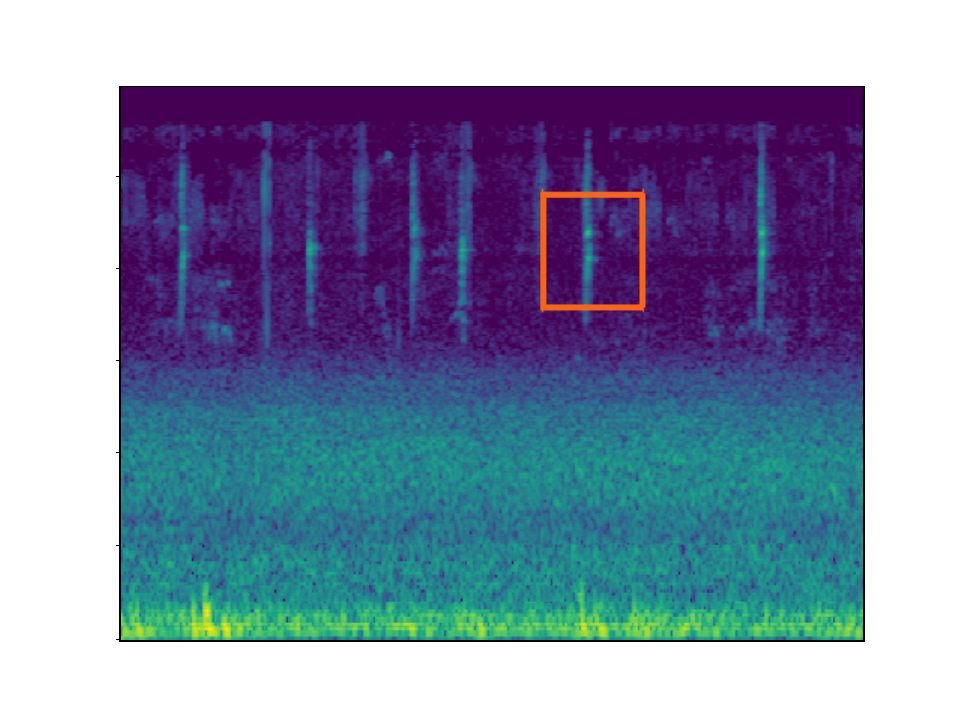}
                \par
                {\scriptsize $s^{(c,j)} = 0.40$}
                \par
                \scriptsize\textbf{yerwar}
            \end{minipage}
            \begin{minipage}{0.19\textwidth}
                \centering
                \includegraphics[trim={57 35 40 20},clip, width=\textwidth]{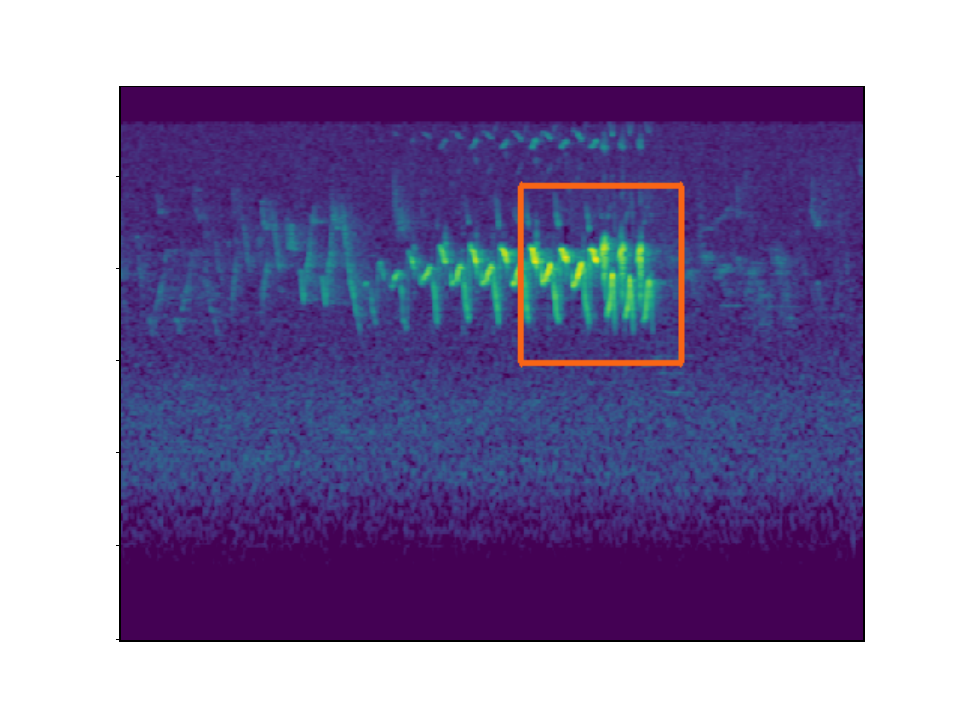}
                \par
                {\scriptsize $s^{(c,j)} = 0.39$}
                \par
                \scriptsize\textbf{yerwar}
            \end{minipage}
            \begin{minipage}{0.19\textwidth}
                \centering
                \includegraphics[trim={57 35 40 20},clip, width=\textwidth]{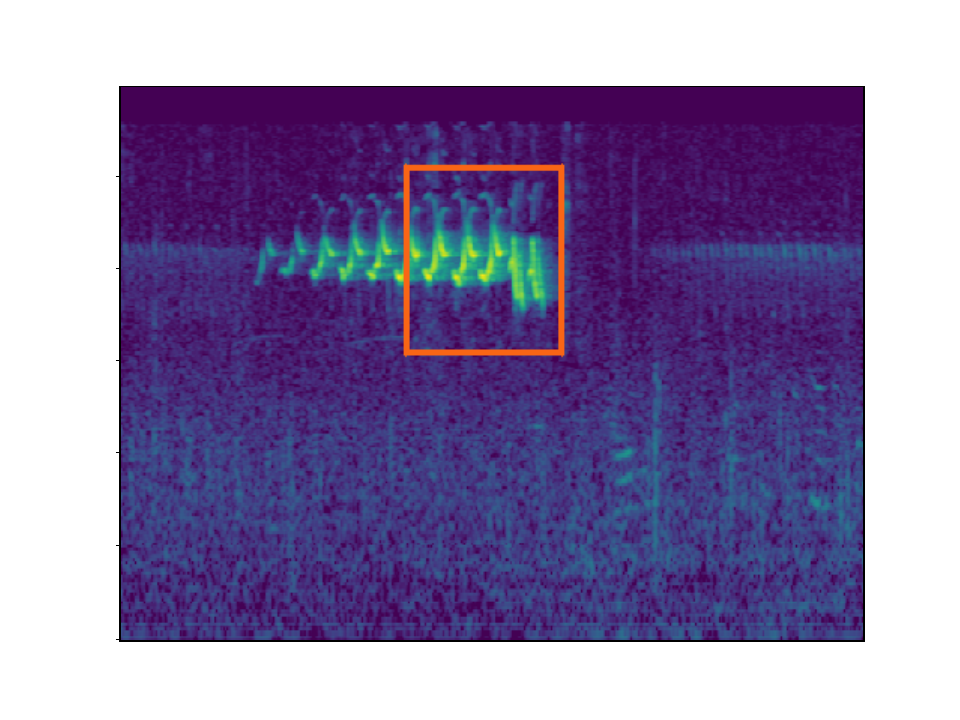}
                \par
                {\scriptsize $s^{(c,j)} = 0.38$}
                \par
                \scriptsize\textbf{yerwar}
            \end{minipage}
            \begin{minipage}{0.19\textwidth}
                \centering
                \includegraphics[trim={57 35 40 20},clip, width=\textwidth]{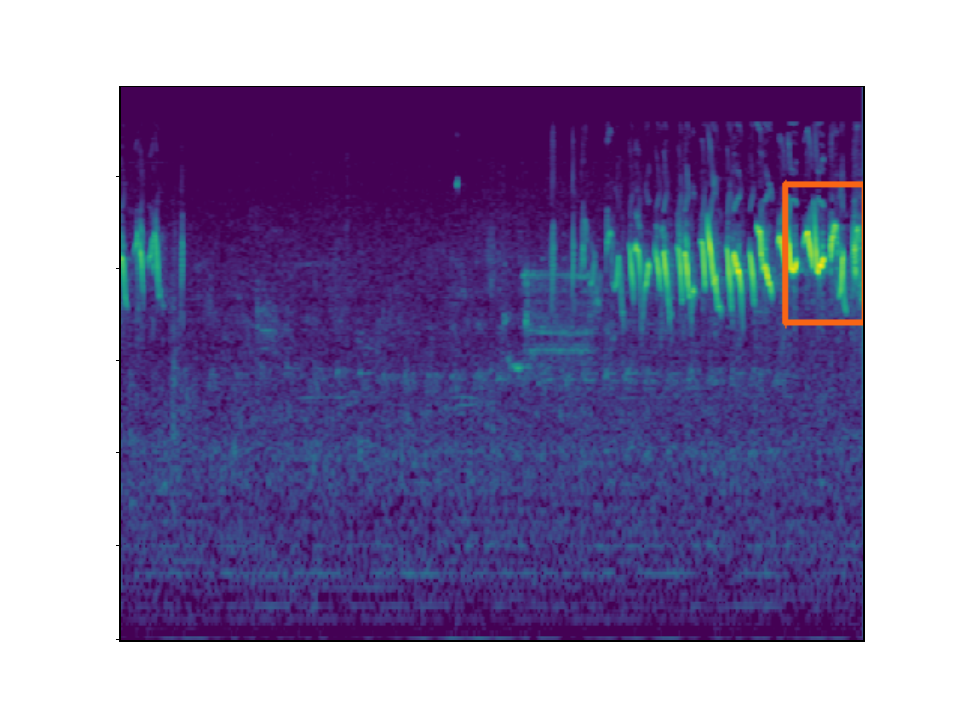}
                \par
                {\scriptsize $s^{(c,j)} = 0.37$}
                \par
                \scriptsize\textbf{yerwar}
            \end{minipage}
            \begin{minipage}{0.19\textwidth}
                \centering
                \includegraphics[trim={57 35 40 20},clip, width=\textwidth]{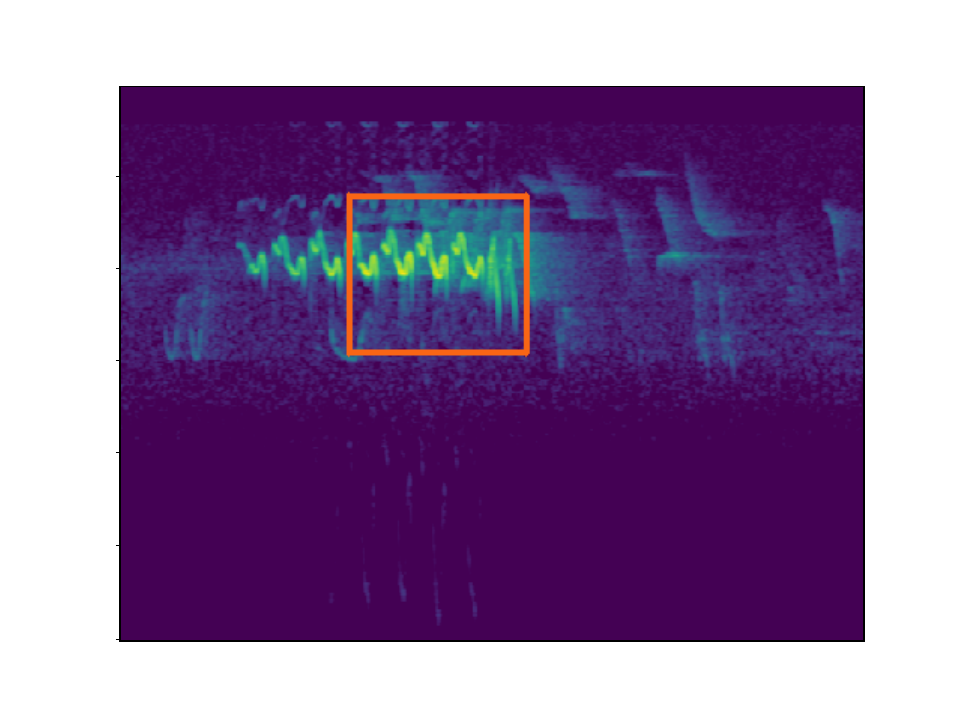}
                \par
                {\scriptsize $s^{(c,j)} = 0.37$}
                \par
                \scriptsize\textbf{yerwar}
            \end{minipage}
        \end{minipage}
    \end{minipage}

    \begin{minipage}{\textwidth}
        \begin{minipage}{0.025\textwidth}
            \rotatebox{90}{\scriptsize\textbf{Prototype 3}}
        \end{minipage}
        \begin{minipage}{0.025\textwidth}
            \rotatebox{90}{\scriptsize $w^{(j,c)} = 1.0$}
        \end{minipage}
        \begin{minipage}{0.95\textwidth}
            \centering
            \begin{minipage}{0.19\textwidth}
                \centering
                \includegraphics[trim={57 35 40 20},clip, width=\textwidth]{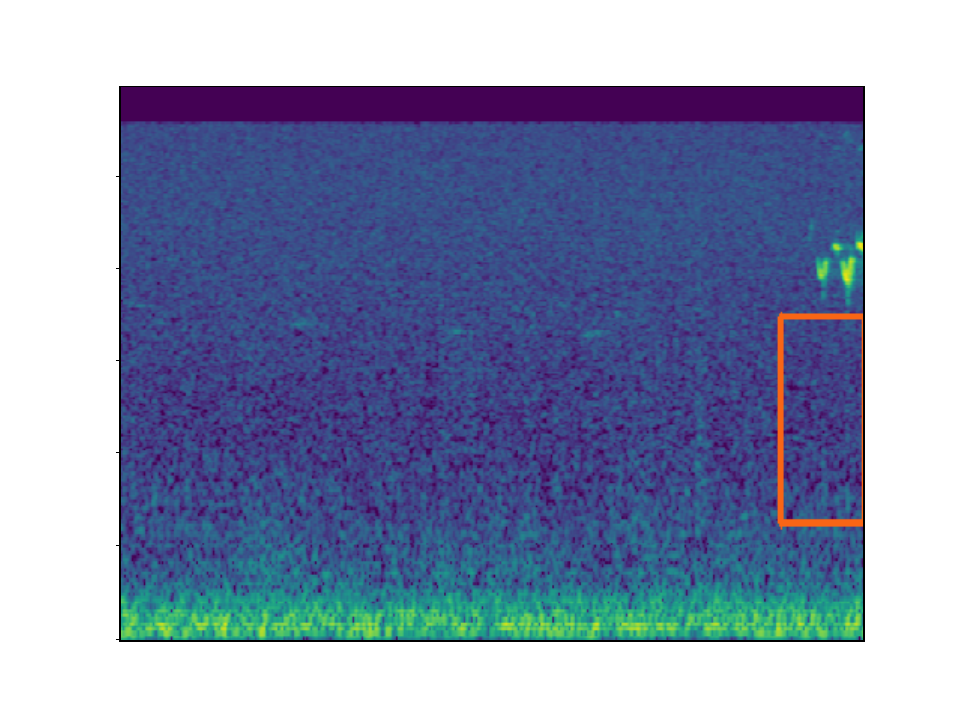}
                \par
                {\scriptsize $s^{(c,j)} = 0.43$}
                \par
                \scriptsize\textbf{yerwar}
            \end{minipage}
            \begin{minipage}{0.19\textwidth}
                \centering
                \includegraphics[trim={57 35 40 20},clip, width=\textwidth]{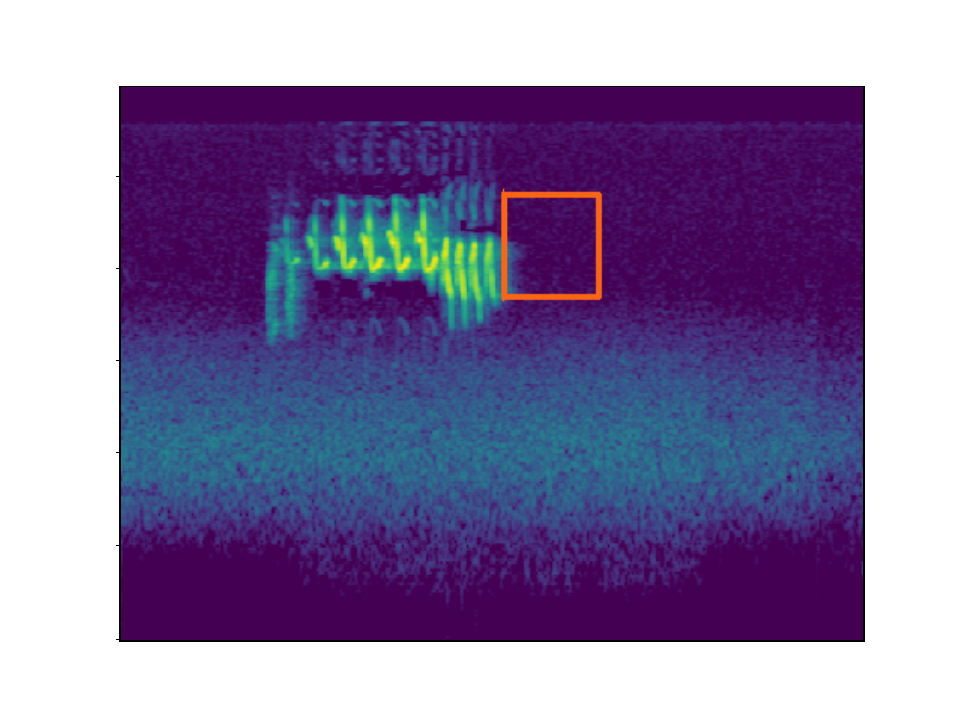}
                \par
                {\scriptsize $s^{(c,j)} = 0.41$}
                \par
                \scriptsize\textbf{yerwar}
            \end{minipage}
            \begin{minipage}{0.19\textwidth}
                \centering
                \includegraphics[trim={57 35 40 20},clip, width=\textwidth]{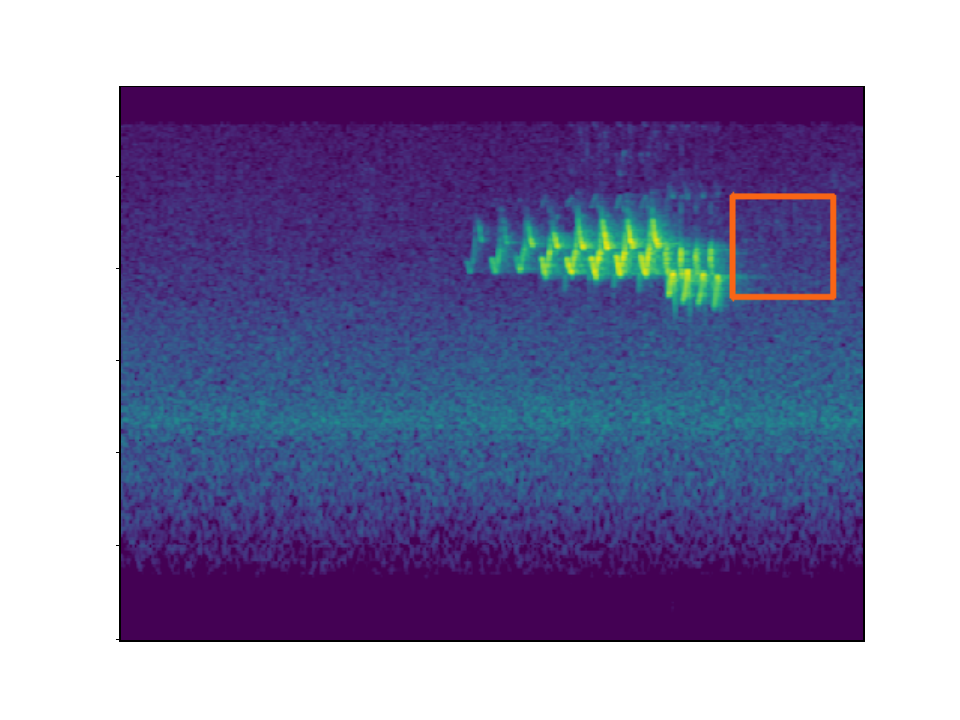}
                \par
                {\scriptsize $s^{(c,j)} = 0.40$}
                \par
                \scriptsize\textbf{yerwar}
            \end{minipage}
            \begin{minipage}{0.19\textwidth}
                \centering
                \includegraphics[trim={57 35 40 20},clip, width=\textwidth]{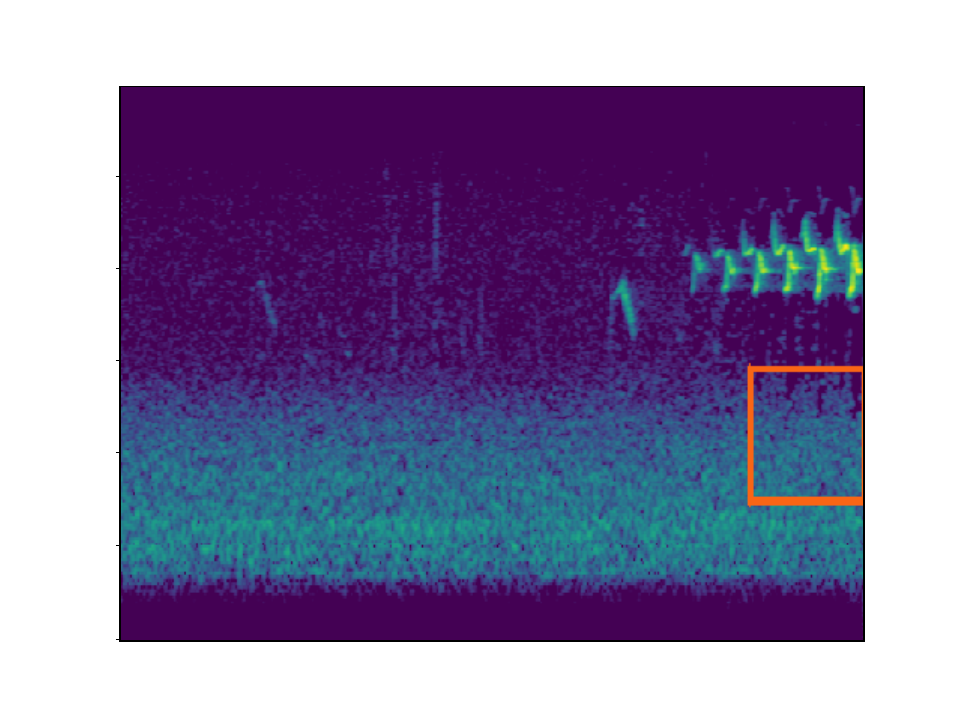}
                \par
                {\scriptsize $s^{(c,j)} = 0.39$}
                \par
                \scriptsize\textbf{yerwar}
            \end{minipage}
            \begin{minipage}{0.19\textwidth}
                \centering
                \includegraphics[trim={57 35 40 20},clip, width=\textwidth]{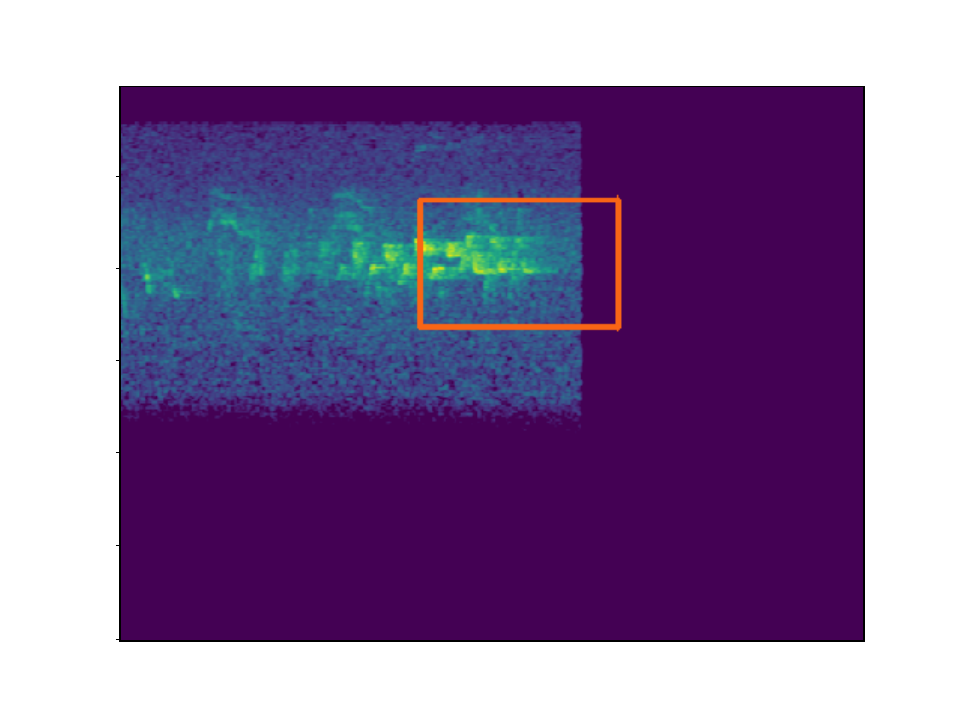}
                \par
                {\scriptsize $s^{(c,j)} = 0.38$}
                \par
                \scriptsize\textbf{yerwar}
            \end{minipage}
        \end{minipage}
    \end{minipage}

    \begin{minipage}{\textwidth}
        \begin{minipage}{0.025\textwidth}
            \rotatebox{90}{\scriptsize\textbf{Prototype 4}}
        \end{minipage}
        \begin{minipage}{0.025\textwidth}
            \rotatebox{90}{\scriptsize $w^{(j,c)} = 2.38$}
        \end{minipage}
        \begin{minipage}{0.95\textwidth}
            \centering
            \begin{minipage}{0.19\textwidth}
                \centering
                \includegraphics[trim={57 35 40 20},clip, width=\textwidth]{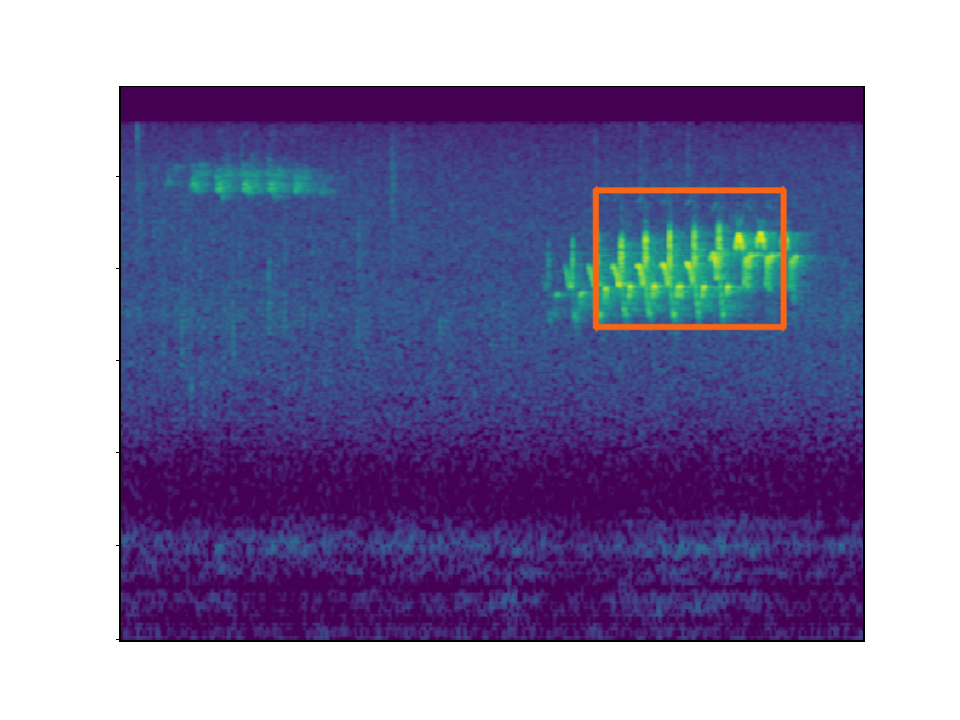}
                \par
                {\scriptsize $s^{(c,j)} = 0.42$}
                \par
                \scriptsize\textbf{yerwar}
            \end{minipage}
            \begin{minipage}{0.19\textwidth}
                \centering
                \includegraphics[trim={57 35 40 20},clip, width=\textwidth]{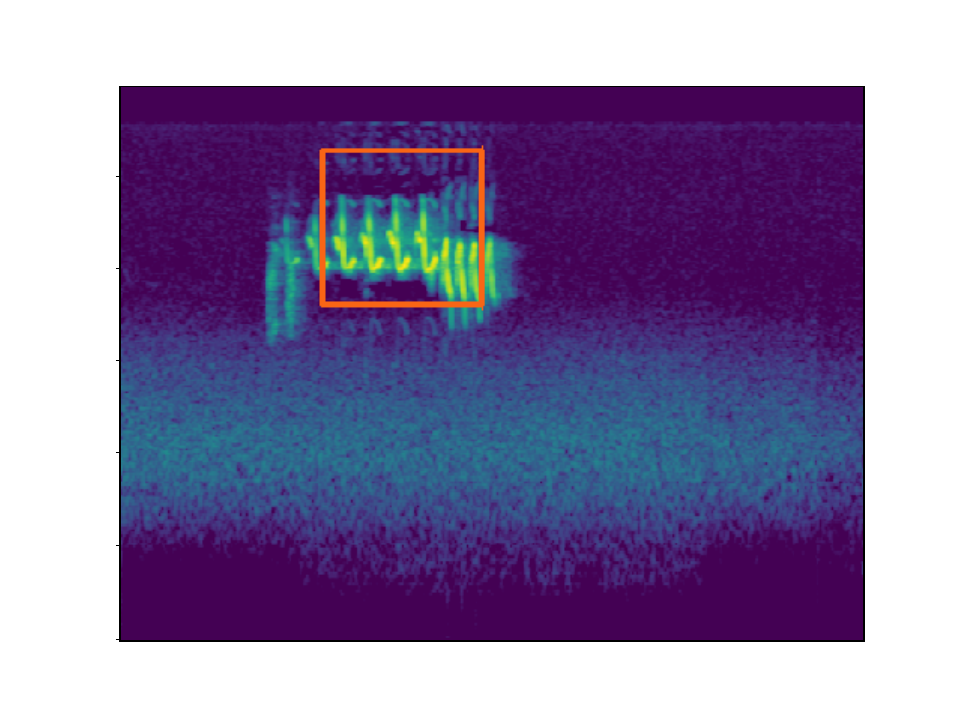}
                \par
                {\scriptsize $s^{(c,j)} = 0.42$}
                \par
                \scriptsize\textbf{yerwar}
            \end{minipage}
            \begin{minipage}{0.19\textwidth}
                \centering
                \includegraphics[trim={57 35 40 20},clip, width=\textwidth]{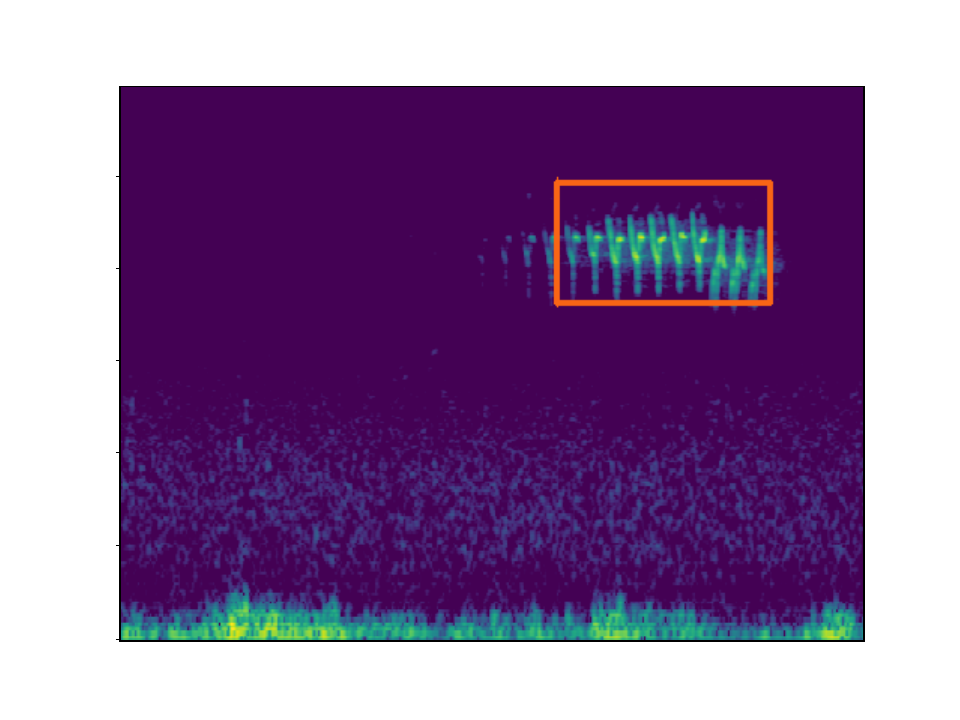}
                \par
                {\scriptsize $s^{(c,j)} = 0.42$}
                \par
                \scriptsize\textbf{yerwar}
            \end{minipage}
            \begin{minipage}{0.19\textwidth}
                \centering
                \includegraphics[trim={57 35 40 20},clip, width=\textwidth]{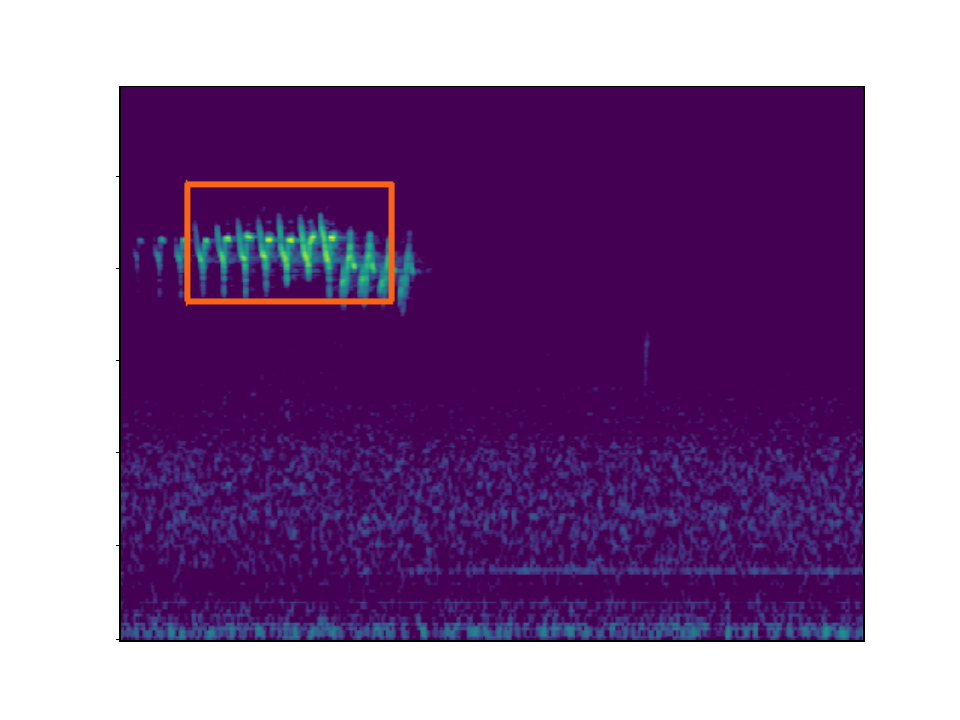}
                \par
                {\scriptsize $s^{(c,j)} = 0.42$}
                \par
                \scriptsize\textbf{yerwar}
            \end{minipage}
            \begin{minipage}{0.19\textwidth}
                \centering
                \includegraphics[trim={57 35 40 20},clip, width=\textwidth]{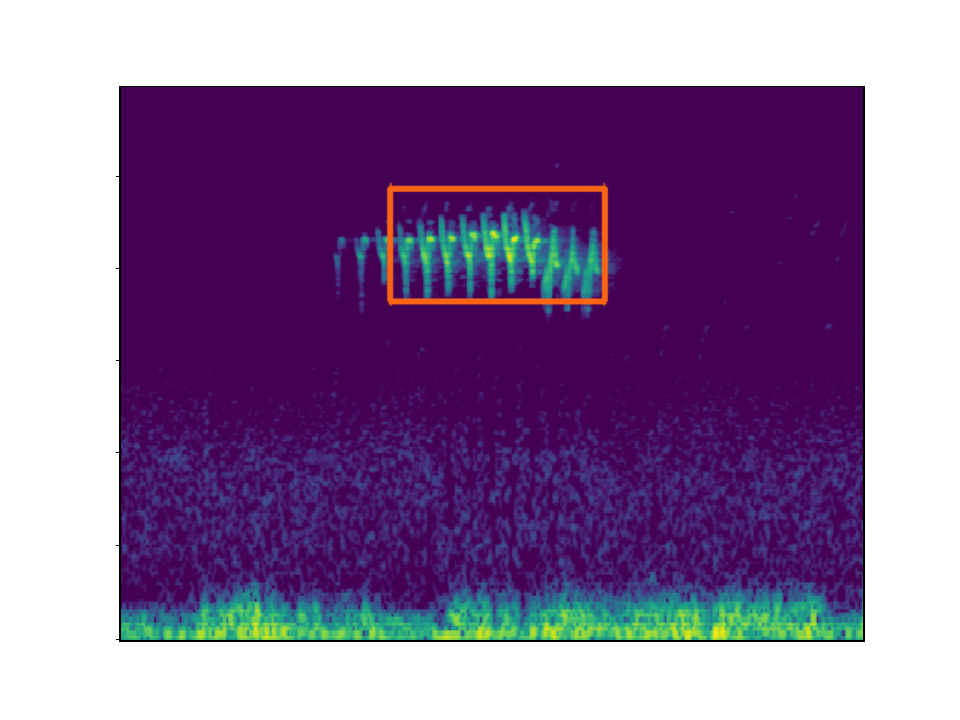}
                \par
                {\scriptsize $s^{(c,j)} = 0.42$}
                \par
                \scriptsize\textbf{yerwar}
            \end{minipage}
        \end{minipage}
    \end{minipage}

    \begin{minipage}{\textwidth}
        \begin{minipage}{0.025\textwidth}
            \rotatebox{90}{\scriptsize\textbf{Prototype 5}}
        \end{minipage}
        \begin{minipage}{0.025\textwidth}
            \rotatebox{90}{\scriptsize $w^{(j,c)} = 1.41$}
        \end{minipage}
        \begin{minipage}{0.95\textwidth}
            \centering
            \begin{minipage}{0.19\textwidth}
                \centering
                \includegraphics[trim={57 35 40 20},clip, width=\textwidth]{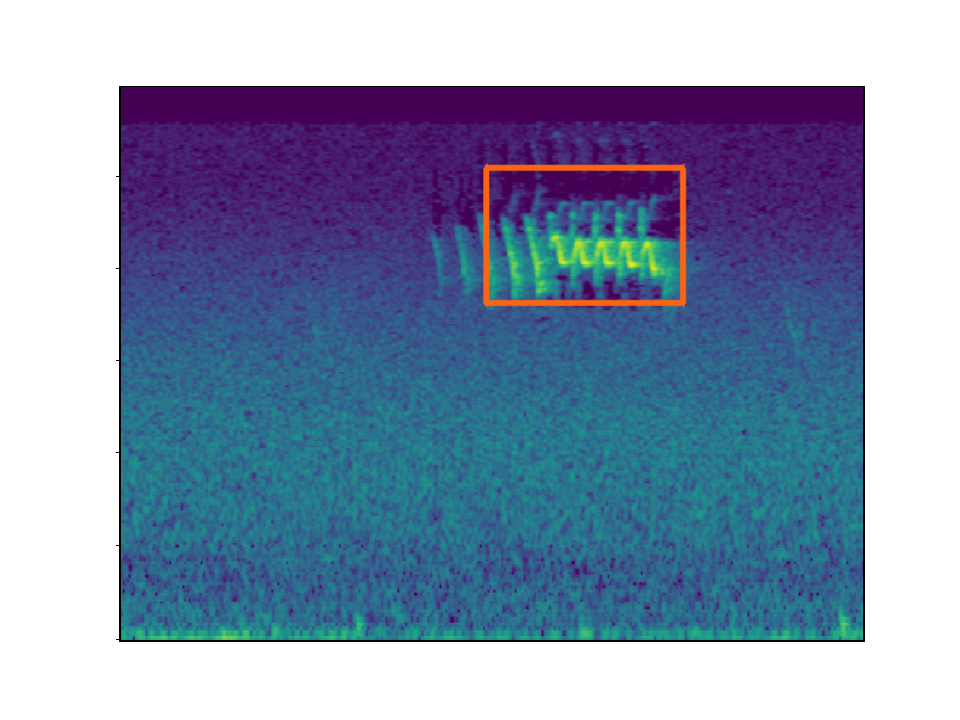}
                \par
                {\scriptsize $s^{(c,j)} = 0.36$}
                \par
                \scriptsize\textbf{yerwar}
            \end{minipage}
            \begin{minipage}{0.19\textwidth}
                \centering
                \includegraphics[trim={57 35 40 20},clip, width=\textwidth]{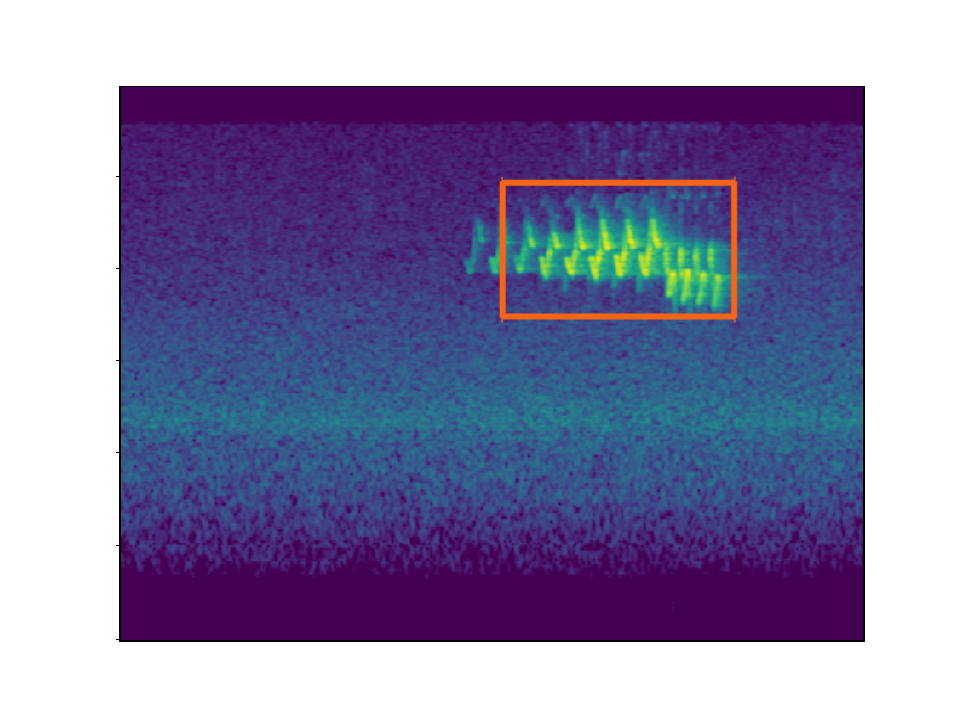}
                \par
                {\scriptsize $s^{(c,j)} = 0.36$}
                \par
                \scriptsize\textbf{yerwar}
            \end{minipage}
            \begin{minipage}{0.19\textwidth}
                \centering
                \includegraphics[trim={57 35 40 20},clip, width=\textwidth]{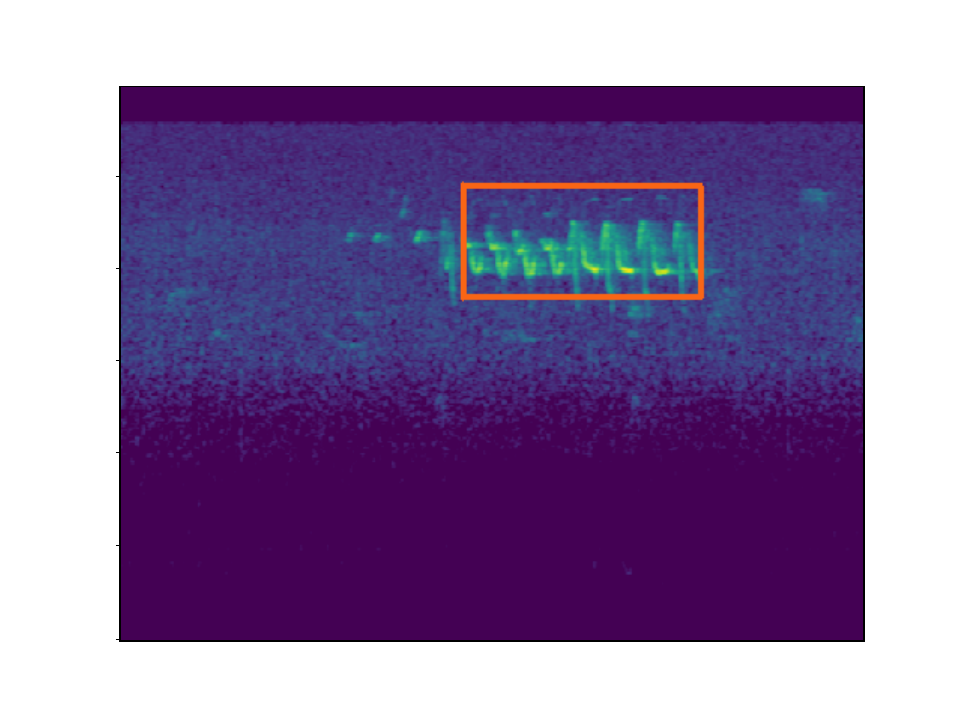}
                \par
                {\scriptsize $s^{(c,j)} = 0.36$}
                \par
                \scriptsize\textbf{yerwar}
            \end{minipage}
            \begin{minipage}{0.19\textwidth}
                \centering
                \includegraphics[trim={57 35 40 20},clip, width=\textwidth]{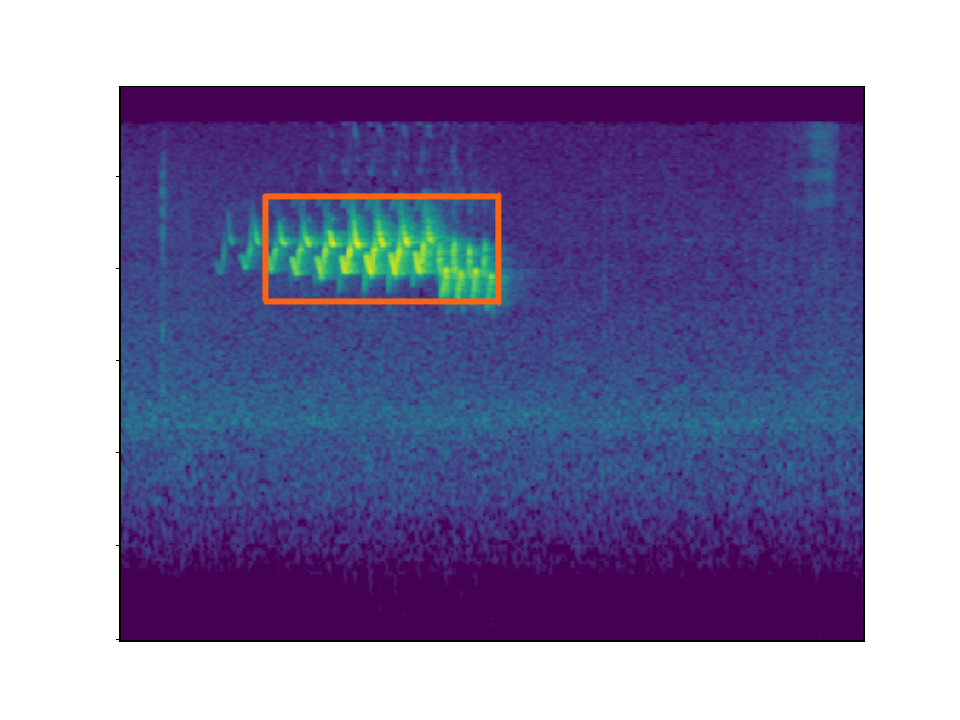}
                \par
                {\scriptsize $s^{(c,j)} = 0.35$}
                \par
                \scriptsize\textbf{yerwar}
            \end{minipage}
            \begin{minipage}{0.19\textwidth}
                \centering
                \includegraphics[trim={57 35 40 20},clip, width=\textwidth]{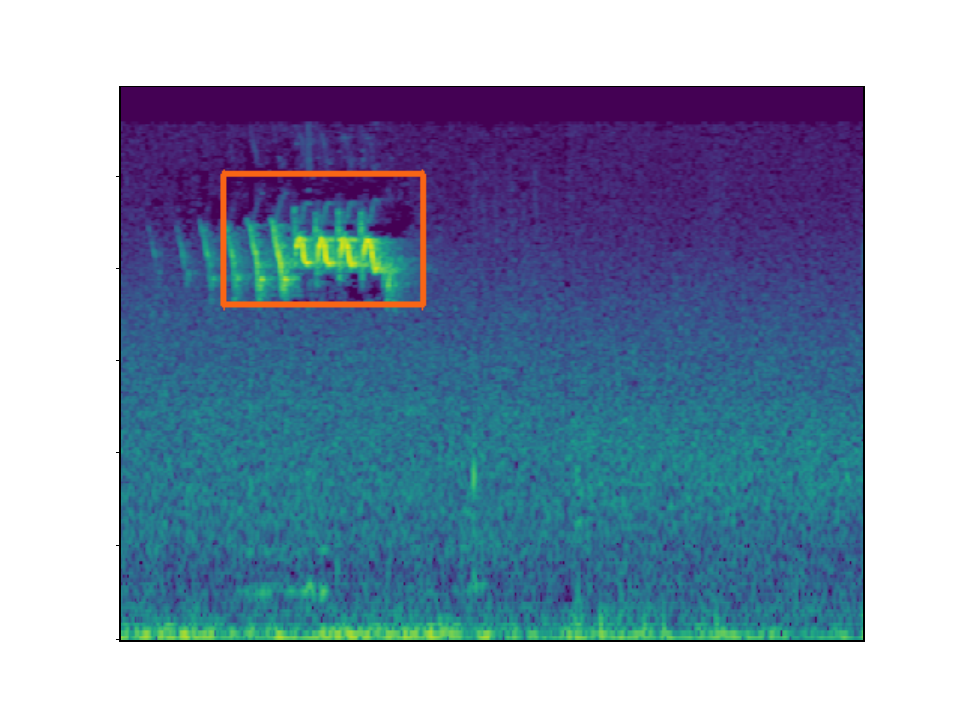}
                \par
                {\scriptsize $s^{(c,j)} = 0.35$}
                \par
                \scriptsize\textbf{yerwar}
            \end{minipage}
        \end{minipage}
    \end{minipage}

    \caption{The spectrograms of the most similar instances from the SNE subset of the training dataset, with their labels highlighted in bold, for the five prototypes learned by \gls*{audioprotopnet}-5 for the Yellow-rumped Warbler (yerwar). The prototypes
correspond to the parts of the spectrograms surrounded by the orange bounding boxes. Additionally, the similarity values to the prototypes $s^{(c,j)}$ and the weights of the respective prototypes $w^{(j,c)}$ in the final layer are shown. Prototype 1 also shows a high similarity to instances of the bird species Wilson's Warbler (wlswar) and Brown Creeper (brncre).}
\label{fig:prototypes_train_yerwar}
\end{figure}

As part of contribution (C4), we performed a qualitative analysis of the learned prototypes. 
For the qualitative analysis of the prototypes learned by \gls*{audioprotopnet}, we focused on the Sierra Nevada (SNE) training dataset provided by BirdSet.
The bird species inhabiting the Sierra Nevada region have well-documented vocalization types, which are described in detail in the literature \cite{pieplow2019peterson}. 
This enables the analysis of the learned prototypes without requiring extensive ornithological expertise.
The SNE training dataset is a subset of the XCL training dataset, containing only the bird species present in the SNE test dataset. 
To visualize the prototypes, we applied the procedure described in Section \ref{subsec:prototype_visualization} to the SNE subset of the training dataset only. 
For each prototype, we identified the five most similar instances from this subset. 
Since only the prototype learning classifier was trained in \gls*{audioprotopnet}, while the backbone network derived from the black-box ConvNeXt model remained frozen, the prototypes also serve as a post-hoc analysis of the most relevant embeddings learned by the ConvNeXt model. \\
In our analysis, we first examined the most similar examples of the learned prototypes for two bird species, the Mountain Chickadee (Poecile gambeli) and the Yellow-rumped Warbler (Setophaga coronata). 
We chose these species because they exemplify the characteristics and quality of the learned prototypes for other bird species, while revealing interesting insights and existing challenges of \gls*{audioprotopnet}. 
Figures \ref{fig:prototypes_train_mouchi} and \ref{fig:prototypes_train_yerwar} show the instances with the highest similarity to the prototypes of these two species.

\begin{figure}[ht]
    \centering
    \begin{subfigure}[b]{0.49\textwidth}
        \centering
        \includegraphics[width=\textwidth]{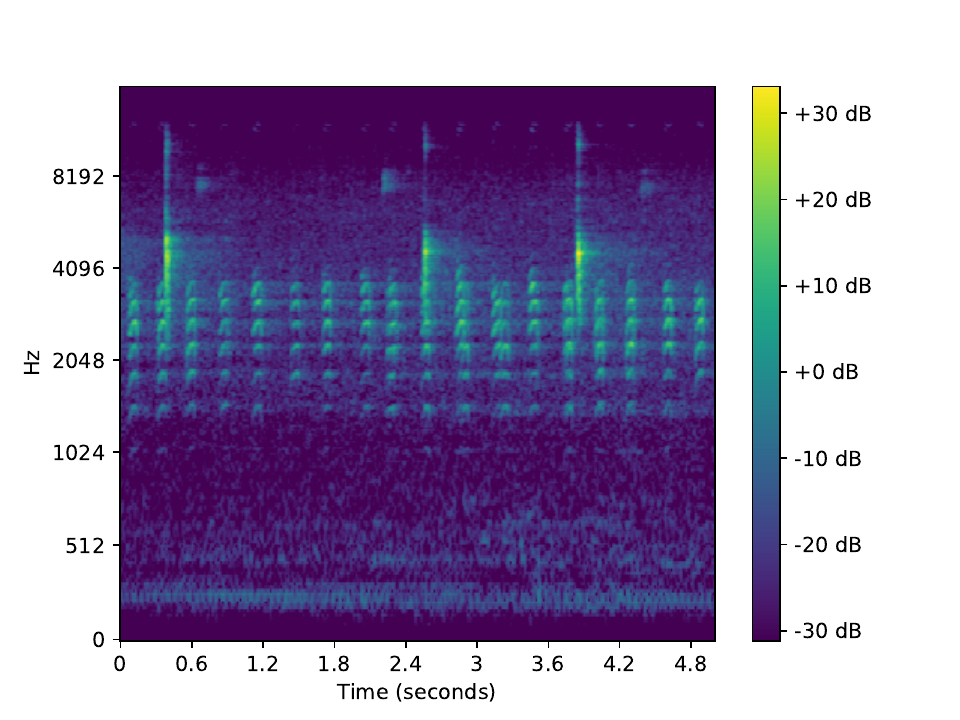}
        \caption{}
        \label{fig:a}
    \end{subfigure}
    \begin{subfigure}[b]{0.49\textwidth}
        \centering
        \includegraphics[width=\textwidth]{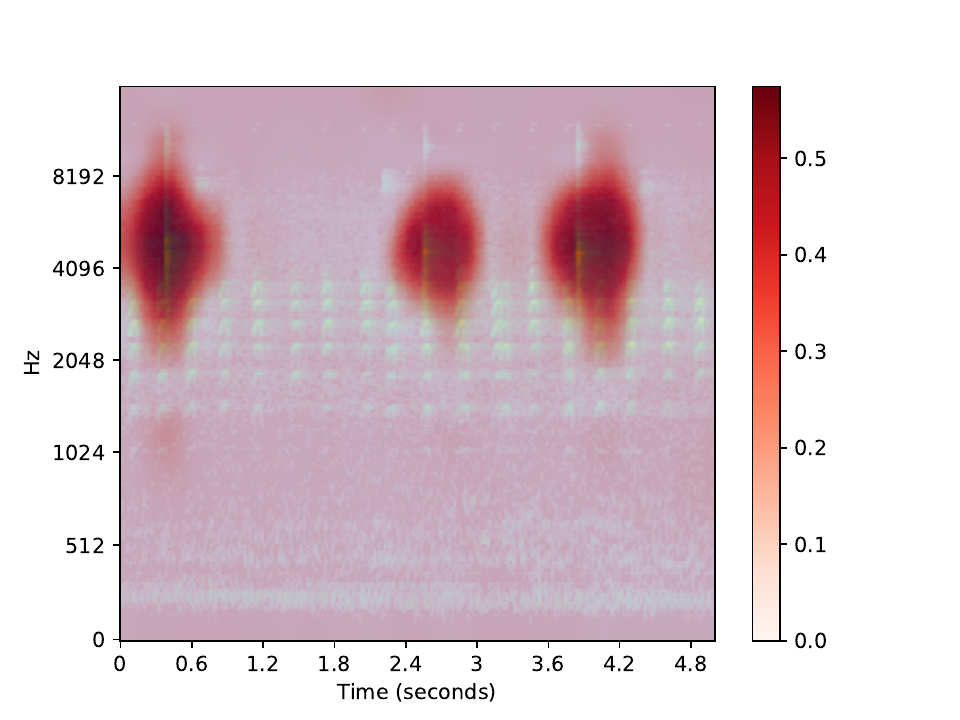}
        \caption{}
    \end{subfigure}
    
    \caption{Illustration of (a) the original spectrogram and (b) the spectrogram with a heatmap overlaid, visualizing the similarity of the regions of the fifth most similar instance of the SNE training dataset to prototype 1 of the Yellow-rumped Warbler. The call of the Yellow-rumped Warbler occurs three times in this spectrogram, and the prototype shows consistently high similarity at all three positions where the call occurs.}
    \label{fig:heatmap_yerwar}
\end{figure}

Figures \ref{fig:prototypes_train_mouchi} and \ref{fig:prototypes_train_yerwar} show that the model learned different types of vocalizations from the two bird species. 
Comparing these to the vocalizations described in \cite{pieplow2019peterson}, we find that prototype 2 in Figure \ref{fig:prototypes_train_mouchi} represents the whistled song of the Mountain Chickadee. Prototypes 3 and 4 represent different parts of the chick-a-dee call, while prototype 5 represents the gurgle song. 
For the Yellow-rumped Warbler, prototype 1 in Figure \ref{fig:prototypes_train_yerwar} represents its most common call, while prototypes 4 and 5 represent its song.
Many of the prototypes, such as prototype 3 for the Mountain Chickadee, are semantically meaningful and consistent. 
This means that they capture the same sound pattern under varying conditions, such as differences in volume and background noise. 
The model is capable of recognizing sound patterns at different temporal locations within a spectrogram and can identify sound patterns that occur multiple times within a spectrogram. 
As shown in Figure \ref{fig:heatmap_yerwar}, this enables a form of object detection. 

\begin{figure}[ht]
\centering

\begin{minipage}{\textwidth}
  \begin{minipage}{0.025\textwidth}
    \rotatebox{90}{\scriptsize\textbf{Original Spectrogram}}
  \end{minipage}
  \begin{minipage}{0.31\textwidth}
    \centering
    \includegraphics[width=\textwidth]{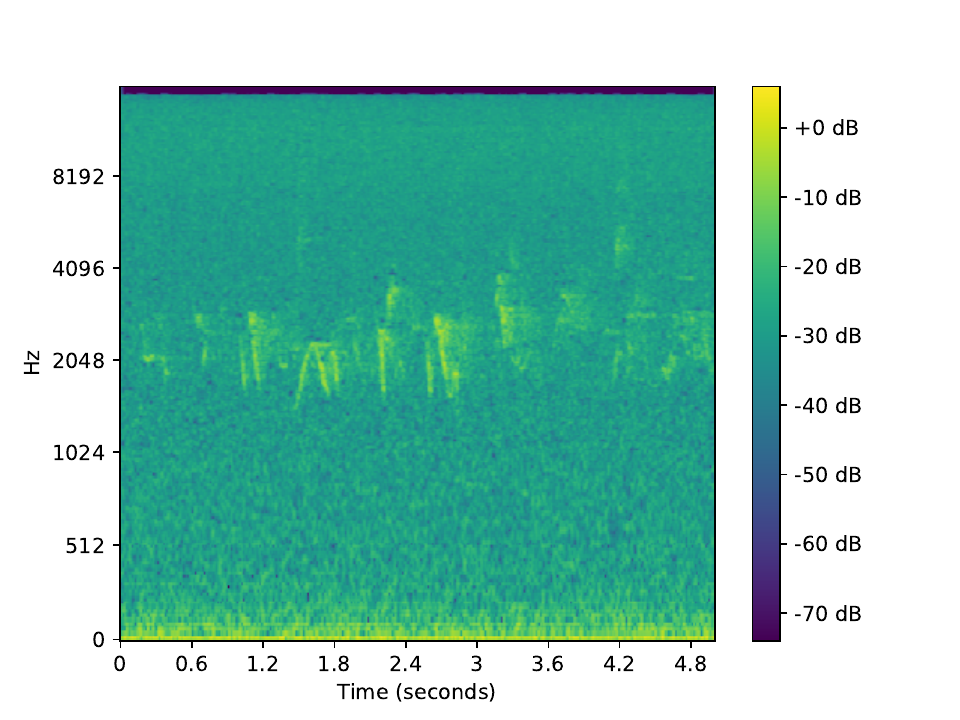}
    \par
    \scriptsize\textbf{amerob, bkhgro}
  \end{minipage}
  \begin{minipage}{0.025\textwidth}
    \rotatebox{90}{\scriptsize\textbf{Heatmaps}}
  \end{minipage}
  \begin{minipage}{0.31\textwidth}
    \centering
    \includegraphics[width=\textwidth]{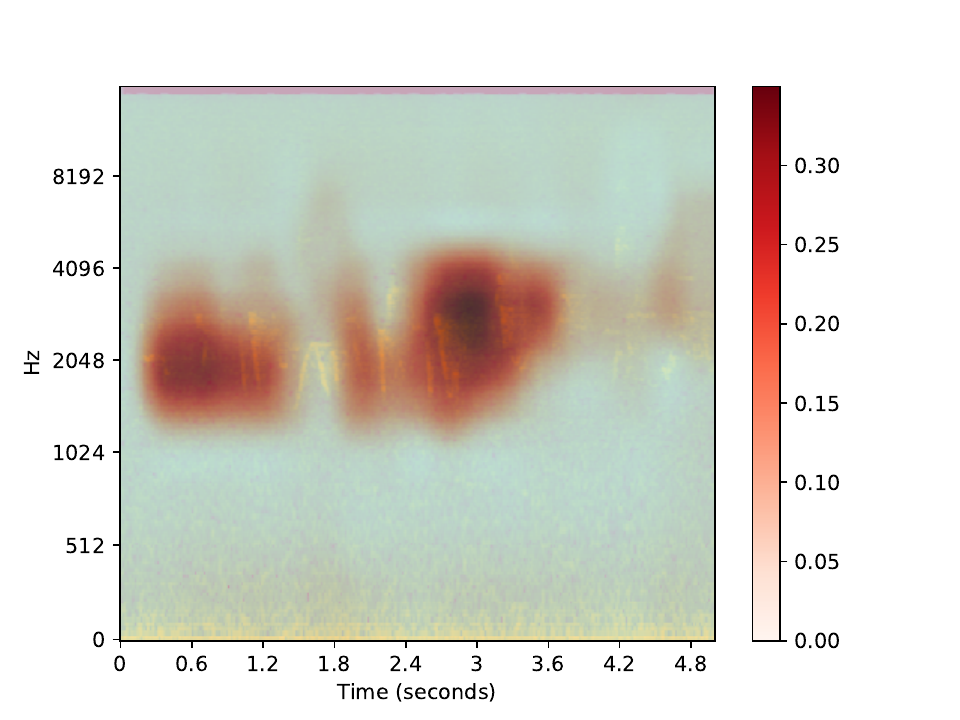}
    \par
    \textbf{\scriptsize{Similarity to prototype amerob}}
  \end{minipage}
  \begin{minipage}{0.31\textwidth}
    \centering
    \includegraphics[width=\textwidth]{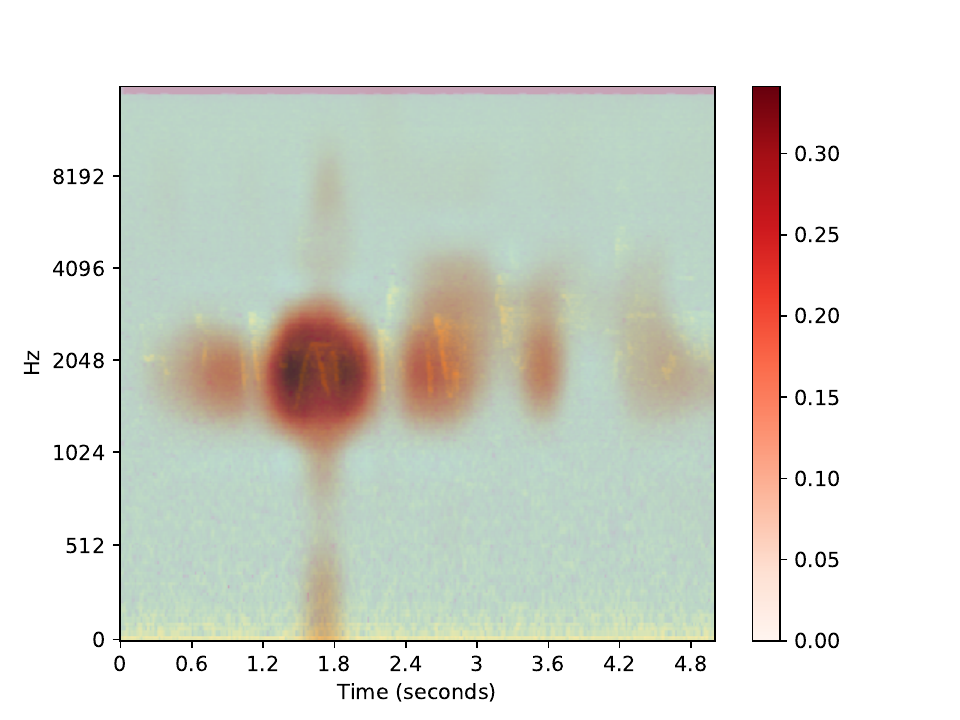}
    \par
    \textbf{\scriptsize{Similarity to prototype bkhgro}}
  \end{minipage}
  
\end{minipage}

\begin{minipage}{\textwidth}
  \begin{minipage}{0.025\textwidth}
    \rotatebox{90}{\scriptsize\textbf{Prototype amerob}}
  \end{minipage}
  \begin{minipage}{0.025\textwidth}
    \rotatebox{90}{\scriptsize $w^{(j,c)} = 1.22$}
  \end{minipage}
  \begin{minipage}{0.95\textwidth}
    \centering
    \begin{minipage}{0.19\textwidth}
      \centering
      \includegraphics[trim={57 35 40 20},clip, width=\textwidth]{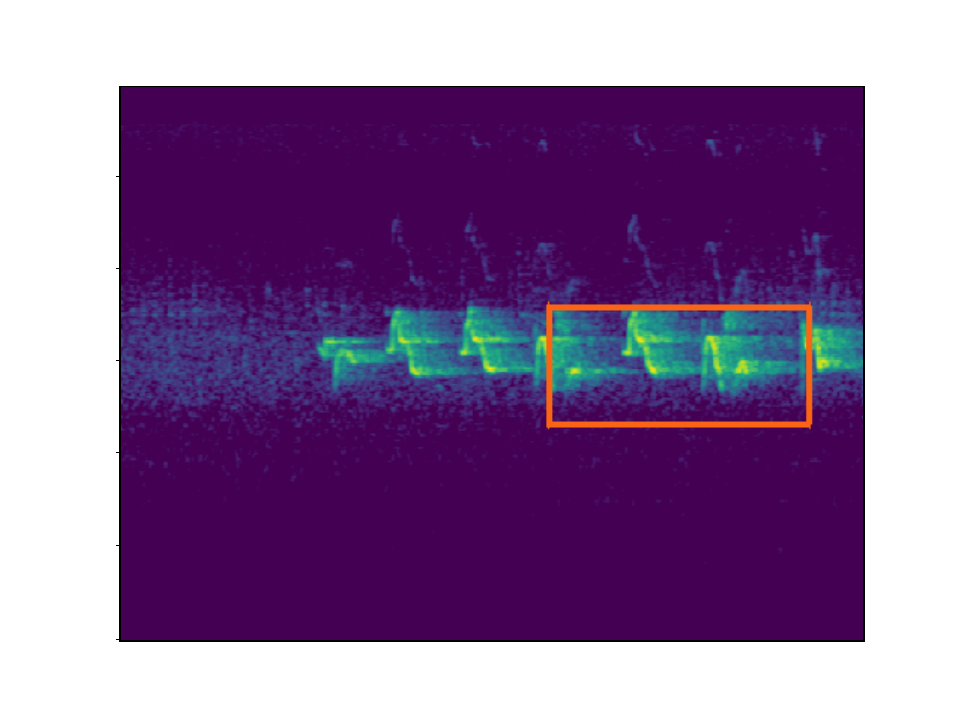}
      \par
      {\scriptsize $s^{(c,j)} = 0.47$}
      \par
      \scriptsize\textbf{amerob}
    \end{minipage}
    \begin{minipage}{0.19\textwidth}
      \centering
      \includegraphics[trim={57 35 40 20},clip, width=\textwidth]{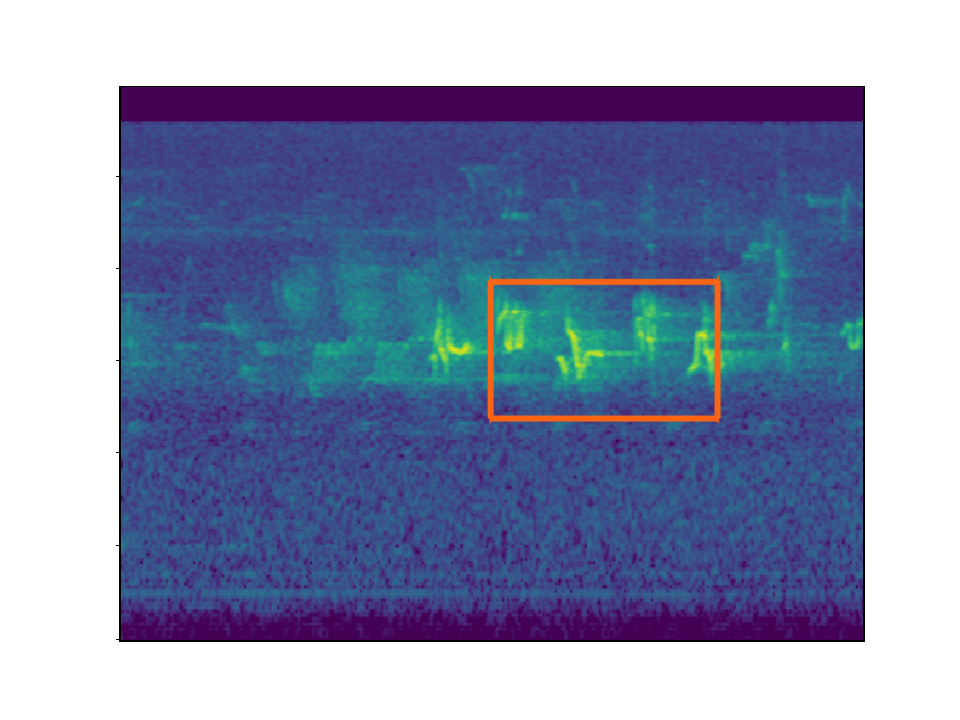}
      \par
      {\scriptsize $s^{(c,j)} = 0.46$}
      \par
      \scriptsize\textbf{amerob}
    \end{minipage}
    \begin{minipage}{0.19\textwidth}
      \centering
      \includegraphics[trim={57 35 40 20},clip, width=\textwidth]{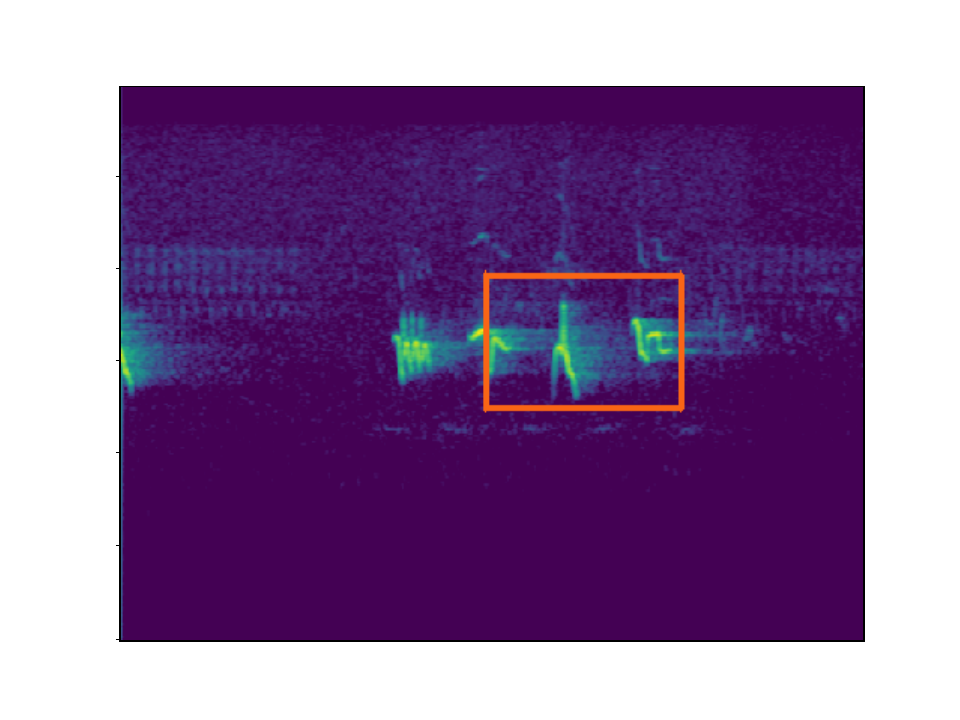}
      \par
      {\scriptsize $s^{(c,j)} = 0.45$}
      \par
      \scriptsize\textbf{amerob}
    \end{minipage}
    \begin{minipage}{0.19\textwidth}
      \centering
      \includegraphics[trim={57 35 40 20},clip, width=\textwidth]{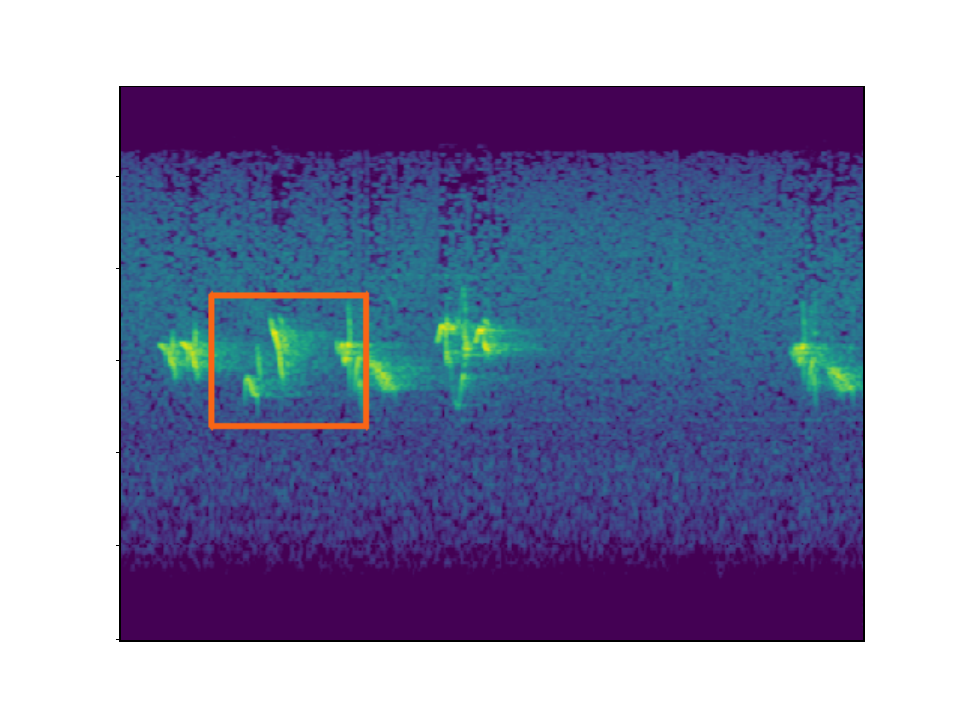}
      \par
      {\scriptsize $s^{(c,j)} = 0.44$}
      \par
      \scriptsize\textbf{amerob}
    \end{minipage}
    \begin{minipage}{0.19\textwidth}
      \centering
      \includegraphics[trim={57 35 40 20},clip, width=\textwidth]{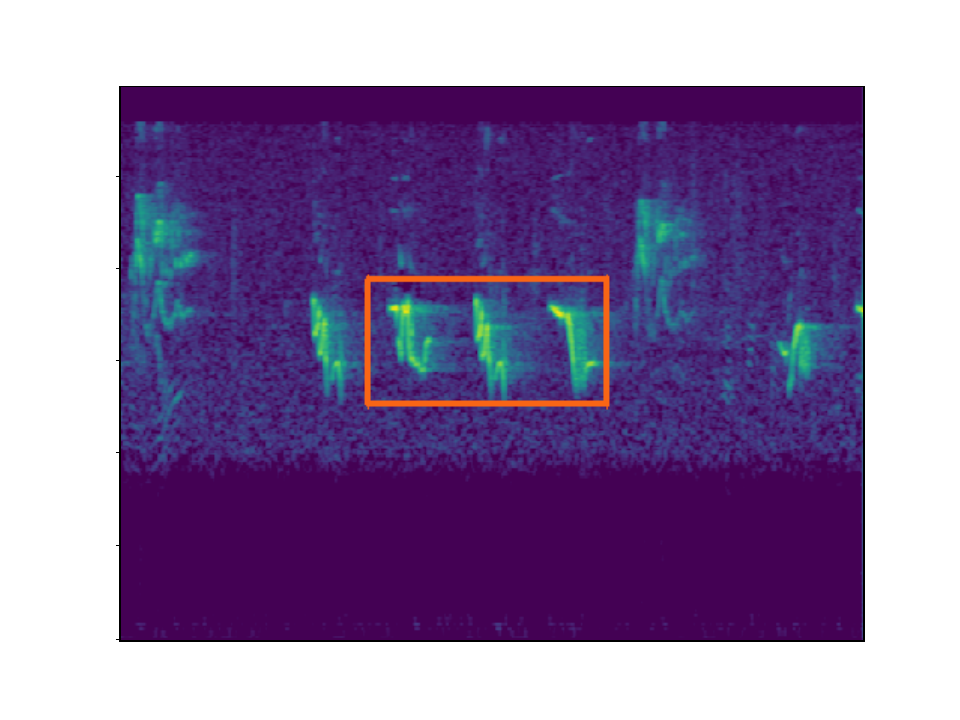}
      \par
      {\scriptsize $s^{(c,j)} = 0.43$}
      \par
      \scriptsize\textbf{amerob}
    \end{minipage}
  \end{minipage}
\end{minipage}

\begin{minipage}{\textwidth}
  \begin{minipage}{0.025\textwidth}
    \rotatebox{90}{\scriptsize\textbf{Prototype bkhgro}}
  \end{minipage}
  \begin{minipage}{0.025\textwidth}
    \rotatebox{90}{\scriptsize $w^{(j,c)} = 2.15$}
  \end{minipage}
  \begin{minipage}{0.95\textwidth}
    \centering
    \begin{minipage}{0.19\textwidth}
      \centering
      \includegraphics[trim={57 35 40 20},clip, width=\textwidth]{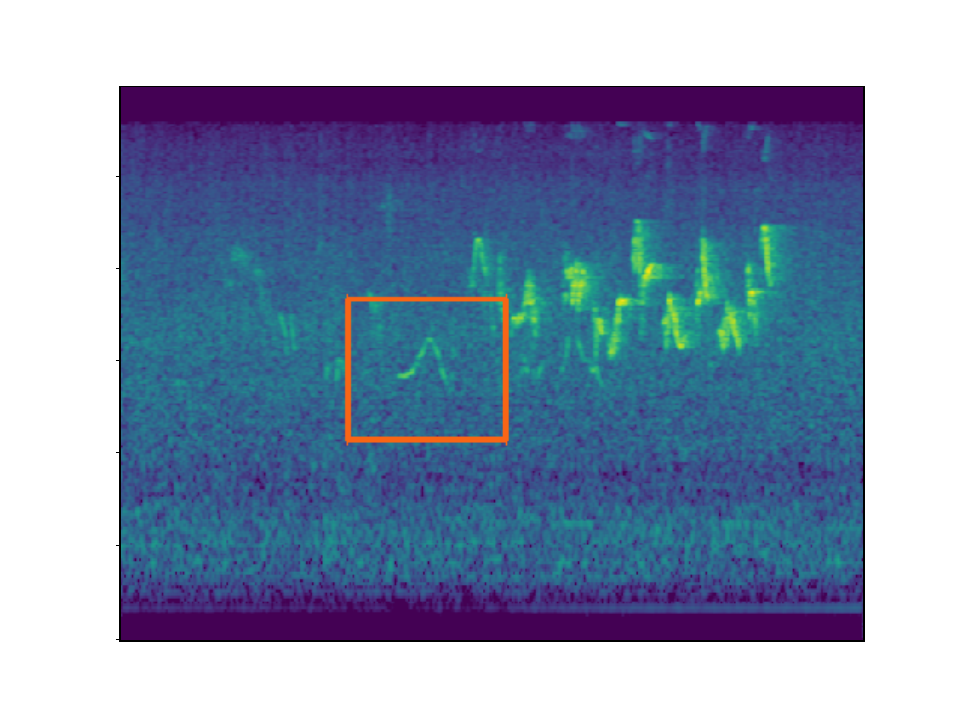}
      \par
      {\scriptsize $s^{(c,j)} = 0.57$}
      \par
      \scriptsize\textbf{warvir}
    \end{minipage}
    \begin{minipage}{0.19\textwidth}
      \centering
      \includegraphics[trim={57 35 40 20},clip, width=\textwidth]{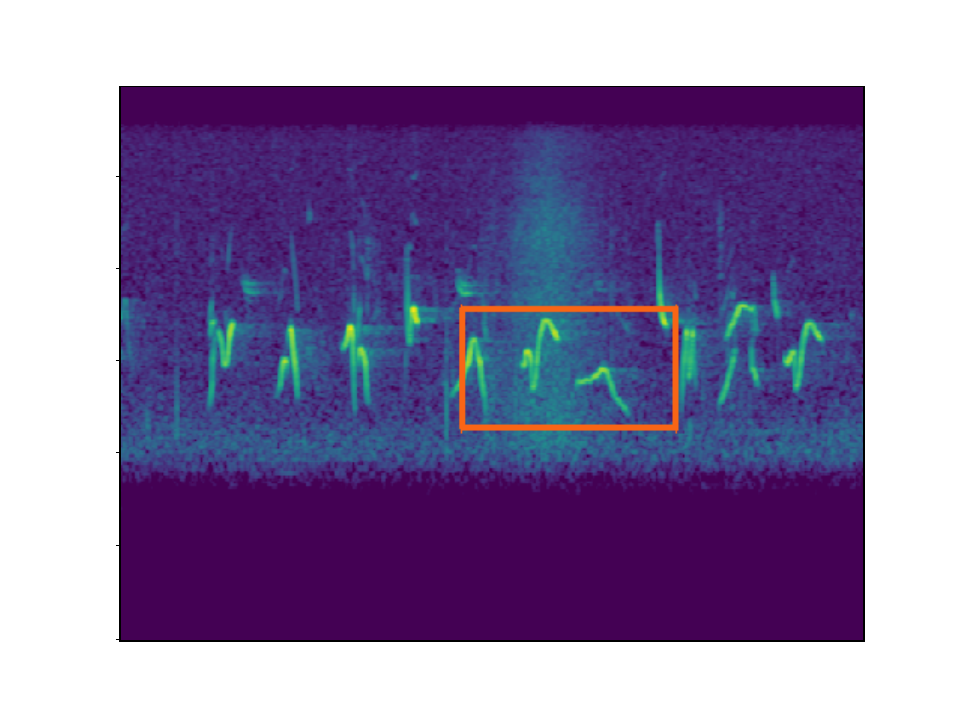}
      \par
      {\scriptsize $s^{(c,j)} = 0.56$}
      \par
      \scriptsize\textbf{bkhgro}
    \end{minipage}
    \begin{minipage}{0.19\textwidth}
      \centering
      \includegraphics[trim={57 35 40 20},clip, width=\textwidth]{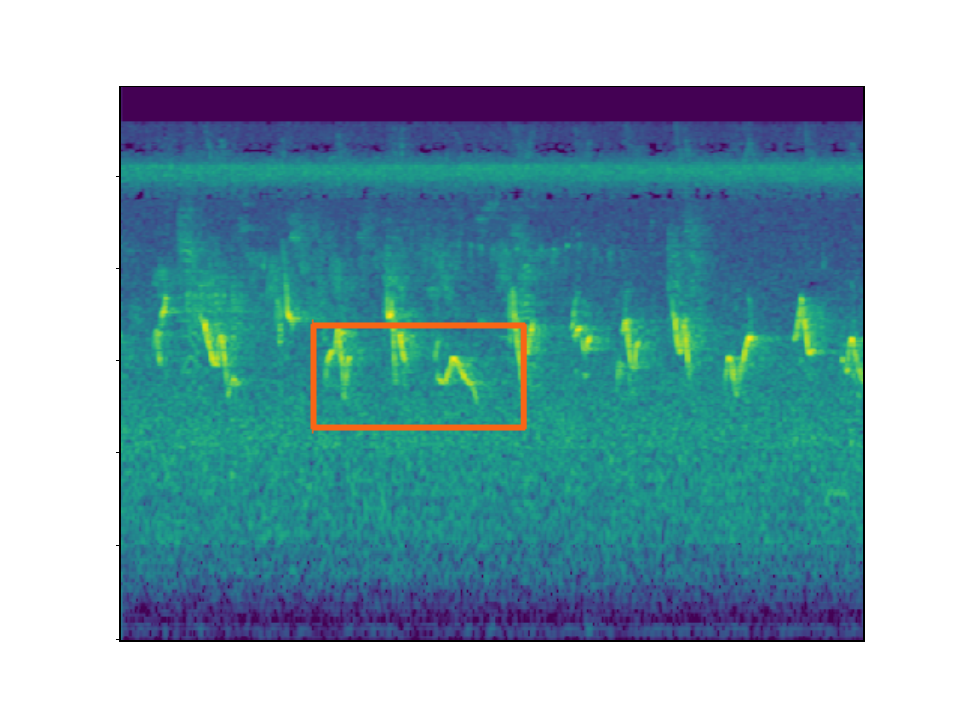}
      \par
      {\scriptsize $s^{(c,j)} = 0.53$}
      \par
      \scriptsize\textbf{bkhgro}
    \end{minipage}
    \begin{minipage}{0.19\textwidth}
      \centering
      \includegraphics[trim={57 35 40 20},clip, width=\textwidth]{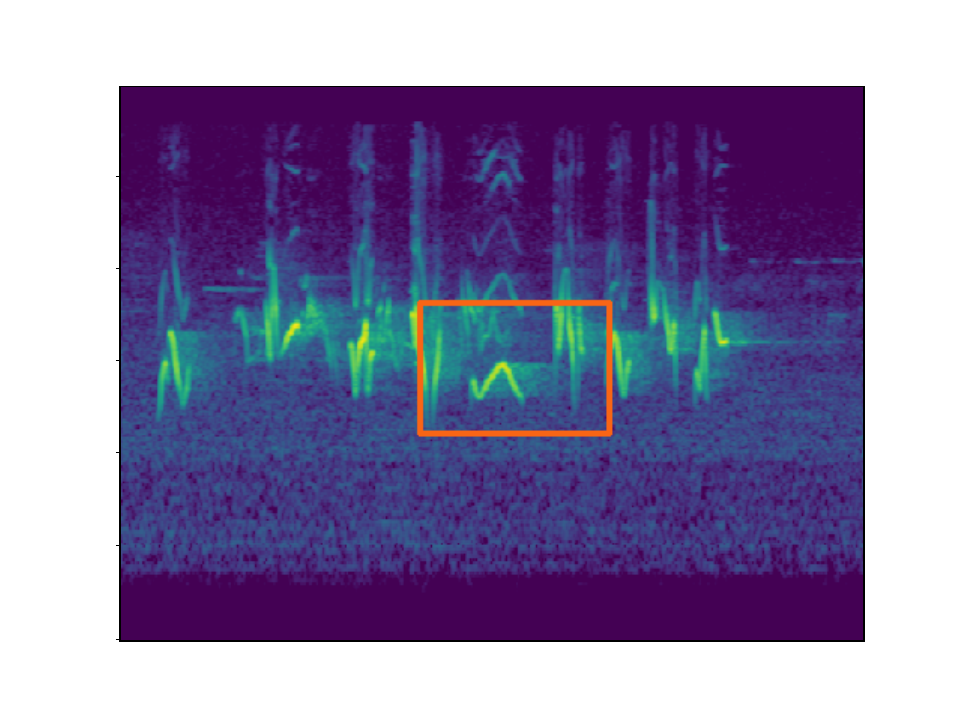}
      \par
      {\scriptsize $s^{(c,j)} = 0.52$}
      \par
      \scriptsize\textbf{bkhgro}
    \end{minipage}
    \begin{minipage}{0.19\textwidth}
      \centering
      \includegraphics[trim={57 35 40 20},clip, width=\textwidth]{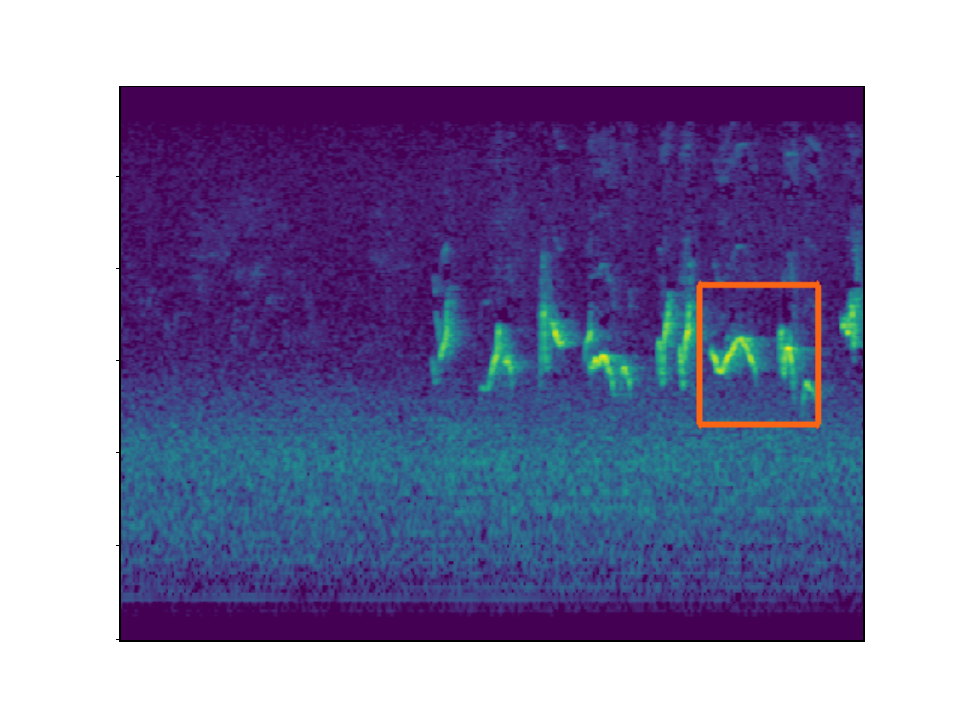}
      \par
      {\scriptsize $s^{(c,j)} = 0.52$}
      \par
      \scriptsize\textbf{bkhgro}
    \end{minipage}
  \end{minipage}
\end{minipage}

\caption{Visualization of a spectrogram of an instance from the SNE test dataset annotated with the bird species American Robin (amerob) and Black-headed Grosbeak (bkhgro). The original spectrogram is accompanied by two superimposed heatmaps showing which areas most closely resemble the prototypes most similar to the spectrogram. These prototypes are from the classes American Robin and Black-headed Grosbeak, respectively. In addition, the five instances most similar to these prototypes from the SNE training dataset are shown, with their labels highlighted in bold, to illustrate the characteristic sound patterns of each prototype. In the instance most similar to the prototype of the Black-headed Grosbeak, the call of the Black-headed Grosbeak appears in the background of a recording of the Warbling Vireo (warvir).}
\label{fig:local_analysis}
\end{figure}

Figure \ref{fig:heatmap_yerwar} shows that the most frequent call of the Yellow-rumped Warbler occurs three times in the given spectrogram. 
The corresponding similarity values in the heatmap indicate a high similarity to prototype 1 of the Yellow-rumped Warbler at all three locations where this call occurs.
The third and fourth most similar instances to prototype 1 of the Yellow-rumped Warbler are also noteworthy. 
Although these examples are from the species Wilson's Warbler (Cardellina pusilla) and Brown Creeper (Certhia americana), respectively, the sound patterns most similar to prototype 1 in both spectrograms bear a strong resemblance to those of the other three most similar instances. 
While the Wilson's Warbler has a very similar call to the Yellow-rumped Warbler, which could indicate misclassification by the model, the Brown Creeper does not have a call similar to the Yellow-rumped Warbler. 
This suggests possible label noise, which would require further verification by an ornithologist. 
This observation suggests that \gls*{audioprotopnet} may also be useful for identifying label noise.
It is also notable that the prototypes sometimes represent entire vocalizations, such as prototype 2 of the Mountain Chickadee with its whistled song, and sometimes only individual parts of vocalizations. 
For example, prototype 3 of the Mountain Chickadee represents the first part and prototype 4 represents the second part of the chick-a-dee call.
Figures \ref{fig:prototypes_train_mouchi} and \ref{fig:prototypes_train_yerwar} also show the current limitations and challenges of \gls*{audioprotopnet}. 
A comparison of the learned prototypes with the vocal repertoires of the Mountain Chickadee and the Yellow-rumped Warbler described in \cite{pieplow2019peterson} shows that the prototypes do not cover the entire sound repertoire of these species, but only a portion of it. 
For example, the model did not learn the flight and alarm calls of the Yellow-rumped Warbler, nor the alarm and contact calls of the Mountain Chickadee. 
This is problematic because the model is unlikely to recognize certain types of vocalizations that are not represented by the prototypes.
In addition, there are redundancies, such as in prototypes 4 and 5 of the Yellow-rumped Warbler, which both represent the same concept, namely the song of this species. 
Sometimes prototypes represent only background noise, as in prototype 3 of the Yellow-rumped Warbler, where the most similar instances are mainly random noise regions in the spectrograms. 
Some prototypes are also not semantically meaningful because they do not represent consistent sound patterns and show high similarity to many different vocalizations. 
An example is prototype 1 of the Mountain Chickadee, whose most similar instances include the chick-a-dee call, the gurgle song, and the whistle song.
Occasionally, prototypes are more similar to instances of other classes than to instances of their own class. 
For example, prototype 1 of the Yellow-rumped Warbler has a high similarity to an instance of the Wilson's Warbler. 
Given these problems, \gls*{audioprotopnet} still has considerable room for improvement. \\
The prototypes can also be used to better understand which sound patterns in a spectrogram were responsible for a classification. 
Figure \ref{fig:local_analysis} illustrates this using an instance from the SNE test dataset. 

The ground truth labels for the spectrogram of this instance are American Robin (Turdus migratorius) and Black-headed Grosbeak (Pheucticus melanocephalus), which are the two classes predicted by the model with the highest confidence scores. 
Figure \ref{fig:local_analysis} shows the spectrogram of this instance along with the most similar prototypes of American Robin and Black-headed Grosbeak. 
The heatmaps show the similarity of the spectrogram regions to these two prototypes.
In this example, the model successfully recognizes the distinct sound patterns of the bird species present in the spectrogram. 
It correctly focuses on the bell-shaped call of the Black-headed Grosbeak, with the corresponding prototype showing a high similarity to this call. 
Comparison with \cite{pieplow2019peterson} suggests that this is a call produced by begging juveniles and occasionally adult females. 
The prototype with the highest similarity is the song of the American Robin. 
Interestingly, this prototype has no similarity to the call of the Black-headed Grosbeak in the spectrogram, but only to the surrounding regions representing the song of the American Robin. 
This shows that in this example the model successfully separated the vocalizations of the two species.

\section{Discussion}
\label{sec:discussion}
In summary, our results show that \gls*{audioprotopnet} outperforms the current leading models in bird sound classification, thus defining the new state-of-the-art in the field. 
In particular, \gls*{audioprotopnet} outperforms the black-box model ConvNeXt, even though both models use the same backbone and therefore rely on identical embeddings. 
This result suggests that the prototype learning classifier of \gls*{audioprotopnet} is able to extract more meaningful and discriminative information from these embeddings than the fully connected classifier used by ConvNeXt. 
A possible reason for this could be the stronger inductive bias of the prototype learning classifier, which likely improves the model's ability to generalize to unseen data. 
In addition, the classifiers differ in how they handle embeddings.
ConvNeXt's fully connected classifier relies on embeddings that have undergone global average pooling, a process that aggregates spatial information. 
In contrast, \gls*{audioprotopnet}'s prototype learning classifier uses embeddings without global average pooling. 
This preserves temporal and spectral information, potentially allowing the prototype learning classifier to make more accurate classifications. 
However, further research is needed to fully understand the exact reasons for \gls*{audioprotopnet}'s improved performance. \\
A major challenge for \gls*{audioprotopnet} is determining the optimal number of prototypes per class. 
Following the original architecture of \gls*{ppnet} \cite{chen2019looks}, we assigned an equal number of prototypes to each bird species. 
However, since the sound repertoires of bird species vary greatly in complexity and diversity, it may be advantageous to assign prototypes based on the individual complexity of each species. 
If done manually, this would require considerable effort and in-depth domain knowledge. 
Therefore, a central research question for future work is how to develop methods to automatically determine the optimal number of prototypes for each class. \\
Our qualitative analysis of the prototypes learned by \gls*{audioprotopnet} showed that the model captured only a subset of the vocalization types of different bird species, rather than covering their entire sound repertoire. 
This suggests that the training dataset may not adequately represent all vocalization types. 
To address this issue, one could improve the data coverage of the training dataset by collecting and including data for the missing vocalization types. 
In addition, it may be beneficial to explicitly label the different vocalization types and include vocalization type classification in the classification problem along with species classification. 
This could force the model to learn all vocalization types, which should significantly improve bird sound classification in practical applications. 
Another possible solution would be to incorporate domain knowledge into the models' prototypes.
The inherent interpretability of \gls*{audioprotopnet} promotes deeper human understanding of classification results, allowing ornithologists and \gls*{ml} engineers to jointly optimize the model.
However, there is currently no mechanism to directly intervene in the selection or removal of prototypes. 
Introducing the possibility of human intervention could be promising, especially if \gls*{audioprotopnet} learns prototypes that are duplicates, do not cover the entire sound repertoire of a bird species, or do not capture the characteristic features of a bird species' vocalizations at all.
Although some progress has been made in the field of image recognition by integrating humans into the prototype verification process \cite{netzorg2023improving}, the incorporation of domain knowledge remains one of the biggest challenges in prototype learning \cite{rudin2022interpretable}.
Especially in ornithology, which has a wealth of expert knowledge, the incorporation of this knowledge into prototype learning models for bird sound classification is a promising direction for future research.

\section{Conclusion}
\label{sec:conclusion}
In this study, we present \gls*{audioprotopnet}, an inherently interpretable deep learning model for bird sound classification. 
Our experiments confirm the broad applicability and scalability of \gls*{audioprotopnet} in classifying audio recordings of nearly ten thousand different bird species in a multi-label setting. 
In contrast to previously used deep learning models for bird sound classification, \gls*{audioprotopnet} is characterized by its inherent interpretability. 
It provides deeper insights into the model's embeddings through class-specific prototypes, which represent vocalization patterns for each bird species in the embedding space and are learned from spectrograms of training instances.
Therefore, \gls*{audioprotopnet} has the potential to facilitate model debugging, knowledge discovery, and interdisciplinary collaboration between machine learning engineers and ornithologists.
Furthermore, our results show that \gls*{audioprotopnet} outperforms the bird sound classification model Perch on the BirdSet multi-label classification benchmark, making \gls*{audioprotopnet} the new state-of-the-art model in the field. 
However, the applicability of \gls*{audioprotopnet} is not limited to bird sound classification. 
\gls*{audioprotopnet} can potentially also be used to classify audio recordings of vocalizations of other animal groups such as insects, fish and mammals.
Our research thus makes an important contribution to the development of inherently interpretable deep learning models for the classification of bioacoustic data, and lays the foundation for a promising field of research.

\section*{Acknowledgements}
This work was carried out as part of the DeepBirdDetect project (Fkz. 67KI31040E) funded by the German Federal Ministry for the Environment, Nature Conservation, Nuclear Safety and Consumer Protection (BMUV). 

\section*{Code availability}
The code for the experiments and visualizations in this paper will be made available on GitHub upon acceptance.

\bibliographystyle{unsrt}  
\bibliography{audioprotopnet_arxiv} 

\appendix


\section{Importance of explainability for bird sound classification}
\label{secA1:importance_xai} 
The classification of bird sounds benefits greatly from the explainability of the models used, especially when it comes to recognizing different song types, call types, and dialects within the sound repertoire of different bird species. 
By using a prototype learning model such as \gls*{audioprotopnet}, prototypical sound patterns can be learned within the input spectrograms that contain these key features. 
Analysis of the learned prototypes provides a deeper understanding of the specific sound patterns used by the model to recognize each bird species.
The learned prototypes then serve not only as a basis for classifying new recordings based on their similarity to these prototypes, but also as local explanations for each individual classification result. 
A local explanation explains the model's classification process for a particular input by highlighting the features or patterns that contributed to that particular classification. 
For example, if the model classifies a recording as belonging to a particular bird species because it matches a prototype of a particular call type, then the local explanation highlights this pattern and helps to better understand the reasoning behind this individual classification.
In addition, analyzing all the learned prototypes in their entirety provides a global explanation of the model. 
A global explanation describes what patterns the model generally uses for its classifications and provides insight into the overall behavior of the model. 
In the case of bird sound classification, a global analysis of all prototypes for a particular bird species could show which song types, call types, and dialects the model has learned to recognize and which it has not. 
This shows whether the model covers the entire sound repertoire of a species, or whether it is only able to recognize certain types of vocalizations.
Building on the work of \cite{adadi2018peeking} and \cite{arrieta2020explainable}, who identified several reasons for the importance of \gls*{xai} in general, we derive below some advantages of using \gls*{ppnet} for bird sound classification. \\ \\
\textbf{Quality assurance and debugging} \\
Prototype learning allows \gls*{ml} engineers to gain deeper insight into the learned embeddings of their models. 
This helps identify and eliminate sources of error in data and models. 
For example, prototypes can be used to evaluate a model's ability to distinguish relevant signals, such as bird calls, from noise, such as traffic noise or human speech, that the model may be using incorrectly for classification. 
By visually or audibly analyzing the learned prototypes, humans can quickly see what sounds they represent. 
This helps determine whether the model is being influenced by spurious features during classification. 
The prototypes also help to assess whether the model can recognize the full repertoire of a bird species, i.e. the different songs, call types and dialects, or whether it has only learned to recognize certain vocalizations of a species. \\ \\
\textbf{Trust and acceptance} \\
Ornithologists' confidence in bird sound classification models is crucial for their acceptance and practical application. 
Providing insight into the embeddings used by the classification models through prototypes makes the correctness of the models verifiable. 
Understanding the specific sound patterns that a model has learned to recognize, as well as understanding the limitations of the model, helps to build confidence in the technology over time. \\ \\
\textbf{Knowledge discovery} \\
It has already been shown that \gls*{dl} models are superior to human abilities in certain areas, such as the recognition of blue whale calls \cite{miller2023deep}. 
In such cases, prototype learning offers promising opportunities for knowledge discovery in avian bioacoustic research. 
It can help discover previously unknown features in bird vocalizations that the \gls*{dl} model uses to distinguish bird species, but that have been overlooked by ornithologists. 
Detailed analysis of the learned prototypes of a \gls*{ppnet} could therefore reveal subtle differences in the acoustic signatures of bird species, such as variations in pitch, rhythm, and frequency, that were previously unknown to humans. 
This could deepen ornithologists' understanding of avian acoustic communication and contribute to the further development of biological theories. \\ \\
\textbf{Interdisciplinary collaboration} \\
Prototype learning also facilitates interaction between \gls*{ml} engineers and ornithologists. 
If the models and their classification results are explainable, it is easier to incorporate the domain knowledge of ornithologists into the modeling process. 
For example, if the model is not yet able to recognize certain songs, call types, or dialects of a bird species, additional recordings of these vocalizations could be included in the training data to improve classification. 
Also, additional penalty terms could be included in the loss function to encourage the model to learn prototypes representing the missing vocalization types. 
This can improve both the accuracy of the models and the alignment of the learned prototypes with human preferences and semantically meaningful features \cite{netzorg2023improving}. \\ \\
\textbf{Education} \\
Prototype learning models can also be used for educational purposes, helping citizen scientists understand how to identify bird species based on their vocalizations.  
In this way, explainability could also be used to support citizen science applications such as eBird \cite{sullivan2009ebird} or iNaturalist \cite{iNaturalist2024}. 
Explainable models could help citizen scientists better understand what features the model is using to classify bird sounds. 
This could give them clues about what specific sound patterns to look for to distinguish different species. 
Enabling laypeople to better understand bird sounds and their characteristics could improve the quality of volunteer contributions and make them more valuable to scientific research.




\end{document}